\documentclass[lettersize,journal,twoside]{IEEEtran}
\usepackage{amsmath,amsfonts}
\usepackage{algorithmic}
\usepackage{algorithm}
\usepackage{array}
\usepackage[caption=false,font=normalsize,labelfont=sf,textfont=sf]{subfig}
\usepackage{textcomp}
\usepackage{stfloats}
\usepackage{url}
\usepackage{verbatim}
\usepackage{graphicx}
\usepackage{cite}
\usepackage{hyperref}
\usepackage{url}

\usepackage{booktabs}
\usepackage{graphicx}
\usepackage{tabularx}

\usepackage{kotex}
\usepackage{multirow}
\usepackage{orcidlink}
\usepackage{placeins}
\usepackage{afterpage}
\usepackage{wrapfig}
\usepackage{amsfonts}
\usepackage{nicefrac} 
\usepackage{microtype}
\usepackage{xcolor} 
\usepackage{colortbl}
\usepackage{color}
\usepackage{amsmath}
\usepackage{caption}
\usepackage{float}
\definecolor{first}{HTML}{547CB1} %
\definecolor{improve}{HTML}{1E73C4} %

\newcommand{\tworow}[2]{\begin{tabular}[c]{@{}c@{}}#1\vspace{-2pt}\\#2\end{tabular}}
\usepackage{enumitem}
\usepackage[table]{xcolor}
\usepackage{colortbl}
\IEEEoverridecommandlockouts
\definecolor{lightgray}{gray}{0.92}
\newcommand{\NA}{\textcolor{lightgray}{\footnotesize N/A}}

\begin{document}

\title{On the Design of Mixture-of-Experts for Dynamic Gaussian Splatting
}

\author{In-Hwan Jin\orcidlink{0009-0008-9202-6510}, Hyeongju Mun\orcidlink{0009-0008-5289-5535}, Joonsoo Kim\orcidlink{0000-0002-6470-0773}, Kugjin Yun\orcidlink{0009-0002-7574-2853}, and Kyeongbo Kong\orcidlink{0000-0002-1135-7502},~\IEEEmembership{Member,~IEEE}

\thanks{Received 3 March 2026; revised 9 June 2026; accepted 10 July 2026. This work was supported by the National Research Foundation of Korea (NRF) grants funded by the Korean government (MSIT) (RS-2024-00414230, RS-2024-00456152). Computational resources were provided by ``Advanced GPU Utilization Support Program'' funded by the Government of the Republic of Korea (Ministry of Science and ICT) and the Cluster Server for Computational Science at Pusan National University. \textit{(Corresponding author: Kyeongbo Kong.)}}

\thanks{In-Hwan Jin and Hyeongju Mun contributed equally to this work. In-Hwan Jin and Hyeongju Mun are with the Department of Electrical and Electronics Engineering, Pusan National University, Busan, Republic of Korea (e-mail: ihjin@pusan.ac.kr; 201924128@pusan.ac.kr).}

\thanks{Kyeongbo Kong is with the Department of Electrical and Electronics Engineering and JYS AI Nexus Institute, Pusan National University, Busan, Republic of Korea (e-mail: kbkong@pusan.ac.kr).}

\thanks{Joonsoo Kim and Kugjin Yun are with the Immersive Media Research Section, Electronics and Telecommunications Research Institute (ETRI), Daejeon, Republic of Korea(e-mail: joonsookim@etri.re.kr; kjyun@etri.re.kr).}
}

\markboth
{IEEE TRANSACTIONS ON PATTERN ANALYSIS AND MACHINE INTELLIGENCE}
{JIN \MakeLowercase{et al.}: ON THE DESIGN OF MIXTURE-OF-EXPERTS FOR DYNAMIC GAUSSIAN SPLATTING}
\maketitle

\begin{abstract}
Dynamic scene reconstruction remains challenging due to the heterogeneous and spatially varying nature of real-world motion.
Although recent 3D Gaussian Splatting methods have introduced diverse deformation formulations for dynamic novel view synthesis, each method typically relies on a single deformation model within its representation, which limits robustness across diverse dynamic scenarios.
In this work, we study a fundamental problem—multi-deformation modeling for dynamic 3D Gaussian representations—under two distinct integration constraints that differ in when and how multiple deformation experts interact during training.
From a Mixture-of-Experts (MoE) perspective, we view multi-deformation modeling as the problem of combining multiple specialized deformation models within a unified 3D representation.
We first introduce \textit{Mixture of Deformation Experts (MoDE)}, which integrates multiple deformation experts directly into the deformable Gaussian Splatting pipeline through joint optimization.
In MoDE, experts operate on a shared canonical Gaussian representation, enabling multi-deformation modeling without introducing additional training stages or modifying the original optimization schedule.
In contrast, we further present \textit{Mixture of Experts for Dynamic Gaussian Splatting (MoE-GS)} under a different integration constraint, where deformation experts are optimized independently and combined through a separate routing stage.
As a result, expert interaction occurs over non-canonical Gaussian representations after individual optimization.
Together, these two approaches provide alternative strategies for multi-deformation modeling, 
clarifying how integration constraints shape the design and behavior of deformation experts 
in dynamic 3D Gaussian representations.
Our code is available at:  \url{https://github.com/cvsp-lab/MoE-GS-studio}.

\end{abstract}


\section{Introduction}
\afterpage{
\begin{figure*}[t!]
\centering  
\includegraphics[width=\linewidth]{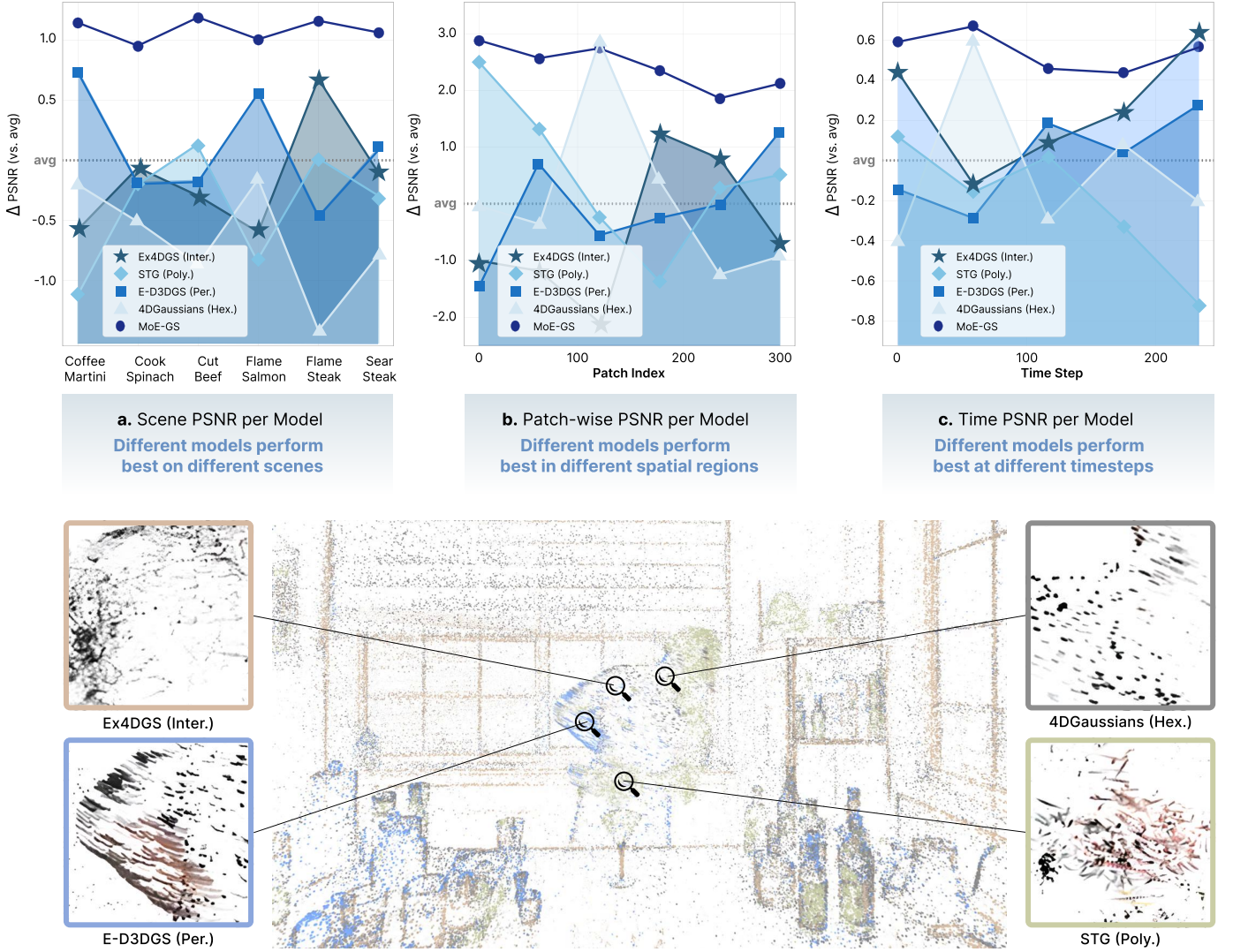}
\vspace{-3pt}
\caption{Limitations of existing dynamic Gaussian splatting methods. (a) Scene-level: No single method consistently dominates across scenes.
(b) Spatial-level: Different spatial regions favor different deformation models.
(c) Temporal-level: The best-performing method changes over time within the same scene.
We also visualize representative motion trajectories of four experts—4DGaussians \cite{wu20244d}(Gray), STG \cite{li2024spacetime}(Green), E-D3DGS \cite{bae2024per}(Blue), and Ex4DGS \cite{lee2024fully}(Beige)—to illustrate their distinct motion behaviors.}
\label{toyEX}
\end{figure*}
}
\IEEEPARstart{S}{patial} 
{
intelligence requires machines to perceive, represent, and reason about dynamic environments over time. In this context, world models aim to provide simulation-capable representations of the physical world, supporting prediction, planning, and interaction in complex environments. Dynamic scene representation is therefore a fundamental building block for such models, as it reconstructs the evolving geometry and appearance of real-world scenes from visual observations.
Realistically modeling such dynamic scenes is also essential for training future AGI models, creating immersive content for spatial computing, and enabling embodied agents to effectively perceive and interact with their environments. Recent advances in novel view synthesis, particularly Neural Radiance Fields (NeRF)~\cite{mildenhall2021nerf}, have significantly improved the quality of static scene reconstruction. However, NeRF's implicit representation and intensive ray-tracing introduce substantial computational overhead, limiting real-time applicability. To address these limitations, explicit representations such as 3D Gaussian Splatting (3DGS)~\cite{kerbl20233d} have emerged, achieving real-time rendering and fast training time without compromising visual fidelity.

Building upon the efficiency of Gaussian-based representations, recent efforts have extended them to dynamic scenes through diverse deformation mechanisms, including MLP-based deformation networks~\cite{wu20244d,bae2024per, xu2024grid4d},
polynomial-based motion models~\cite{li2024spacetime},
and interpolation-based methods~\cite{lee2024fully}.
These approaches aim to enable temporally consistent and physically plausible motion modeling while preserving the efficiency of Gaussian-based representations.
Despite these advances, our empirical analysis (Fig.~\ref{toyEX}) reveals that existing approaches do not consistently generalize across diverse real-world dynamic scenarios.
Specifically, our analysis identifies three key limitations:
\vspace{-2pt}
\begin{itemize}
\item \textbf{Scene-level variations}: Different reconstruction methods exhibit significant variability in performance across scenes, indicating that each model has a restricted range of optimal applicability (Fig.~\ref{toyEX}a). 
This observation suggests that reconstruction quality is highly scene-dependent, and that the effectiveness of a given deformation formulation can vary substantially across different scenes.
\item \textbf{Spatial-level inconsistencies}: Within the same scene, reconstruction quality varies across different spatial regions when processed by distinct methods (Fig. \ref{toyEX}b). This spatial variability demonstrates that no single model consistently excels across all regions of a given scene.
\item \textbf{Temporal fluctuations}: Within a video sequence, the best-performing method changes dynamically from frame to frame, reflecting the inherent temporal complexity of real-world dynamics (Fig. \ref{toyEX}c). Such temporal fluctuations highlight the inability of individual models to consistently capture complex temporal patterns.
\vspace{-2pt}
\end{itemize}

Importantly, the performance variations observed above are not incidental, but stem directly from how each method parameterizes deformation.
In dynamic Gaussian Splatting, the deformation formulation acts as a strong inductive bias, implicitly favoring specific spatial and temporal motion regimes.
While such inductive biases regularize optimization and improve stability within preferred motion regimes, they also limit the diversity of motions that can be faithfully represented in heterogeneous real-world scenes.

Figure~\ref{toyEX} (bottom) illustrates representative motion trajectories produced by different deformation models.
Although all methods operate on the same underlying Gaussian primitives, their distinct deformation formulations give rise to qualitatively different motion behaviors.
Models with strong spatial or temporal regularization tend to generate smooth and coherent trajectories, yielding robust performance in static or smoothly varying regions.
In contrast, more flexible formulations produce diverse and high-velocity trajectories that better capture fast or irregular motion, often at the expense of coherence in rigid or low-frequency regimes.
These qualitative differences in motion trajectories highlight that different deformation formulations induce fundamentally different motion behaviors, even when applied to the same underlying Gaussian primitives.

Taken together, these observations indicate that the limitations of existing dynamic Gaussian Splatting methods arise not from insufficient modeling capacity, but from their reliance on a single deformation prior to represent inherently heterogeneous motion.
Motivated by this insight, we investigate multi-deformation modeling, where multiple deformation models are combined within a unified 3D Gaussian representation.
This direction naturally aligns with the Mixture-of-Experts (MoE) paradigm, which addresses heterogeneous data by leveraging multiple specialized models rather than a single monolithic formulation.
A key design consideration in this context is \emph{when and how} multiple deformation experts are integrated into the reconstruction pipeline.
Specifically, deformation experts may be combined either during the core 3D Gaussian optimization through joint training within a shared pipeline, or through a separate routing stage following independent expert optimization.
Guided by this distinction, we explore two different integration strategies for multi-deformation modeling, leading to two structurally distinct approaches.

The first perspective integrates multiple deformation experts directly into the standard deformable Gaussian Splatting pipeline, which we refer to as \emph{Mixture of Deformation Experts (MoDE)}.
In this design, multiple deformation experts operate on a shared canonical Gaussian representation and are optimized jointly during training.
Conceptually, MoDE performs expert composition at the representation level: multiple experts generate candidate deformations for the same underlying 3D Gaussian primitives, and their outputs are combined within a unified 3D representation.
By preserving the original training schedule and optimization procedure, MoDE offers an efficient and minimally invasive extension of existing dynamic Gaussian Splatting methods.
Moreover, because all experts are anchored to a common canonical representation, MoDE naturally supports direct and consistent 3D reconstruction.
However, this canonical constraint limits the applicability of deformation
models that do not share a common representation, and joint optimization
can restrict the extent to which the complementary strengths of
individual experts are fully exploited.

The second perspective combines multiple deformation experts that are optimized independently in separate training runs, followed by a separate routing stage to combine their outputs, which we refer to as \emph{Mixture of Experts for Dynamic Gaussian Splatting (MoE-GS)}.
In contrast to MoDE, experts in MoE-GS do not share a common canonical representation, which precludes direct Gaussian-level correspondence across experts.
As a result, expert combination is performed in a shared image-space domain rather than at the level of individual Gaussians.
By decoupling expert optimization from expert combination, MoE-GS allows each deformation model to fully specialize within its preferred motion regime, leading to stable and well-behaved experts.
To mitigate the limitations of image-space combination, MoE-GS employs a volume-aware pixel routing mechanism that accounts for underlying 3D structure during expert integration.
This increased flexibility, however, comes at the cost of additional computation due to multiple training runs and the need for a separate routing stage, and requires additional strategies to recover a unified 3D reconstruction.

\begin{table}[t]
\centering
\caption{Qualitative comparison between MoDE and MoE-GS across key design trade-offs.
○: fully supported, △: partially supported, ×: not supported.}
\label{tab:comparison}
\begin{tabular}{lcc}
\toprule
\textbf{Criterion} & \textbf{MoDE} & \textbf{MoE-GS} \\
\midrule
Direct Gaussian-level 3D reconstruction      & ○ & × \\
Compatibility with non-canonical deformations& × & ○ \\
Training stability                           & △ & ○ \\
Training efficiency                          & ○ & △ \\
Expert specialization flexibility            & △ & ○ \\
\bottomrule
\end{tabular}
\end{table}

The key trade-offs between MoDE and MoE-GS are summarized in Table~\ref{tab:comparison}. MoDE and MoE-GS represent two distinct approaches to modeling multiple deformations in dynamic Gaussian Splatting, where different integration constraints induce different trade-offs in reconstruction fidelity, training stability, and computational efficiency. Rather than claiming a single universally optimal solution, we clarify the design space of deformation composition and show how alternative integration strategies systematically shape model behavior under practical constraints. Through systematic analysis and empirical evaluation, we demonstrate that multi-deformation modeling can be realized in qualitatively different ways, each better suited to specific deployment requirements. In summary, our contributions are threefold:
\begin{itemize}
\item We provide a systematic analysis of dynamic Gaussian Splatting methods, revealing that performance variations across scenes, spatial regions, and time are closely tied to the deformation priors underlying each method.

\item Motivated by this observation, we investigate multi-deformation modeling from a Mixture-of-Experts perspective and introduce two distinct integration strategies, \textbf{MoDE} and \textbf{MoE-GS}, which differ in when and how multiple deformation experts are combined.

\item We analyze and compare these two approaches, highlighting their respective trade-offs in 3D reconstruction fidelity, training stability, and computational efficiency, and clarifying the design space of deformation composition for dynamic Gaussian representations.
\end{itemize}

\section{Related Works}
\vspace{3pt}
\subsection{Dynamic Novel View Synthesis}
\vspace{2pt}
Dynamic novel view synthesis aims to reconstruct novel views of temporally varying scenes from a limited set of observations given camera poses and timestamps, and can be regarded as a more general setting that extends novel view synthesis for static scenes into the temporal domain~\cite{bergen1991plenoptic}. While modeling spatial consistency is the primary concern in static scenes, dynamic scenes require jointly accounting for time-varying geometry and appearance, which significantly increases the difficulty of the problem~\cite{levoy2023light}. Effectively modeling such complex scene dynamics typically demands advanced temporal modeling techniques, which often lead to increased computational complexity and memory consumption.
To model dynamic scenes, prior works have proposed various approaches that either decouple a canonical space from time-dependent deformations or directly extend spatiotemporal representations. NeRF\cite{yin2025ms,verbin2024ref}-based methods~\cite{guo2023forward, liu2022devrf, park2021hypernerf, pumarola2021d, song2023nerfplayer, wang2021neural, park2021nerfies, duisterhof2024deformgs} provide flexible representations by learning deformation fields to capture temporal variations. However, these approaches suffer from substantially increased training and rendering costs due to dense ray sampling and additional neural network inference. Although several efforts have been made to alleviate these limitations~\cite{cao2023hexplane, fridovich2023k, lin2023high, shao2023tensor4d, wang2023masked, wang2023neural}, they still face significant bottlenecks in simultaneously achieving real-time rendering and modeling complex dynamic motions.
Against this backdrop, recent studies have focused on extending the efficiency and representational power of 3D Gaussian Splatting~\cite{kerbl20233d} to dynamic scenes, and Gaussian-based representations have emerged as a promising alternative for dynamic novel view synthesis.

\vspace{3pt}
\subsection{Dynamic 3D Gaussians}
\vspace{2pt}
Recent in Gaussian-based neural rendering,
particularly 3D Gaussian Splatting (3DGS)~\cite{kerbl20233d}
and its numerous extensions~\cite{chen2025hac++,qu2024z}, have motivated a growing body of work
that extends their rendering efficiency and expressive power
to dynamic scenes. Unlike NeRF-based approaches, Gaussian-based methods adopt an explicit, rasterization-based representation, enabling fast training and real-time rendering. Owing to these advantages, Gaussian representations have emerged as a promising alternative for dynamic novel view synthesis.

Within this line of research, many Gaussian-based dynamic scene models~\cite{wu20244d,bae2024per,lee2024fully, li2024spacetime, sun20243dgstream, liu2026dynamics, yan20244d, liu2024swings, duan20244d, lin2024gaussian, luiten2024dynamic, yang2024deformable, kratimenos2024dynmf, liang2025gaufre, huang2024sc, yu2024cogs,chen2025dash,wu2025localdygs,chen2026haif,jiang2025timeformer,wang2025freetimegs,gao20257dgs, liu2025modgs, yang2024real, yang20244d, zhang2025mega, cho20264d, wang2025shape}
explore diverse strategies for extending static Gaussian representations
to dynamic environments by introducing time-varying Gaussian attributes
or motion representations.
Recent works further broaden this design space by considering casually captured
monocular videos~\cite{liu2025modgs,wang2025shape} and native 4D Gaussian
primitive representations~\cite{yang2024real,yang20244d,zhang2025mega,cho20264d}.
These approaches differ in how they model temporal variations in scenes,
leading to distinct inductive biases across methods. One representative class of approaches models scene dynamics implicitly
by learning deformation fields that map points from a canonical space to
observed frames.
These methods typically rely on MLPs, while differing in how spatial and
temporal information is embedded.
For example, 4DGaussians~\cite{wu20244d} combine coordinate-conditioned embeddings with HexPlane-style feature representations and employ lightweight MLPs to estimate Gaussian deformations over time. E-D3DGS~\cite{bae2024per} observes that individual Gaussians exhibit distinct dynamic behaviors and proposes a deformation model that combines per-Gaussian latent embeddings with temporal embeddings, together with a coarse-to-fine strategy~\cite{feichtenhofer2019slowfast} to capture deformations at multiple temporal scales. Grid4D~\cite{xu2024grid4d} also adopts a canonical Gaussian formulation and predicts temporal deformations, but replaces plane-based factorization with an explicit feature representation that decomposes the 4D spatiotemporal domain into multiple 3D hash grids for deformation prediction. Despite their architectural differences, these methods share the common principle of modeling temporal variations through deformation fields. In contrast, several approaches avoid canonical deformation fields altogether and instead explicitly parameterize the temporal evolution of Gaussians. Spacetime Gaussians (STG)~\cite{li2024spacetime} model Gaussian motion and rotation using polynomial functions and introduce time-dependent opacity to represent dynamic changes. Ex4DGS~\cite{lee2024fully} explicitly defines Gaussian positions and rotations at sparse keyframes and reconstructs continuous temporal dynamics via interpolation~\cite{bartels1995introduction, shoemake1985animating} between adjacent keyframes.
Collectively, these Gaussian-based dynamic scene rendering methods explore different temporal modeling strategies and representation designs to address dynamic novel view synthesis. In this work, we investigate an alternative direction by integrating multiple Gaussian-based approaches within a unified mixture-of-experts framework.

\vspace{-3pt}
\subsection{Mixture of Experts}
\vspace{2pt}

MoE is an ensemble \cite{lakshminarayanan2017simple} learning technique where multiple expert models specialize in distinct subtasks, with a gating network dynamically selecting the most relevant experts per input instance. MoE architectures have demonstrated scalability and efficiency by introducing sparsity \cite{shazeer2017outrageously, fedus2022switch, chen2023mod, lewis2021base, hazimeh2021dselect, ma2018modeling, chi2022representation}, enabling conditional computation to scale model capacity without excessive computational cost \cite{he2021fastmoe, rajbhandari2022deepspeed, yu2024moesys, he2022fastermoe, singh2023hybrid, nie2023flexmoe, zhai2023smartmoe}. This approach has been particularly successful in large-scale deep learning models, including large language models (LLMs) for machine translation \cite{lepikhin2021gshard, fedus2022switch}, multimodal and vision--language models (VLMs) \cite{li2025uni, yu2025moe, yu2024boosting}, and computer vision applications such as multi-task learning \cite{ma2018modeling}, face forgery detection \cite{kong2022efficient}, and anomaly detection \cite{meng2024moead}.
While MoE has traditionally been studied in vision primarily from the perspective of computational sparsity and efficiency, recent vision-based MoE research~\cite{mi2025learning} shows that superior accuracy can be achieved even without relying on computational sparsity. In particular, combining heterogeneous expert models and properly calibrating their predictions can outperform any single expert. For example, object detection approaches such as Mixture of Calibrated Experts (MOCAE)\cite{oksuz2023mocae} and post-hoc detector calibration frameworks demonstrate that effectively integrating diverse expert predictions improves accuracy without introducing additional conditional computation. These findings suggest that MoE should not be viewed solely as a mechanism for reducing computation, but also as a model combination strategy for enhancing representational capacity and robustness.
Inspired by this perspective, MoE-GS~\cite{jin2026moe} applies mixture-of-experts to Dynamic Gaussian Splatting through rendering-level expert routing. It demonstrates that heterogeneous Gaussian-based reconstruction models can be adaptively combined according to scene characteristics.
Building upon this line of research, the present work introduces MoDE, a structurally distinct integration design that composes multiple deformation experts within a shared canonical Gaussian representation. We examine the structural differences between MoE-GS~\cite{jin2026moe} and MoDE, and analyze how their respective integration designs influence geometric representation, optimization behavior, and rendering quality. This comparative analysis offers a deeper understanding of how expert integration strategies affect dynamic scene representation.

\vspace{-4pt}

\section{Method}
\vspace{3pt}
We present two MoE formulations for dynamic Gaussian Splatting under different integration constraints.
Sec.~\ref{sec:prelim} first introduces the standard MoE formulation and notation used throughout the paper.
Sec.~\ref{sec:mode} presents \textit{MoDE}, which composes multiple canonical deformation experts within a shared Gaussian representation through temporally continuous gating.
Sec.~\ref{sec_moegs} then introduces \textit{MoE-GS}, which integrates independently trained heterogeneous experts without enforcing a shared canonical space, and combines them via a volume-aware routing mechanism.
Together, these two formulations reveal how different integration
constraints lead to fundamentally different design choices in expert interaction, optimization strategy, and reconstruction behavior.

\begin{figure*}[tb]
\begin{center}
\centerline{\includegraphics[width=\textwidth]{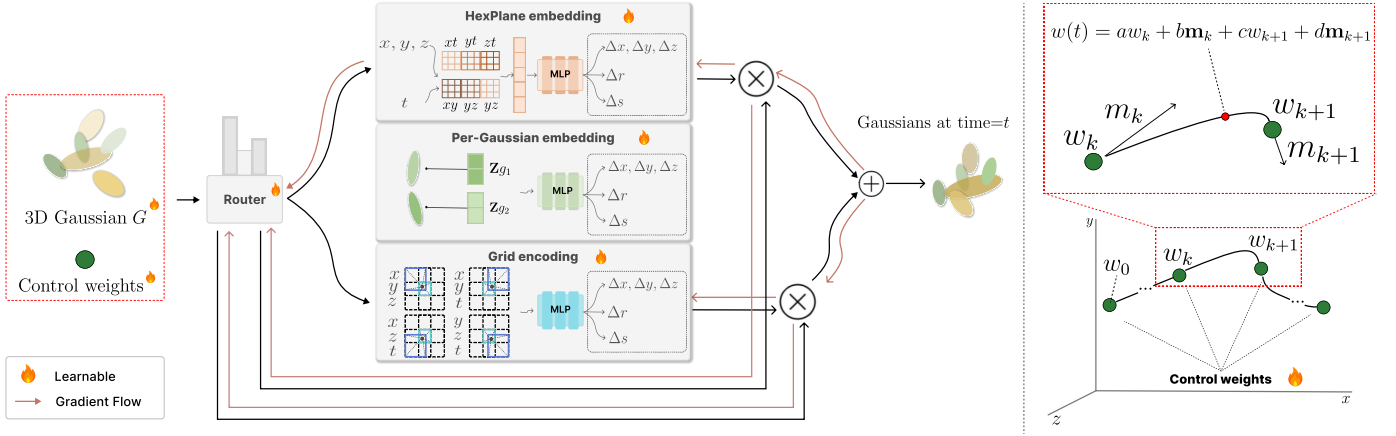}}
\vspace{-3pt}
\caption{Overview of Mixture of Deformation Experts (MoDE). MoDE augments a canonical Gaussian deformation model with multiple deformation experts that share a common canonical Gaussian representation.
Each expert encodes a distinct deformation prior using different embedding strategies (e.g., HexPlane, per-Gaussian embedding, grid encoding) and predicts time-dependent Gaussian deformations.
A router assigns continuous, time-varying gating weights to each expert, which are interpolated via a spline-based formulation (right).
The final deformation at time $t$ is obtained by a weighted composition of expert outputs.
During training, gradient propagation to the canonical Gaussians is restricted to the baseline expert to ensure stable optimization, as indicated by the gradient flow.}

\label{mode}
\vspace{-2mm}
\end{center}
\end{figure*}

\begin{figure*}[tb]
\begin{center}
\centerline{\includegraphics[width=\textwidth]{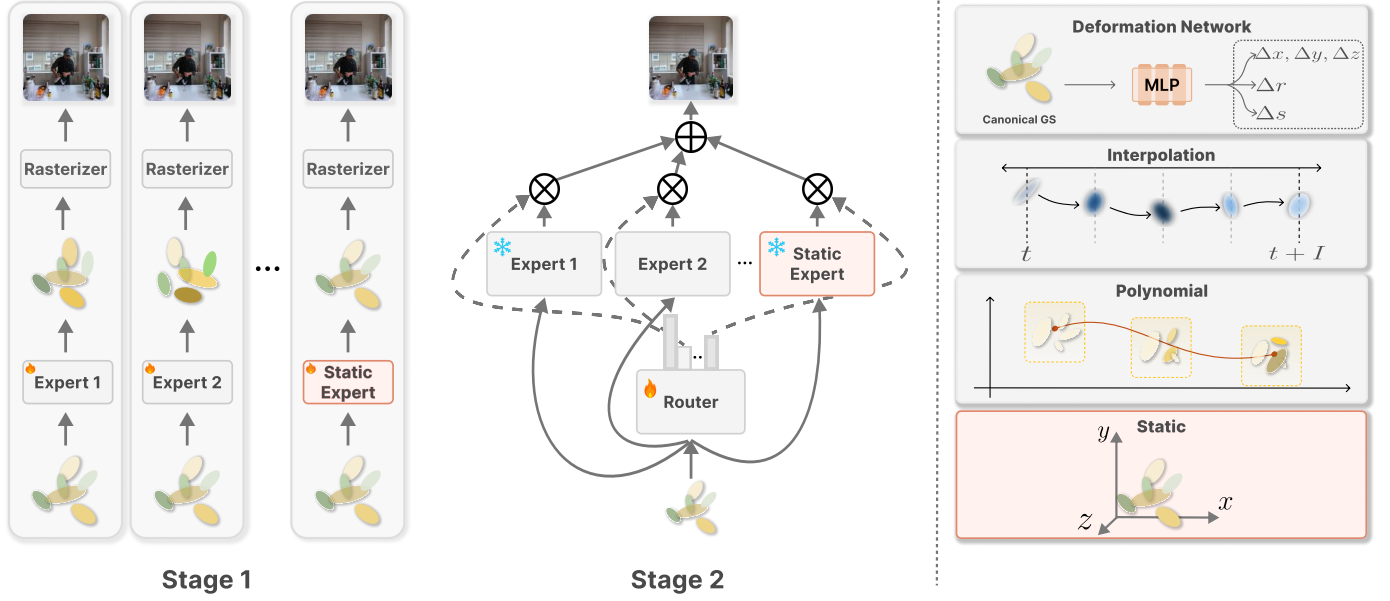}}
\vspace{-3pt}
\caption{
    Overview of the MoE-GS framework. In Stage 1 (Expert Training), each expert is independently trained to reconstruct the dynamic scene by optimizing its own Gaussian representation, ensuring diverse modeling capabilities. In Stage 2 (Router Training), with all expert parameters fixed, the Volume-aware Pixel Router learns to dynamically blend expert-rendered images by computing spatially and temporally adaptive gating weights. The Candidate Experts (right) illustrate diverse Gaussian-based reconstruction methods integrated into our framework, including deformation-network-based, interpolation-based, polynomial-based, and static approaches, each designed to model complementary motion and appearance characteristics of dynamic scenes.}

\label{framework}
\vspace{-10pt}
\end{center}
\end{figure*}

\vspace{-3pt}
\subsection{Preliminary}
\vspace{2pt}
\label{sec:prelim}

We briefly review the standard Mixture-of-Experts (MoE) architecture \cite{shazeer2017outrageously}, which forms the basis for our proposed method. A standard MoE consists of multiple parallel expert networks \( E_1, E_2, \dots, E_N \) and a \textit{Router} that adaptively combines expert outputs based on the input.
Formally, given an input \( x \), the MoE output is computed as follows:
\begin{equation}
\text{MoE}(x) = \sum_{k=1}^{N} G_k(x)\cdot E_k(x),
\end{equation}
where \( E_k(x) \) is the output of the \( k \)-th expert, and \( G_k(x) \) represents the corresponding \textit{gating weight} computed by the Router.
The gating weights are typically computed by a Router, often implemented as a lightweight neural network (e.g., linear layer or MLP), defined as:
\begin{equation}
G_k(x) = \text{Softmax}(R_k(x)),
\end{equation}
where \( R_k(x) \) is the router output for the \( k \)-th expert, computed as:
\begin{equation}
R_k(x) = x^\top W_{r,k},
\end{equation}
with \( W_{r,k} \) being the \( k \)-th column vector of trainable weight matrix \( W_r \). This structure enables adaptive blending of experts based on input characteristics, enhancing performance and flexibility.
In the following sections, we detail how this MoE formulation is applied to dynamic Gaussian Splatting under two distinct integration strategies.

\subsection{Mixture of Deformation Experts (MoDE)}
\label{sec:mode}

We begin by introducing \textit{MoDE}, a representation-level integration design of the standard MoE formulation for dynamic scene reconstruction (Fig.~\ref{mode}).
Conceptually, MoDE applies the MoE principle directly to Gaussian deformation: given the same canonical Gaussian input, multiple deformation experts independently predict candidate deformations, and their outputs are adaptively combined to produce the final deformed Gaussian representation.
Specifically, all deformation experts in MoDE operate on a shared set of canonical Gaussians.
Each expert encodes a distinct deformation prior and produces an expert-specific deformation output for the same underlying Gaussian primitives.
These candidate deformations are then aggregated through time-dependent gating weights, analogous to the weighted summation of expert outputs in standard MoE models.
As a result, the time-varying deformation of each Gaussian is represented as a weighted combination of expert-specific deformation outputs, enabling the model to adaptively compose complementary motion behaviors within a single representation. This requirement naturally motivates the use of canonical Gaussian deformation models, where temporal motion is expressed as deviations from a shared canonical representation.

\subsubsection{Candidate Canonical Deformation Experts}
Under this constraint, MoDE restricts its expert set to dynamic Gaussian
deformation models that follow the canonical Gaussian formulation.
These methods share a common structure: a set of canonical Gaussians is defined in a reference space, and temporal deformation is modeled by a learnable deformation network that maps canonical coordinates to time-dependent positions.
Formally, each Gaussian is represented in a canonical space as
\begin{equation}
\mathcal{G}_i = \{ \mathbf{X}_i, \mathbf{r}_i, \mathbf{s}_i, \sigma_i, \mathbf{C}_i \},
\end{equation}
where $\mathbf{X}_i$, $\mathbf{r}_i$, and $\mathbf{s}_i$ denote the canonical position, rotation, and scale, respectively.
Given a time step $t$, a canonical deformation model extracts a time-dependent deformation feature
\begin{equation}
\mathbf{f}_i(t) = \Psi(\mathcal{G}_i, t),
\end{equation}
where $\Psi(\cdot)$ denotes a model-specific feature encoding function.
Despite differences in how $\Psi(\cdot)$ is instantiated across methods, the deformation itself is predicted through a common multi-head mapping:
\begin{equation}
\Delta \mathbf{X}_i = \phi_x(\mathbf{f}_i), \quad
\Delta \mathbf{r}_i = \phi_r(\mathbf{f}_i), \quad
\Delta \mathbf{s}_i = \phi_s(\mathbf{f}_i).
\end{equation}
The deformed Gaussian is then given by
\begin{equation}
(\mathbf{X}', \mathbf{r}', \mathbf{s}')
= (\mathbf{X} + \Delta \mathbf{X},\;
   \mathbf{r} + \Delta \mathbf{r},\;
   \mathbf{s} + \Delta \mathbf{s}).
\end{equation}

While they follow the same canonical paradigm, existing canonical deformation methods differ substantially in how canonical information is embedded and how deformation is parameterized, resulting in distinct inductive biases over motion.

\paragraph{HexPlane-based Canonical Deformation (4DGaussians~\cite{wu20244d})}
4DGaussians models deformation using a shared spatio-temporal HexPlane embedding
that conditions a deformation MLP.
Because all Gaussians query a common canonical feature field,
the induced deformation signals are highly correlated across space,
leading to strong spatial regularization and smooth motion trajectories.
This prior favors stable reconstruction in static or slowly varying regions.

\paragraph{Hash-Encoded Canonical Deformation (Grid4D~\cite{xu2024grid4d})}
Grid4D adopts a higher-capacity spatio-temporal hash encoding
within the canonical deformation framework.
By relaxing low-rank assumptions in the feature representation,
it can capture more localized and higher-frequency motion patterns,
inducing a bias toward sharper and more abrupt temporal variations.

\paragraph{Per-Gaussian Embedding-based Deformation (E-D3DGS~\cite{bae2024per})}
E-D3DGS assigns learnable per-Gaussian embeddings to condition deformation,
enabling more flexible and locally adaptive motion modeling.
This formulation supports higher-frequency yet spatially aligned deformation,
making it effective for structured but dynamic regions.

Although all three experts follow the canonical Gaussian formulation,
their distinct embedding and parameterization strategies impose
different motion priors.
These complementary deformation characteristics motivate
the adaptive composition mechanism in MoDE.

\subsubsection{Temporal Gating via Weight Spline}
To enable temporally adaptive composition of canonical deformation experts, we model expert gating weights as a continuous function of time using a lightweight spline-based formulation.
For each Gaussian, we introduce a discrete set of learnable control weights that parameterize the temporal evolution of expert contributions:
\begin{equation}
\mathcal{W} = \{ \mathbf{w}_k \mid \mathbf{w}_k \in \mathbb{R}^{E},\; k = 0, \dots, N_w - 1 \},
\end{equation}
where $E$ denotes the number of deformation experts and $N_w$ is the number of control points.
These per-Gaussian control weights define expert preferences at discrete temporal
anchors and are interpolated to obtain a continuous gating signal.

Given a normalized time $t_n \in [0,1]$, the time-dependent weight vector is computed as
\begin{equation}
\mathbf{w}(t) = \mathcal{S}(t_n;\, \mathcal{W}),
\end{equation}
where $\mathcal{S}(\cdot)$ denotes a cubic Hermite spline.
This construction ensures first-order temporal continuity while allowing flexible
expert transitions without introducing additional temporal networks.
An illustration of the spline-based temporal gating mechanism is provided in Fig.~\ref{mode} (right).

The interpolated weights $\mathbf{w}(t)$ serve as unnormalized expert scores.
Let $\mathcal{K}(t)$ denote the index set of the Top-$K$ elements of
$\mathbf{w}(t)$.
We compute sparse gating weights by applying softmax over the selected
experts:
\begin{equation}
G_i(t) =
\begin{cases}
\operatorname{Softmax}\!\big(\mathbf{w}(t)\big)_i,
& i \in \mathcal{K}(t), \\
0, & \text{otherwise},
\end{cases}
\end{equation}
where the softmax is computed only over the indices in $\mathcal{K}(t)$.
The Top-$K$ constraint suppresses weak responses and reduces interference between experts,
while softmax normalization ensures stable weighting among selected experts.
This sparse gating mechanism encourages specialization while preserving smooth
temporal transitions.

Finally, the gated expert responses are aggregated directly at the Gaussian
attribute level.
Let $\mathbf{A}_i(t)$ denote the deformation output of expert $i$.
The final Gaussian attributes are computed as
\begin{equation}
\mathbf{A}(t) = \sum_{i=1}^{E} G_i(t)\, \mathbf{A}_i(t).
\end{equation}
Through this process, each Gaussian primitive adaptively composes complementary
deformation behaviors over time while maintaining temporal coherence.

\subsubsection{Training Strategy}
Training MoDE requires careful optimization due to the interaction between multiple deformation experts and a shared canonical Gaussian representation.
We adopt two complementary strategies to stabilize training and encourage balanced expert utilization.

(i) \textit{Stabilizing the canonical Gaussian space.}
Naively updating the canonical Gaussians using gradients from all experts
often leads to unstable convergence, as different deformation priors may
induce conflicting gradient signals in the shared canonical space.
To address this issue, only the baseline deformation expert is allowed
to update the canonical Gaussian parameters, while gradients from the
remaining experts are blocked at the canonical level.
This preserves a stable reference space while still allowing diverse
deformation behaviors to be learned through expert-specific networks.
For clarity, we visualize this asymmetric gradient flow in the framework overview (Fig.~\ref{mode}).

(ii) \textit{Random gating warm-up.}
During early training, learned gating weights may prematurely favor
a single expert due to random initialization or transient optimization noise.
To prevent this self-reinforcing behavior, we randomly activate experts
with equal probability for a fixed number of initial iterations.
After this warm-up phase, the spline-based gating mechanism is enabled,
allowing experts to specialize according to their deformation strengths.

\begin{figure*}[tb]
\begin{center}
\centerline{\includegraphics[width=\linewidth]{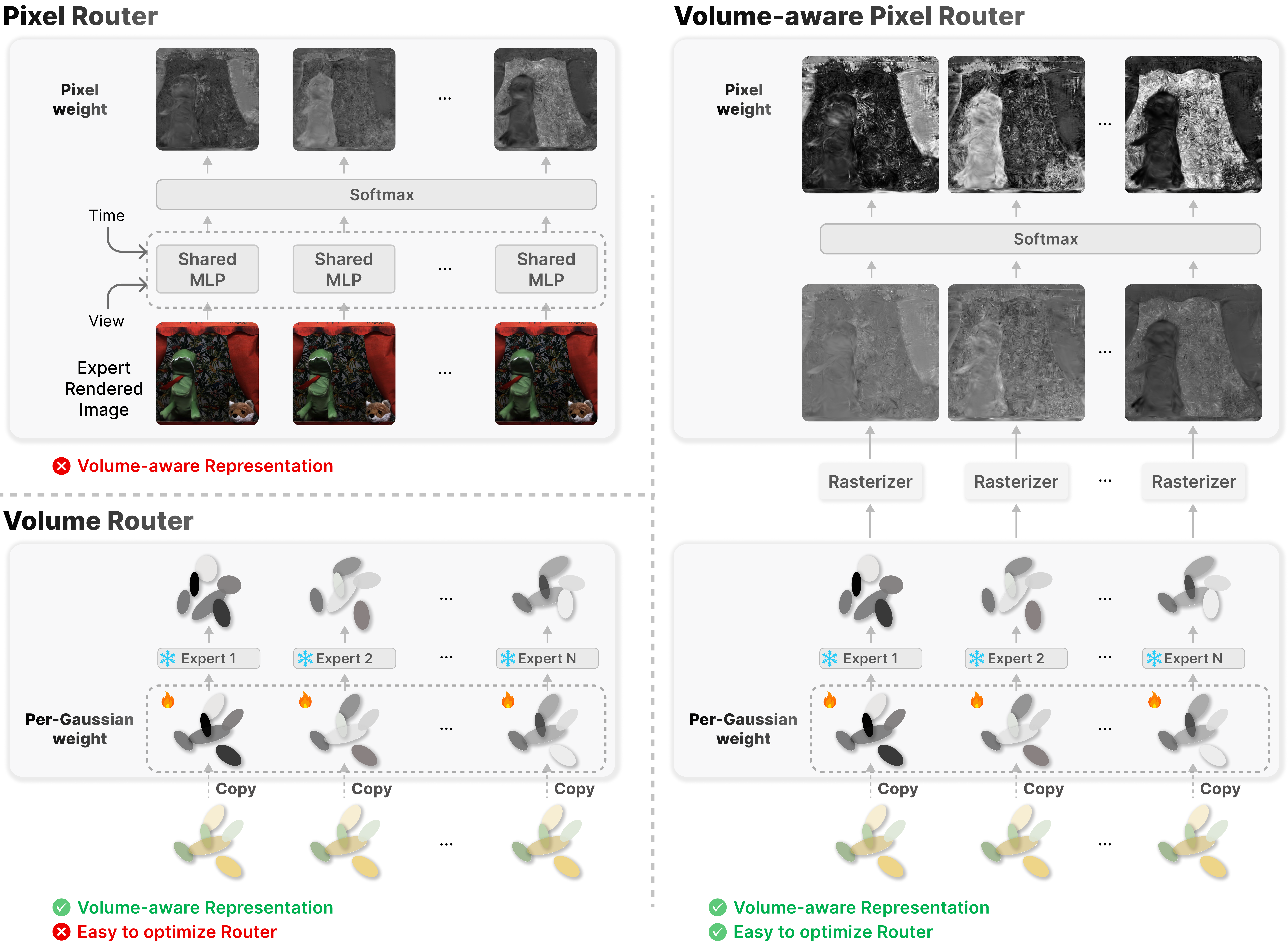}}
\caption{
Comparison of Router Architectures.
The Pixel Router (top-left) assigns weights purely at the pixel level, ignoring volumetric features. The Volume Router (bottom-left) uses Gaussian-level weights but is difficult to optimize. Our Volume-aware Pixel Router (right) combines Gaussian-level weights with rasterization-based splatting.
}
\label{Router_detail}
\end{center}
\vspace{-10pt}
\end{figure*}

\vspace{-5pt}
\subsection{Mixture of Experts for Dynamic Gaussian Splatting}
\label{sec_moegs}
\vspace{2pt}

While MoDE provides a direct and efficient way to compose multiple deformation behaviors within a shared canonical Gaussian representation, its joint optimization setting inherently couples expert behavior through the underlying deformation pipeline.
As a result, the degree of expert specialization achievable in MoDE is constrained by the shared representation and training dynamics.
In contrast, many practical dynamic reconstruction scenarios involve multiple Gaussian-based models that are independently designed and optimized, each adopting fundamentally different motion parameterizations and inductive biases.
In such cases, direct composition at the deformation level becomes ill-defined,
as these models do not share a common canonical representation.

Motivated by this setting, MoE-GS decouples expert optimization from expert integration.
Each expert model is trained independently until convergence, allowing it to fully specialize within its preferred motion regime.
Expert interaction is then introduced through an additional routing stage, which determines how expert outputs should be selected and combined in the absence of shared Gaussian correspondence.
As illustrated in Fig.~\ref{framework}, MoE-GS treats each independently trained dynamic Gaussian reconstruction model as an expert and introduces a learnable routing mechanism to integrate their outputs.

\subsubsection{Candidate Heterogeneous Gaussian Experts.}
Unlike MoDE, which requires all experts to share a canonical Gaussian representation, MoE-GS places no such constraint on expert design.
This allows MoE-GS to integrate a diverse set of dynamic Gaussian models with fundamentally different motion parameterizations and inductive biases.
As illustrated on the right side of Fig.~\ref{framework}, our framework incorporates both canonical deformation-based models and non-canonical trajectory-based models as experts, enabling composition across heterogeneous motion representations.

\paragraph{Canonical deformation experts}
Canonical deformation models parameterize temporal motion through a shared reference space and a learnable deformation field.
These experts excel at capturing spatially coherent motion and provide stable reconstructions in static or smoothly varying regions.
Within MoE-GS, canonical experts offer strong regularization and serve as reliable baselines for consistent deformation modeling.

\paragraph{Keyframe interpolation-based motion modeling (Ex4DGS~\cite{lee2024fully})}
Ex4DGS models motion through independent keyframe interpolation, allowing each Gaussian to update its position without enforcing a shared deformation field.
As a result, Gaussians may follow free-form and multi-directional trajectories, even within local neighborhoods.
This formulation is well suited to abrupt, irregular, or highly non-rigid motion regimes, where stronger structural priors may oversmooth the dynamics.
However, the lack of explicit spatial coherence can lead to instability in regions dominated by rigid or low-frequency motion.

\paragraph{Polynomial trajectory-based motion modeling (STG~\cite{li2024spacetime})}
STG parameterizes motion using low-order polynomial trajectories for each Gaussian.
This induces a strong bias toward globally smooth, low-curvature motion, yielding directionally consistent trajectories over time.
Such an inductive bias makes STG particularly effective in scenes dominated by rigid or near-rigid global motion, but limits its ability to represent highly irregular or rapidly changing dynamics.

\paragraph{Static experts (3DGS-MCMC~\cite{kheradmand20243d})}
Most dynamic Gaussian reconstruction methods extend vanilla 3D Gaussian Splatting~\cite{kerbl20233d} to model temporal deformation.
In parallel, recent work has focused on improving static Gaussian Splatting through enhanced densification strategies, optimization schemes, and regularization techniques.
In MoE-GS, we incorporate such static Gaussian Splatting variants as additional experts by training them on individual frames.
These static experts provide high-fidelity reconstruction in regions with little or no motion, such as backgrounds or rigid structures, where dynamic deformation modeling is unnecessary or even detrimental.

\subsubsection{Volume-Aware Pixel Router}
In MoE-GS, expert models are trained independently and do not share a canonical Gaussian representation, which precludes establishing direct Gaussian-level correspondence across experts.
As a result, expert outputs cannot be composed through deformation-level or
Gaussian-level aggregation.
The most natural domain for expert interaction is therefore the image space, where all models produce rendered observations.

A straightforward solution is to predict routing weights from pixel-level features
using an MLP conditioned on time and viewing direction (Fig.~\ref{Router_detail}, top-left).
However, purely pixel-based routing lacks volumetric awareness, as it ignores
intrinsic Gaussian attributes such as position, scale, rotation, and opacity.
Conversely, assigning gating weights directly to 3D Gaussians before rasterization
preserves volumetric structure but remains ill-posed in the absence of shared
Gaussian correspondence across experts (Fig.~\ref{Router_detail}, bottom-left).
In this case, routing decisions must be inferred solely from rendered observations,
making expert competition indirect and difficult to stabilize. To address these limitations, we propose the \emph{Volume-aware Pixel Router} (Fig.~\ref{Router_detail}, right),
which combines image-space supervision with Gaussian-level structural cues.
Our design introduces temporally and view-dependent per-Gaussian routing
parameters that are splatted to the image plane through differentiable
Gaussian rasterization.

Specifically, for each Gaussian $\mathcal{G}_i$, we introduce learnable
per-Gaussian routing weights
\begin{equation}
\boldsymbol{w}_i^{per} = [w_i, w_i^{dir}, (t \cdot w_i^{time})]^T,
\end{equation}
which encode base, view-dependent, and time-dependent components.
These weights are treated as Gaussian attributes and rasterized to the image plane,
producing pixel-aligned embeddings that aggregate contributions from overlapping
Gaussians.

The resulting 2D routing features are refined by a lightweight MLP $\Phi$:
\begin{equation}
R'(u) = w_{2D}(u) +
\Phi\big(w_{2D}^{dir}(u),\, w_{2D}^{time}(u),\, r(u)\big),
\end{equation}
where $r(u)$ denotes the pixel viewing direction.
Here, $w_{2D}(u)$ provides a base routing signal aggregated from overlapping
Gaussians, while the MLP $\Phi$ injects additional view-dependent and
temporal context to adapt routing decisions to dynamic scene changes.
This residual formulation preserves Gaussian-level structure while allowing
flexible, pixel-adaptive refinement.

Expert gating weights are then obtained via softmax normalization:
\begin{equation}
G'_k(u) = \operatorname{Softmax}(R'_k(u)),
\end{equation}
which ensures that expert contributions at each pixel form a normalized
convex combination.
This competition mechanism allows the router to selectively emphasize
experts that better explain the observed appearance at $(u,t)$.

Finally, the routed image is synthesized by blending expert-rendered outputs:
\begin{equation}
I_{\text{MoE}}(u) =
\sum_{k=1}^{N} G'_k(u)\, I_{E_k}(u).
\end{equation}
Through this formulation, expert interaction occurs in image space,
while routing decisions remain informed by underlying Gaussian attributes,
enabling spatially coherent and temporally adaptive expert composition.

We optimize Router parameters ($\boldsymbol{w}_i^{per}$ and $\Phi$) using standard Gaussian Splatting loss terms (L1 and SSIM). To account for differences in expert convergence behavior, we employ a two-stage training strategy: experts are first optimized independently, and then the router is trained with these fixed experts (details in Appendix~\ref{sup_two-stage}).

\subsubsection{Practical Efficiency Design}
While MoE-GS enables flexible composition of heterogeneous dynamic Gaussian models,
it introduces two practical challenges:
(i) the lack of an explicit unified 3D representation due to image-space expert interaction,
and (ii) increased computational overhead arising from multiple experts and redundant rasterization.
In the following, we describe three complementary mechanisms that address these challenges
and make MoE-GS practical for large-scale dynamic reconstruction.

(i) \textit{Gaussian-Level Interpretation of Pixel Gating.}
\label{gaussian-level}
Although MoE-GS performs expert gating at the pixel level,
the routing weights are computed from per-Gaussian signals
(e.g., depth, visibility, and deformation) prior to rasterization.
Therefore, the learned gating implicitly reflects geometric structure
rather than relying solely on image-space cues.
Since the routing weights are defined before projection,
they can be lifted back to the Gaussian domain via
responsibility-weighted aggregation.
This enables post-hoc fusion and geometric analysis directly in 3D space.
Details are provided in Appendix~\ref{MoE_3D_interp}.

(ii) \textit{Single-Pass Multi-Expert Rendering and Gate-Aware Pruning.}
\label{app:render_prune}
In a naive multi-expert pipeline, each expert is rasterized independently,
leading to redundant projection and visibility computation.
We eliminate this redundancy by aggregating all Gaussians from different experts into a single batch and associating each Gaussian with a one-hot expert indicator $e_j \in \mathbb{R}^K$.
The rendered color for expert $k$ at pixel $u$ is then computed as Multi-Expert Rendering.
\begin{equation}
C_k(u) = \sum_{j=1}^{M} T_j(u)\,\alpha_j(u)\,c_j \cdot (e_j)_k,
\end{equation}
where $M$ is the total number of Gaussians across experts, $T_j(u) = \prod_{m=1}^{j-1}(1-\alpha_m(u))$ is the transmittance, 
$\alpha_j$ the opacity, and $(e_j)_k$ selects Gaussians of expert $k$. 
This design computes projection and visibility only once for all Gaussians, while expert-specific outputs are separated during alpha blending. 
As a result, redundant kernel launches and memory traversals inherent in the multi-pass pipeline are eliminated, 
improving GPU utilization without altering the rendering formulation.

To further reduce overhead, we introduce gate-aware Gaussian pruning.
Instead of heuristic ratio-based removal, we measure the sensitivity of routing
weights with respect to per-Gaussian routing parameters and compute an importance
score for each Gaussian across training views.
Specifically, we measure how strongly each Gaussian influences expert selection by accumulating the gradient of the routing weights $G'_k$
with respect to the corresponding per-Gaussian routing parameters
$\boldsymbol{w}_i^{per}$.
This gradient-based signal directly reflects the sensitivity of the MoE output to each Gaussian.
The importance of Gaussian $i$ across all training views $\mathcal{D}$ is computed as
\begin{equation}
\mathcal{E}_i = \frac{1}{|\mathcal{D}|} \sum_{v \in \mathcal{D}} \Bigl\| \frac{\partial G'_k(v)}{\partial \boldsymbol{w}_i^{per}(v)} \Bigr\|.
\end{equation}
Gaussians with $\mathcal{E}_i < \tau$ are progressively pruned, yielding a compact yet faithful representation. 
This strategy is effective because Gaussians with negligible gradients consistently show little impact on the gating weights and thus contribute minimally to the final image. 
By removing only these Gaussians, the model reduces rendering cost without sacrificing visual fidelity, 
unlike naive ratio-based pruning which destabilizes optimization.

(iii) \textit{Distillation-Based Expert Training.}
As the number of experts increases, pruning alone becomes insufficient to
maintain efficient inference.
We therefore introduce a distillation strategy~\cite{hinton2015distilling,xie2024mode,bucilua2006model,ba2014deep} that transfers the behavior of
the full MoE-GS model to individual experts.
Each expert $E_k$ is trained from scratch under dual supervision from both ground-truth images and the outputs of the MoE model.
Specifically, the MoE-rendered image $I_{\text{MoE}}$ serves as pseudo supervision, while the corresponding routing weights $G'_k$ provide confidence estimates for each expert. 
The distillation loss is
\begin{equation}
\begin{aligned}
\mathcal{L}_k^{\text{KD}}
&= \lambda\, \mathcal{L}\!\left(G'_k I_{E_k},\, G'_k I_{GT}\right) \\
&\quad + (1-\lambda)\, \mathcal{L}\!\left((1-G'_k) I_{E_k},\, (1-G'_k) I_{\text{MoE}}\right).
\end{aligned}
\end{equation}

where $\mathcal{L}$ combines L1 and SSIM losses, and $\lambda$ balances ground-truth vs. MoE supervision. 
This encourages each expert to specialize in reliable regions guided by ground truth while leveraging MoE outputs in uncertain areas. 
As a result, individual experts approximate the performance of the full MoE-GS model with significantly reduced complexity, enabling efficient real-time deployment. 

\begin{figure*}[tb]
\begin{center}
\centerline{\includegraphics[width=\textwidth]{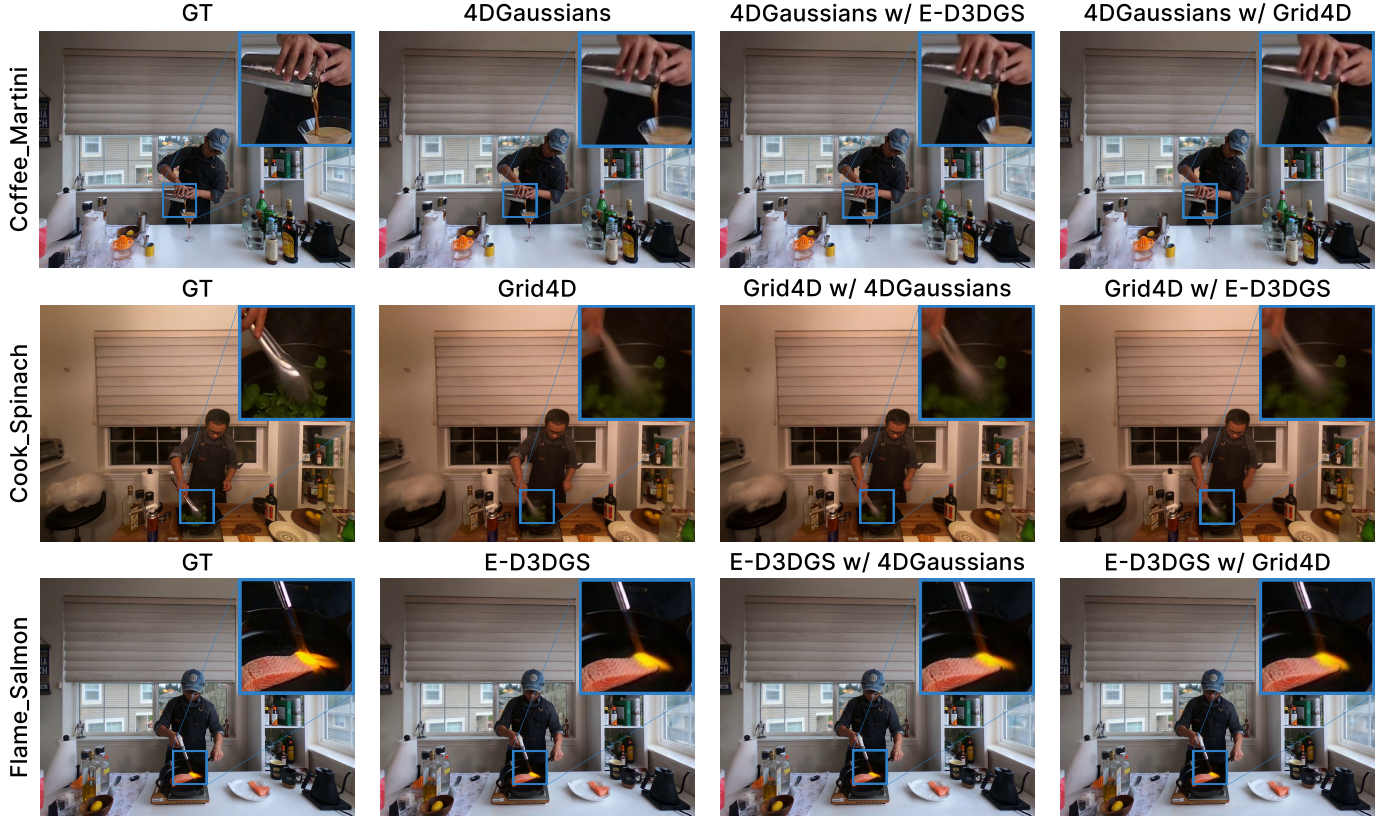}}
\caption{\textbf{MoDE Qualitative Results} Comparison of MoDE with baseline dynamic Gaussian splatting methods on Neural 3D Video dataset~\cite{li2022neural}.}
\label{mode_quali}
\end{center}
\vspace{-8mm}
\end{figure*}

\section{Experiments}
In this section, we evaluate MoDE and MoE-GS from the perspectives of
reconstruction quality, efficiency, and training stability.
Sec.~\ref{sec:Experimental Setup} describes the experimental setup,
including datasets, evaluation metrics, and implementation details.
Sec.~\ref{sec:MoDE exp} presents quantitative and qualitative results
for MoDE, analyzing the effect of deformation-level expert composition
under a shared canonical representation.
Sec.~\ref{sec:MoE-GS exp} reports results for MoE-GS, including
performance across different expert configurations and ablation studies
on routing design, efficiency mechanisms, and training stability.
Finally, Sec.~\ref{sec:mode_moegs_comparison} provides a direct
comparison between MoDE and MoE-GS from both structural and
efficiency--performance perspectives, highlighting the trade-offs
induced by their different integration constraints.

\vspace{-3mm}
\subsection{Experimental Setup}
\vspace{1mm}
\label{sec:Experimental Setup}
We conduct experiments on two standard benchmarks for dynamic scene reconstruction, Neural 3D Video (N3V)~\cite{li2022neural} and Technicolor~\cite{sabater2017dataset}, and report PSNR, SSIM~\cite{wang2004image}, and LPIPS~\cite{zhang2018unreasonable} to evaluate visual quality.
Both MoDE and MoE-GS are implemented within the same Gaussian Splatting framework and share identical training settings unless otherwise specified. 
All experiments are conducted on NVIDIA A6000 GPUs. 
Additional dataset-specific configurations are provided in the supplementary material.

Due to structural differences between MoDE and MoE-GS, their baseline evaluation protocols differ. 
MoDE jointly optimizes multiple deformation experts within a shared canonical Gaussian representation, which prevents direct reuse of pretrained single-deformation models. 
Therefore, for MoDE, all baseline models are retrained under the same configuration, and the retrained single-deformation results are used as reference baselines. 
In contrast, MoE-GS adopts a decoupled optimization strategy that allows independently trained experts to be directly utilized. 
Accordingly, publicly available pretrained models are used without additional retraining, and their original reported performance serves as the baseline reference.

\renewcommand{\arraystretch}{1.0}

\vspace{-1mm}

\subsection{Results on MoDE}
\label{sec:MoDE exp}
\vspace{2mm}

\subsubsection{Baselines and Expert Configurations}
We evaluate MoDE using three canonical deformation-based
dynamic Gaussian models: 4DGaussians~\cite{wu20244d},
Grid4D~\cite{xu2024grid4d}, and E-D3DGS~\cite{bae2024per}.
We use each method in its original single-deformation configuration as a baseline reference.
MoDE variants are constructed by augmenting each baseline with one additional canonical deformation expert operating on the same set of canonical Gaussians.
All deformation experts are trained jointly within a single
optimization process, sharing the canonical Gaussian
representation and combining their outputs through
time-dependent gating.

\subsubsection{Qualitative Evaluation}
Fig.~\ref{mode_quali} presents qualitative comparisons between single-deformation baselines and their MoDE variants. Compared to the corresponding single-deformation models, MoDE produces more expressive motion patterns in dynamic regions, particularly around articulated and rapidly moving structures. These improvements arise from representation-level integration, where multiple deformation formulations interact within a shared canonical representation and capture complementary motion characteristics. As a result, MoDE yields qualitatively distinct dynamic behaviors and temporal appearance patterns.

\begin{table*}[t!]
\small
\caption{Results of Mixture of Deformation Experts (MoDE) on the N3V dataset~\cite{li2022neural}. 
MoDE composes multiple deformation experts within a single Gaussian representation.}
\vspace{-1mm}
\centering
\resizebox{0.9\textwidth}{!}{%
\begin{tabular}{l|cccccc|c}
\toprule
\multirow{2}{*}{Model} & \multicolumn{7}{c}{PSNR (dB) $\uparrow$} \\ \cmidrule(lr){2-8} 
 & \tworow{Coffee}{~Martini~} 
 & \tworow{Cook}{~Spinach~} 
 & \tworow{\!\!\!\!\!Cut\,Roasted\!\!\!\!\!}{Beef} 
 & \tworow{Flame}{~Salmon~} 
 & \tworow{Flame}{~Steak~} 
 & \tworow{Sear}{~Steak~} 
 & Average \\ 
\midrule
4DGaussians~\cite{wu20244d} 
& 28.42 & 32.24 & \textbf{30.71} & 29.32 & 31.19 & 31.80 & 30.61 \\
\rowcolor{lightgray}
4DGaussians~\cite{wu20244d} w/ E-D3DGS~\cite{bae2024per}
& \textbf{28.90} & 32.41 & 30.22 & \textbf{29.90} & 31.99 & \textbf{32.42} & \textbf{30.97} \\
\rowcolor{lightgray}
4DGaussians~\cite{wu20244d} w/ Grid4D~\cite{xu2024grid4d}
& 28.65 & \textbf{32.42} & 30.50 & 29.37 & \textbf{32.08} & 31.97 & 30.83 \\

\midrule
Grid4D~\cite{xu2024grid4d} 
& 28.15 & 32.39 & 31.82 & 28.60 & 33.05 & \textbf{33.30} & 31.22 \\
\rowcolor{lightgray}
Grid4D~\cite{xu2024grid4d} w/ 4DGaussians~\cite{wu20244d} 
& 28.97 & \textbf{32.70} & 32.45 & 29.07 & \textbf{33.06} & 33.22 & 31.58 \\
\rowcolor{lightgray}
Grid4D~\cite{xu2024grid4d} w/ E-D3DGS~\cite{bae2024per}
& \textbf{29.04} & 32.57 & \textbf{33.15} & \textbf{29.16} & 32.90 & 32.94 & \textbf{31.63} \\

\midrule
E-D3DGS~\cite{bae2024per} 
& 28.92 & \textbf{32.45} & 32.47 & 30.09 & 31.99 & \textbf{33.37} & \textbf{31.55} \\
\rowcolor{lightgray}
E-D3DGS~\cite{bae2024per} w/ 4DGaussians~\cite{wu20244d} 
& 28.90 & 31.28 & \textbf{33.06} & \textbf{30.25} & \textbf{32.61} & 32.69 & 31.46 \\
\rowcolor{lightgray}
E-D3DGS~\cite{bae2024per} w/ Grid4D~\cite{xu2024grid4d} 
& \textbf{28.98} & 32.04 & 30.89 & 30.12 & 32.35 & 33.10 & 31.25 \\

\bottomrule
\end{tabular}
}
\label{tab:mode_n3v}
\vspace{-2mm}
\end{table*}

\begin{figure}[t]
\centering
\includegraphics[width=\columnwidth]{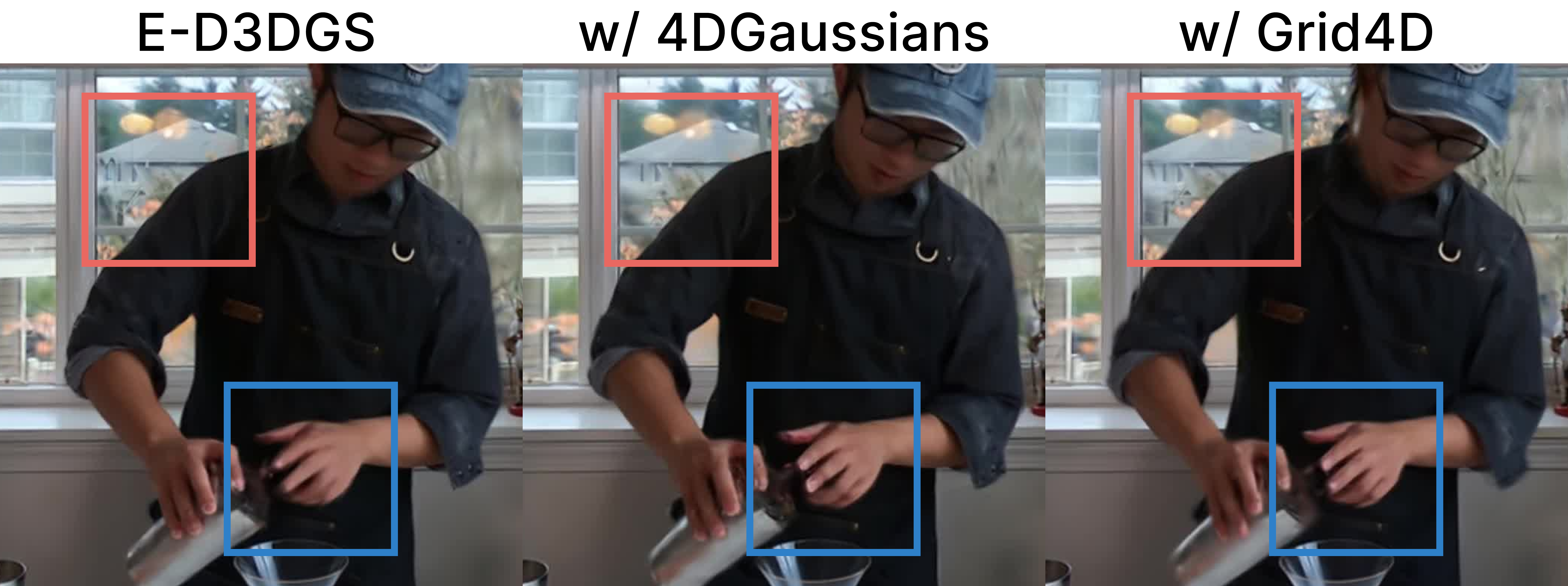}
\caption{\textbf{Trade-off analysis of the E-D3DGS-based MoDE variant.} Dynamic-region gains (blue) and static-region degradation (red).}
\label{fig:mode_abl}
\vspace{-2mm}

\end{figure}

\subsubsection{Quantitative Evaluation}

Quantitative evaluation is conducted on the N3V dataset~\cite{li2022neural}. As reported in Table~\ref{tab:mode_n3v}, MoDE improves the average PSNR over the corresponding single-deformation baselines for 4DGaussians~\cite{wu20244d} and Grid4D~\cite{xu2024grid4d}. As discussed in the qualitative analysis, these gains are accompanied by enhanced motion expressiveness in dynamic regions through the integration of complementary canonical deformation experts within a shared representation. This effect is visible in Fig.~\ref{mode_quali}, where the MoDE variants produce more expressive reconstructions around articulated or rapidly moving regions compared with the corresponding single-deformation baselines.

However, when E-D3DGS~\cite{bae2024per} is used as the baseline, MoDE does not consistently improve the full-image PSNR and can even reduce the averaged score. This is because E-D3DGS already provides strong locally coherent deformation modeling, and jointly optimizing additional deformation experts over the same canonical Gaussian representation introduces a more constrained optimization problem. As shown in Fig.~\ref{fig:mode_abl}, the E-D3DGS-based MoDE variants still improve dynamic regions, such as the moving hand, indicating increased motion expressiveness. At the same time, they can degrade static regions, such as the background visible through the window. Since the N3V scenes contain large static areas and PSNR is computed over the full image, the degradation in static regions can offset or outweigh the gains in dynamic regions. Therefore, the practical benefit of MoDE should be understood as improving dynamic-region motion expressiveness through representation-level deformation composition, while its full-image quantitative gain depends on the strength of the underlying baseline and the static--dynamic trade-off induced by shared canonical optimization.

\vspace{-1mm}
\subsection{Results on MoE-GS}
\label{sec:MoE-GS exp}
\vspace{2mm}

\subsubsection{Baselines and Expert Configurations}

For MoE-GS, we adopt independently trained expert models
and treat them as fixed during router training.
Publicly available pretrained models are used when available,
and otherwise reproduced following their original training protocols.
The expert pool includes representative dynamic Gaussian Splatting
methods with distinct motion parameterizations, namely
4DGaussians~\cite{wu20244d},
E-D3DGS~\cite{bae2024per},
Ex4DGS~\cite{lee2024fully}, and
STG~\cite{li2024spacetime}.
We additionally report representative NeRF-based baselines
for comparison.
Unless otherwise specified, we evaluate MoE-GS with
$N\!=\!2,3,4$ experts using fixed combinations:
$N\!=\!2$: \{Ex4DGS, STG\};
$N\!=\!3$: + E-D3DGS;
$N\!=\!4$: + 4DGaussians.
These configurations are used consistently across all experiments.

\begin{table*}[t!]
\small
\caption{Performance comparison on the N3V dataset~\cite{li2022neural}. {\small\textdagger}: Models were trained on a dataset split into 150 frames. $N\!=\!2,3,4$ experts using
the following fixed combinations:
$N\!=\!2$: \{Ex4DGS~\cite{lee2024fully}, STG~\cite{li2024spacetime}\};
$N\!=\!3$: + E-D3DGS~\cite{bae2024per};
$N\!=\!4$: + 4DGaussians~\cite{wu20244d}.}
\vspace{-1mm}
\centering
\resizebox{0.75\textwidth}{!}{%
\begin{tabular}{l|cccccc|c}
\toprule
\multirow{2}{*}{Model} & \multicolumn{7}{c}{PSNR (dB) $\uparrow$} \\ \cmidrule(lr){2-8} 
 & \tworow{Coffee}{~Martini~} & \tworow{Cook}{~Spinach~} & 
 \tworow{\!\!\!\!\!Cut\,Roasted\!\!\!\!\!}{Beef} & \tworow{Flame}{~Salmon~} & \tworow{Flame}{~Steak~} & \tworow{Sear}{~Steak~} & Average \\ \midrule
HyperReel~\cite{attal2023hyperreel}        & 28.37 & 32.30 & 32.92 & 28.26 & 32.20 & 32.57 & 31.10\\
K-Planes~\cite{fridovich2023k}                          & 29.99 & 32.60 & 31.82 & 30.44 & 32.38 & 32.52 & 31.63 \\ 
MixVoxels-L~\cite{wang2023mixed}                        & 29.63 & 32.25 & 32.40 & 29.81 & 31.83 & 32.10 & 31.34  \\ 
\cmidrule(lr){1-8}

3DGStream~\cite{sun20243dgstream}                                  & 27.75   & 33.31 & 33.21 & 28.42   & 34.30   & 33.01 & 31.67  
\\ 
DASS~\cite{liu2026dynamics}                                  & 28.15   & 33.83 & 33.54 & 28.84   & 34.26   & 33.33 & 31.99  
\\ 

SaRO-GS~\cite{yan20244d}                                & 28.96 & 33.19 & 33.91 & 29.14 & 33.83 & 33.89 & 32.15 \\
SwinGS~\cite{liu2024swings}                                & 27.99 & 33.66 & 34.03 & 28.24 & 32.94 & 33.32 & 31.69  \\

4DGaussians~\cite{wu20244d}                    & 29.09 & 32.78 & 33.15 & 29.76 & 31.81 & 32.01 & 31.43 \\ 

E-D3DGS~\cite{bae2024per} & 30.04 & 33.11 & 33.85 & 30.49 & 32.77 & 33.70 & 32.33 \\

STG\textsuperscript{\textdagger}~\cite{li2024spacetime} & 28.16 & 33.09 & 34.15 & 29.09 & 33.25 & 33.77 & 31.92 \\

Ex4DGS~\cite{lee2024fully}                                & 28.72 & 33.24 & 33.73 & 29.33 & 33.91 & 33.69 & 32.10 \\
\addlinespace[2pt]
\rowcolor{lightgray}
\textbf{MoE-GS (N=2)}                    & 29.39 & 33.87 & 34.65 & 29.88 & 34.59 & 34.58 & 32.82 \\
\addlinespace[1.5pt]
\rowcolor{lightgray}
\addlinespace[1.5pt]
\rowcolor{lightgray}
\textbf{MoE-GS (N=3)  }                  & 30.27 & 33.86 & 34.90 & 30.92 & 34.52 & 34.88 & 33.23 \\
\addlinespace[1.5pt]
\rowcolor{lightgray}
\textbf{MoE-GS (N=4)}                    & \textbf{30.43} & \textbf{34.24} & \textbf{35.20} & \textbf{30.92} & \textbf{34.38} & \textbf{34.42} & \textbf{33.27} \\
\bottomrule
\end{tabular}
}
\vspace{-2mm}
\label{tab:N3V}
\end{table*}

\begin{table}[t!]
\small
\caption{Comparison results on\,the\,Technicolor\,dataset \cite{sabater2017dataset}. $N\!=\!3$: \{Ex4DGS~\cite{lee2024fully}, STG~\cite{li2024spacetime}, E-D3DGS~\cite{bae2024per}\}.}
\vspace{-2mm}
\centering
\resizebox{0.48\textwidth}{!}{%
\begin{tabular}{l|ccccc|c}
\toprule
\multirow{2}{*}{Model} & \multicolumn{6}{c}{PSNR (dB) $\uparrow$}  \\ \cmidrule(lr){2-7}  
 & Birthday & Fabien & Painter & Theater & Train & Average \\
\midrule
 
DyNeRF~\cite{li2022neural}                              & 29.20 & 32.76 & 35.95 & 29.53 & 31.58 & 31.80 \\ 

HyperReel~\cite{attal2023hyperreel}        & 29.99 & 34.70 & 35.91 & \textbf{33.32} & 29.74 & 32.73 \\

4DGaussians~\cite{wu20244d}                    & 30.87 & 33.56 & 34.36 & 29.81 & 25.35 & 30.79  \\
STG~\cite{li2024spacetime}                                & 32.16 & 35.70 & 37.18 & 31.00 & 32.39 & 33.69  \\

E-D3DGS~\cite{bae2024per} & 32.38 & 34.24 & 36.20 & 31.10 & 31.37 & 33.06  \\

Ex4DGS~\cite{lee2024fully}                                & 32.35 & 35.18 & 36.60 & 31.77 & 31.37 & 33.45  \\

\rowcolor{lightgray}
\textbf{MoE-GS (N=3) }                   & \textbf{33.26} & \textbf{36.26} & \textbf{37.63} & 32.88 & \textbf{32.89} & \textbf{34.55}  \\
\bottomrule
\end{tabular}%
}
\vspace{-2mm}
\label{tab:technicolor}
\end{table}

\begin{table}[t]

\centering
\caption{Efficiency evaluation with N=2 Expert Variants on N3V dataset~\cite{li2022neural}. $N\!=\!2$: \{Ex4DGS~\cite{lee2024fully}, STG~\cite{li2024spacetime}\}.}
\label{tab:n23-tfm}
\resizebox{\linewidth}{!}{%
\begin{tabular}{l|ccc}
\toprule
Model & PSNR $\uparrow$ & FPS $\uparrow$ & Memory $\downarrow$ \\
\midrule
STG\textsuperscript{\textdagger}~\cite{li2024spacetime}   & 31.92 & 88.5  & 609.5 \\
Ex4DGS~\cite{lee2024fully}        & 32.01 & 120  & 122.8 \\
\addlinespace[1.5pt]
\rowcolor{lightgray}
MoE\text{-}GS (N=2)        & 32.82  & 44  & 878.7 \\
\addlinespace[1.5pt]
\rowcolor{lightgray}
MoE\text{-}GS (55\% pruning)        & 32.80  & 83  & 351.2 \\
\addlinespace[1.5pt]
\rowcolor{lightgray}
MoE\text{-}GS (75\% pruning)        & 32.45  & 101  & 281.3 \\
\bottomrule
\end{tabular}
}
\vspace{-2mm}
\end{table}

\subsubsection{Quantitative Evaluation} 
\label{Quantitative}
Tables~\ref{tab:N3V} and~\ref{tab:technicolor} report per-scene results on the
N3V~\cite{li2022neural} and Technicolor~\cite{sabater2017dataset} datasets.
Across both benchmarks, MoE-GS achieves the highest average performance and
outperforms most baseline methods on individual scenes, demonstrating consistent
reconstruction quality across diverse dynamic scenarios.
Because MoE-GS combines multiple independently trained experts through an
additional routing stage, it incurs higher inference and memory cost compared to
single-expert baselines.
To analyze this overhead, Table~\ref{tab:n23-tfm} reports an efficiency study of
MoE-GS along with the effect of Gate-Aware Pruning.
The results show that pruning substantially reduces inference time and memory
usage while preserving reconstruction quality.
In particular, the $N\!=\!2$ configuration, which combines STG~\cite{li2024spacetime}
and Ex4DGS~\cite{lee2024fully}, outperforms both individual experts in PSNR while
maintaining moderate computational overhead.
To further validate the generality of MoE-GS, we provide additional evaluations on the monocular HyperNeRF dataset~\cite{park2021hypernerf} in Appendix~\ref{sup_additonal_quanti}, as well as on the large-motion PanopticSports and D-NeRF benchmarks in Appendix~\ref{large_motion}.

\begin{figure*}[tb]
\begin{center}
\centerline{\includegraphics[width=\textwidth]{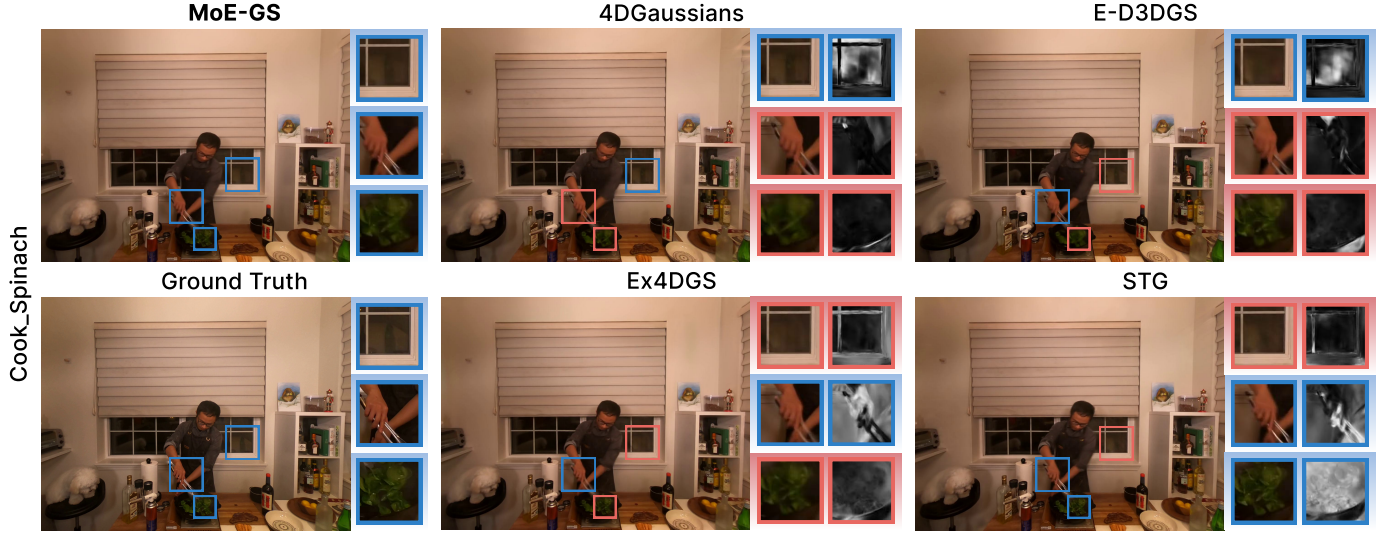}}
\caption{\textbf{N3V Qualitative Results} Comparison of our MoE-GS with other dynamic Gaussian splatting methods on Neural 3D Video dataset~\cite{li2022neural}. \colorbox[rgb]{0.337, 0.671, 0.961}{Blue backgrounds} highlight the method that produces the most visually accurate result among the baselines for each region.}
\label{n3v_qualitative}
\end{center}
\vspace{-8mm}
\end{figure*}

\subsubsection{Qualitative Evaluation} 
\label{Qualitative}
Figure~\ref{n3v_qualitative} visualizes the routing weights predicted by the router,
the outputs of individual experts, and the final MoE-GS renderings.
The spatial distribution of routing weights is consistent with the visual
characteristics of the experts’ outputs, indicating that MoE-GS combines experts
in a spatially adaptive manner.
Additional qualitative results are provided in Appendix~\ref{sup_additonal_quali}.

\begin{figure}[t]
\begin{center}
\centerline{\includegraphics[width=\columnwidth]{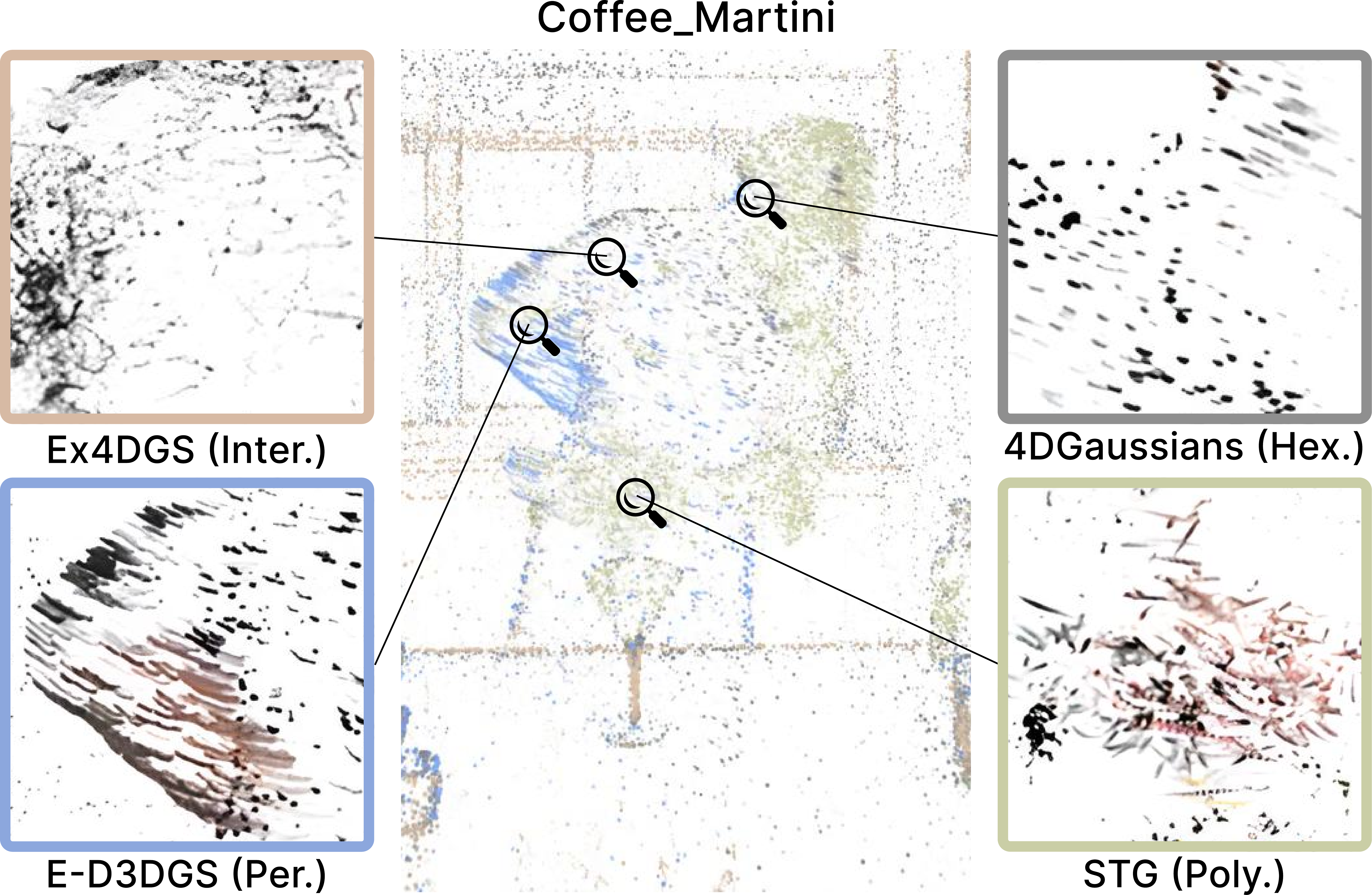}}
\caption{\textbf{Expert-specific motion patterns.} Representative motion trajectories produced by different dynamic Gaussian Splatting experts.}
\vspace{-3pt}
\label{motion_pattern}
\vspace{-10pt}
\end{center}
\end{figure}

\vspace{2mm}
\subsubsection{Expert Specialization Analysis}
To better understand the behavior of the proposed MoE framework, we analyze the characteristic motion patterns captured by each expert. As illustrated in Fig.~\ref{motion_pattern}, different deformation formulations produce qualitatively distinct motion trajectories even when applied to the same underlying Gaussian primitives. We further analyze the resulting expert-specific motion characteristics below.

\noindent\textbf{4DGaussians~\cite{wu20244d} (HexPlane canonical deformation).}
4DGaussians tends to favor smooth and spatially coherent low-frequency motion because neighboring Gaussians share a common HexPlane feature representation, resulting in highly correlated deformation signals.

\noindent\textbf{E-D3DGS~\cite{bae2024per} (Per-Gaussian embedding deformation).}
E-D3DGS is well suited for locally coherent high-velocity motion. This tendency comes from its two-branch deformation network, which separately captures coarse and fine motion components, while per-Gaussian embeddings promote spatially coherent motion patterns among neighboring Gaussians.

\noindent\textbf{Ex4DGS~\cite{lee2024fully} (Keyframe interpolation).}
Ex4DGS is effective for irregular and free-form motion because each Gaussian is independently interpolated between keyframes without enforcing a shared deformation field, allowing highly diverse local trajectories.

\noindent\textbf{STG~\cite{li2024spacetime} (Polynomial trajectory model).}
STG is particularly suitable for globally smooth and near-rigid motion due to its low-order polynomial trajectory parameterization, which imposes a strong bias toward smooth and directionally consistent motion.

These observations suggest that different deformation formulations induce distinct motion priors, which in turn lead to different expert specialization behaviors in MoE-GS. This complementary behavior motivates the expert composition strategy adopted in MoE-GS. Furthermore, additional analyses, representative examples, and trajectory visualizations are provided in Appendix~\ref{expert_analysis}.}

\vspace{2mm}
\subsubsection{Ablation Studies on MoE-GS.} 
We conduct ablation studies to analyze the core design choices of MoE-GS.
Given that MoE-GS differs from MoDE in its decoupled expert optimization
and rendering-level composition, we focus on three aspects:
(i) routing architecture,
(ii) efficiency design,
(iii) training stability, and 
(iv) expert candidate ablation.

(i) \textit{MoE-GS Router Variants.}
\label{Qualitative_router}
We compare three routing strategies:
\emph{Pixel Router}, \emph{Volume Router}, and the proposed
\emph{Volume-aware Pixel Router}.
As reported in Table~\ref{tab:MoE_arch},
the Pixel Router performs expert blending purely in image space.
While this leads to stable optimization,
it yields inferior reconstruction quality,
particularly near structural boundaries.
The Volume Router applies gating directly in 3D
by modulating Gaussian opacities.
Although this preserves geometric structure more explicitly,
it often suffers from optimization instability and oversmoothing artifacts.
In contrast, the proposed Volume-aware Pixel Router
consistently achieves superior performance across all metrics.
These results indicate that effective expert composition
requires both image-space optimization stability
and geometry-aware conditioning.

\begin{table}
\centering
\caption{Performance Comparison of Different MoE Router Variants}
\label{tab:MoE_arch}
\resizebox{\linewidth}{!}{%
\begin{tabular}{l|ccc}
\toprule
Model & PSNR $\uparrow$ & SSIM $\uparrow$ & LPIPS $\downarrow$ \\
\midrule
Pixel Router & 31.12 & 0.952 & 0.022 \\
Volume Router & 32.05 & 0.951 & 0.022 \\
\rowcolor{lightgray}
Volume-aware Pixel Router & \textbf{33.23} & \textbf{0.954} & \textbf{0.021} \\
\bottomrule
\end{tabular}
}
\vspace{-1mm}
\end{table}

\begin{table}[t]

\vspace{-1mm}
\centering
\caption{Ablation of Efficiency Optimizations. $N\!=\!3$: \{Ex4DGS~\cite{lee2024fully}, STG~\cite{li2024spacetime}, E-D3DGS~\cite{bae2024per}\}.}
\label{tab:abl_render_prune}
\resizebox{\linewidth}{!}{%
\begin{tabular}{l|ccc}
\toprule
Model & PSNR $\uparrow$ & FPS $\uparrow$ & Memory $\downarrow$ \\
\midrule
w/o Single-Pass \& Pruning & 32.54 & 36 & 747 \\
w/o Single-Pass & 33.23 & 40 & \textbf{270} \\
w/o Pruning & 32.54 & 60 & 747 \\
\rowcolor{lightgray}
\textbf{MoE-GS (N=3)} & \textbf{33.23} & \textbf{68} & \textbf{270} \\
\bottomrule
\end{tabular}
}
\vspace{-1mm}
\end{table}

(ii) \textit{Efficiency Optimizations.}
Because MoE-GS integrates multiple independently trained experts,
rendering efficiency is a primary concern.
We evaluate two complementary strategies:
\emph{Single-Pass Multi-Expert Rendering}
and \emph{Gate-Aware Gaussian Pruning}.
As shown in Table~\ref{tab:abl_render_prune},
removing either component leads to substantial degradation
in rendering speed or memory usage.
The full MoE-GS configuration eliminates redundant rasterization
and prunes Gaussians with negligible routing influence,
achieving significant improvements in FPS and memory efficiency
while maintaining reconstruction quality.

We further evaluate distillation-based expert training
as a complementary efficiency strategy.
As reported in Table~\ref{tab:KD_MoE},
experts distilled under MoE-GS supervision
consistently outperform their original counterparts.
These results indicate that MoE-GS not only improves
reconstruction directly,
but also serves as an effective teacher
for training compact expert models. More detailed results and analyses are provided in Appendix~\ref{sup_weigth_ablation}.

\begin{table}[t]

\vspace{-1mm}
\centering
\caption{Quantitative comparison of MoE-GS distillation methods on the Technicolor dataset. $N\!=\!3$: \{Ex4DGS~\cite{lee2024fully}, STG~\cite{li2024spacetime}, E-D3DGS~\cite{bae2024per}\}.}
\label{tab:KD_MoE}
\vspace{-2mm}
\resizebox{0.95\linewidth}{!}{%
\begin{tabular}{l|ccc}
\toprule
Model & PSNR $\uparrow$ & SSIM $\uparrow$ & LPIPS $\downarrow$ \\
\midrule
E-D3DGS \cite{bae2024per} & 32.88 & 0.902 & 0.111 \\
\rowcolor{lightgray}
E-D3DGS \cite{bae2024per} (Distilled) & \textbf{33.67} & \textbf{0.915} & \textbf{0.091} \\
\midrule
STG \cite{li2024spacetime}                             & 32.83 & 0.915 & 0.083 \\
\rowcolor{lightgray}
STG \cite{li2024spacetime} (Distilled)                                          & \textbf{33.10} & \textbf{0.917} & \textbf{0.082} \\
\midrule
Ex4DGS \cite{lee2024fully}                                         & 33.57 & 0.918 & 0.086 \\
\rowcolor{lightgray}
Ex4DGS \cite{lee2024fully} (Distilled)                                         & \textbf{33.91} & \textbf{0.923} & \textbf{0.079} \\
\bottomrule
\end{tabular}
}
\vspace{-0.5mm}
\end{table}

\begin{table}[t]
\vspace{-1mm}
\centering
\caption{\textbf{Stability across repeated trainings.} 
Variability exists across runs, but MoE-GS consistently outperforms all single-expert variants. $N\!=\!3$: \{Ex4DGS~\cite{lee2024fully}, STG~\cite{li2024spacetime}, E-D3DGS~\cite{bae2024per}\}.}
\label{best_of_n}
\resizebox{0.8\linewidth}{!}{
\begin{tabular}{l|ccc}
\toprule
\multirow{2}{*}{Model} & \multicolumn{3}{c}{PSNR (dB)} \\
\cmidrule(lr){2-4}
 & Min & Max & Average \\
\midrule
E\mbox{-}D3DGS~\cite{bae2024per}      & 30.78 & 32.33 & 31.19 \\
Ex4DGS~\cite{lee2024fully}            & 31.43 & 32.10 & 31.78 \\
4DGaussians~\cite{wu20244d}           & 30.18 & 31.43 & 30.70 \\
\rowcolor{lightgray}
\textbf{MoE-GS (N=3)}                  & \textbf{32.72} & \textbf{33.23} & \textbf{33.01} \\
\bottomrule
\end{tabular}
}
\end{table}

(iii) \textit{Stability Across Repeated Trainings.}
Dynamic Gaussian Splatting methods
often exhibit noticeable run-to-run variance
due to optimization sensitivity.
As shown in Table~\ref{best_of_n},
single-expert baselines display significant performance variability,
whereas MoE-GS consistently achieves higher
and more stable reconstruction quality.
Notably, even the best individual run
of any single-expert baseline
does not match the average performance of MoE-GS.
This suggests that the gains of MoE-GS
are not attributable to favorable initialization,
but arise from its decoupled expert specialization mechanism.

\begin{table}[t]
\vspace{-1mm}
\centering
\caption{\textbf{Expert candidate ablation on the N3V dataset.}
The original $N\!=\!2$ configuration consists of \{Ex4DGS~\cite{lee2024fully}, STG~\cite{li2024spacetime}\}. 
$N\!=\!3$ variants are obtained by adding different third experts.}
\label{expert_pool_ablation}
\resizebox{0.68\linewidth}{!}{
\begin{tabular}{l|c}
\toprule
Configuration & PSNR (dB) \\
\midrule
MoE-GS (N=2) & 32.82 \\
\rowcolor{lightgray}
\textbf{+ E-D3DGS~\cite{bae2024per}} & \textbf{33.23} \\
+ Grid4D~\cite{xu2024grid4d} & 33.06 \\
+ DeformableGS~\cite{yang2024deformable} & 32.87 \\
+ 3DGS-MCMC~\cite{kheradmand20243d} & 33.07 \\
\bottomrule
\end{tabular}
}
\vspace{-1mm}
\end{table}

\begin{figure}[t]
\centering
\includegraphics[width=\columnwidth]{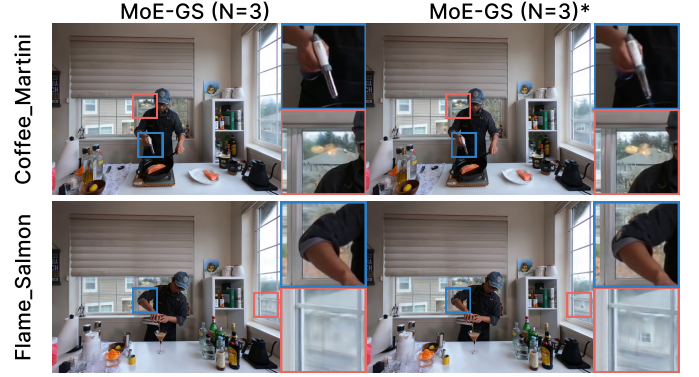}
\caption{\textbf{Effect of Adding a Static Expert.}
Qualitative comparison between two MoE-GS variants built upon the same two-expert baseline \{Ex4DGS~\cite{lee2024fully}, STG~\cite{li2024spacetime}\}.
MoE-GS (N=3): + E-D3DGS~\cite{bae2024per}.
MoE-GS (N=3)*: + 3DGS-MCMC~\cite{kheradmand20243d} (static expert). Red boxes indicate detail-rich static ROIs, while blue boxes indicate dynamic regions.}
\vspace{-2mm}
\label{fig:static_expert_qual}
\end{figure}

(iv) \textit{Expert Candidate Ablation.}
To evaluate MoE-GS with a broader set of Gaussian Splatting experts, we expand the original $N\!=\!2$ expert set, \{Ex4DGS~\cite{lee2024fully}, STG~\cite{li2024spacetime}\}, by adding different third experts. Table~\ref{expert_pool_ablation} reports the results of $N\!=\!3$ variants using E-D3DGS~\cite{bae2024per}, Grid4D~\cite{xu2024grid4d}, DeformableGS~\cite{yang2024deformable}, and 3DGS-MCMC~\cite{kheradmand20243d}.
Among the deformation-based variants, E-D3DGS achieves the highest reconstruction performance, followed by Grid4D and DeformableGS.
Nevertheless, all evaluated $N\!=\!3$ configurations improve over the original $N\!=\!2$ expert set, indicating that expert composition remains beneficial across different third-expert choices.

We further quantify the static regions highlighted in Fig.~\ref{fig:static_expert_qual} using the same fixed ROIs over all 300 frames. Since low-frequency static areas are reconstructed similarly by both variants, we focus on detail-rich static ROIs, such as window and background structures, where the benefit of a static expert is expected to be more visible. In these ROIs, MoE-GS ($N=3$)$^*$ improves PSNR/SSIM from 21.10/0.710 to 22.23/0.726 on Coffee\_Martini and from 23.53/0.794 to 23.90/0.813 on Flame\_Salmon. This indicates that the static expert is particularly beneficial for high-frequency static structures, while its limited contribution to dynamic regions explains why it still underperforms the E-D3DGS-based $N=3$ configuration in full-image PSNR.

\begin{figure}[t]
\centering
\includegraphics[width=\columnwidth]{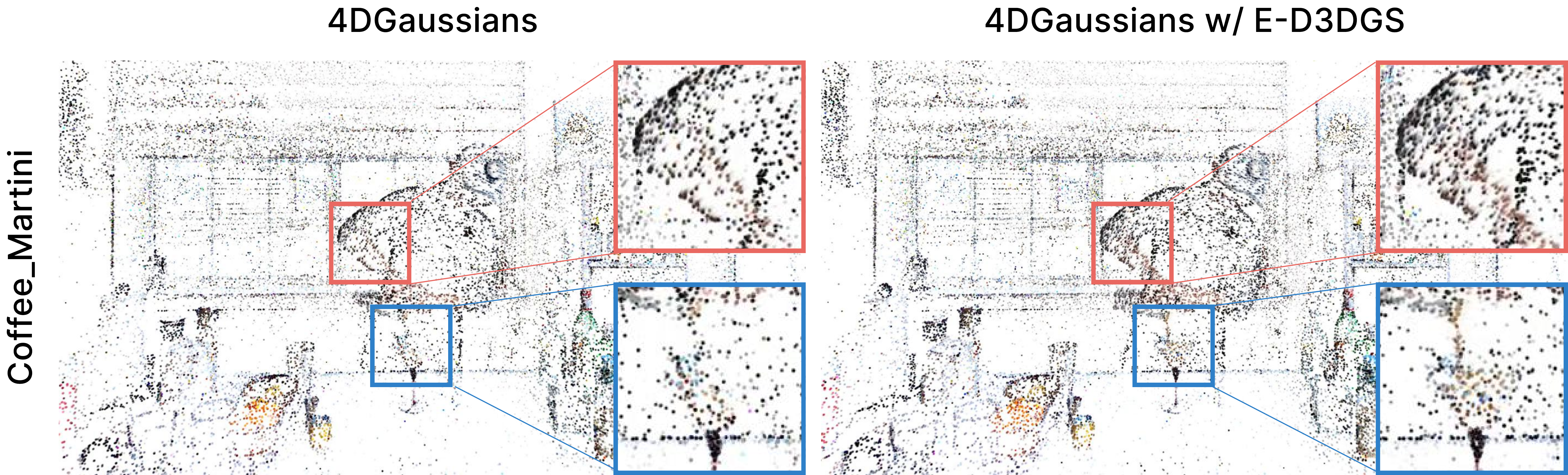}
\caption{\textbf{Deformation-Level Composition Improves Motion Representation.}
4DGaussians~\cite{wu20244d} shows rigid and spatially coherent motion due to its shared canonical deformation.
In contrast, MoDE increases per-Gaussian motion flexibility,
capturing finer dynamics such as the pouring coffee.
Zoomed regions highlight motion patterns not represented by the single-deformation baseline.
}
\label{fig:mode_motion}
\vspace{-2mm}
\end{figure}

\begin{figure*}[t]
\begin{center}
\centerline{\includegraphics[width=\textwidth]{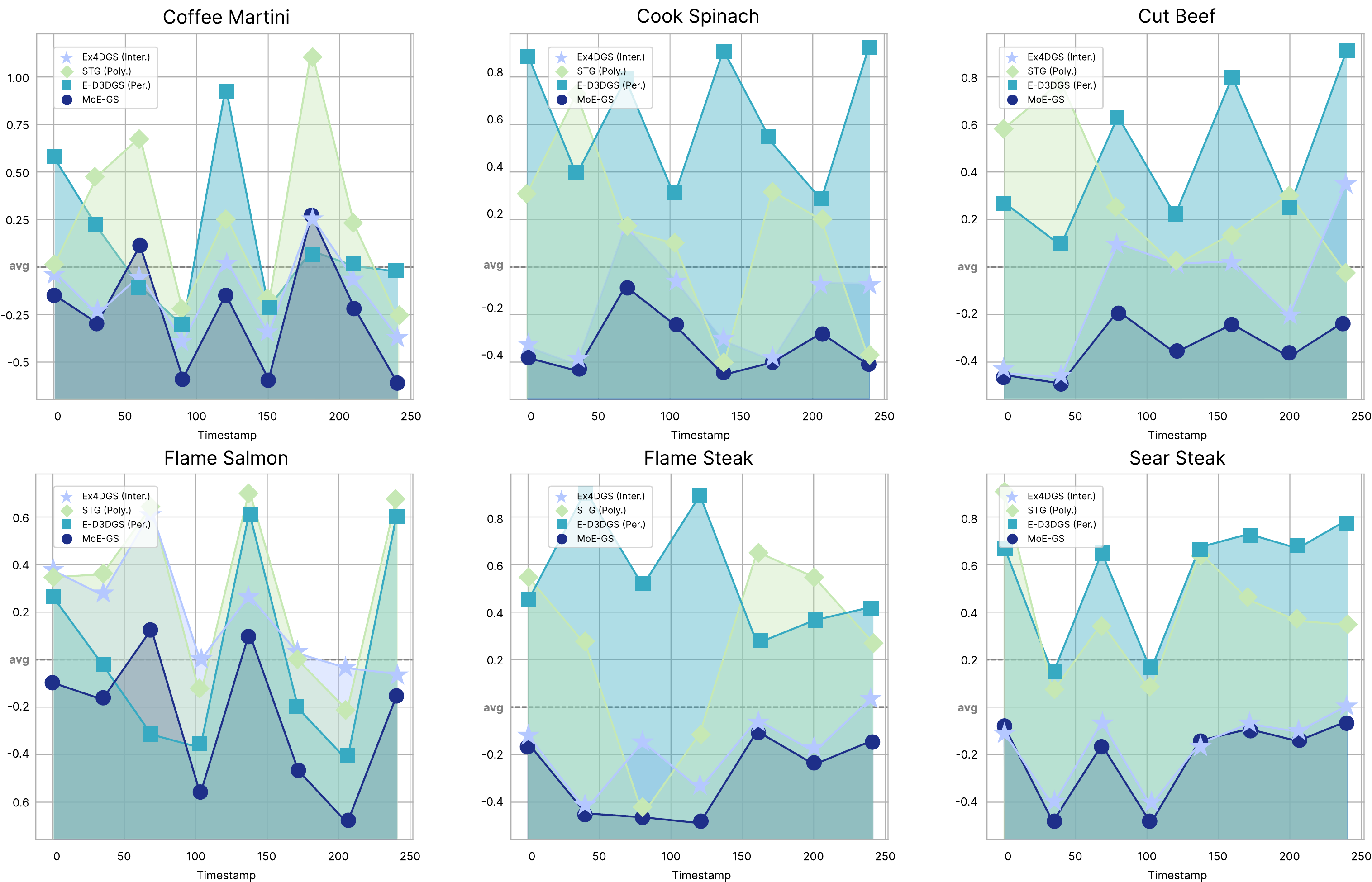}}
\caption{Multi-view Depth Consistency on the N3V dataset~\cite{li2022neural}. Comparison of our MoE-GS with other dynamic Gaussian Splatting methods (\textbf{lower is better}).}
\vspace{-3pt}
\label{MDC_ana}
\vspace{-10pt}
\end{center}
\vspace{-5mm}
\end{figure*}

\vspace{-1mm}
\subsection{Comparison Between MoDE and MoE-GS}
\label{sec:mode_moegs_comparison}
\vspace{1mm}

This section provides a direct comparison between MoDE and MoE-GS,
highlighting how their different integration constraints lead to
fundamentally distinct structural properties and optimization behaviors.
We analyze (i) how each formulation composes 3D structure,
and (ii) the resulting efficiency–performance trade-offs.

(i) \textit{Structural Perspective on 3D Reconstruction.}
MoDE and MoE-GS differ fundamentally in how 3D structure is composed.
MoDE directly combines multiple deformation experts at the 3D level
within a shared canonical Gaussian representation.
As illustrated in Fig.~\ref{fig:mode_motion}, this deformation-level
composition visibly alters Gaussian motion behavior.
Compared to 4DGaussians~\cite{wu20244d}, which exhibits rigid and spatially coherent motion,
MoDE (4DGaussians w/ E-D3DGS~\cite{bae2024per}) increases per-Gaussian motion flexibility,
capturing dynamic effects that are not represented by the single-deformation baseline.
These results confirm that composing experts at the 3D level
enhances the representational capacity of the underlying geometry.
In contrast, MoE-GS does not directly compose experts at the 3D
representation stage.
Expert integration is performed in image space during rendering,
and therefore does not explicitly modify a shared canonical 3D structure.
From a structural standpoint, it is not immediately clear whether
such image-space mixture preserves geometric consistency at the 3D level.
However, as described in Sec.~\ref{gaussian-level},
although expert gating is performed at the pixel level,
the routing weights are computed from per-Gaussian signals
(e.g., depth, visibility, and deformation) prior to rasterization.
Because these weights are defined before projection,
they can be lifted back to the Gaussian domain through
responsibility-weighted aggregation, enabling geometric analysis in 3D space.
Leveraging this property, we evaluate the lifted Gaussian-level
geometry using Multi-view Depth Consistency (MDC).
Since MDC measures cross-view alignment via depth reprojection,
independent of image-space blending,
it provides a direct assessment of 3D geometric consistency.
As shown in Fig.~\ref{MDC_ana}, MoE-GS achieves the lowest MDC errors
across multiple scenes and timestamps on the N3V dataset.
These results indicate that, although MoE-GS performs mixture in image space,
its routing mechanism remains geometry-aware and preserves coherent 3D structure.
Detailed definitions and formulations of MDC are provided in Appendix~\ref{MDC}.

(ii) \textit{Training Dynamics and Optimization Trade-off.}
The structural differences between MoDE and MoE-GS lead to
distinct optimization characteristics and cost--performance trade-offs. As shown in Table~\ref{tab:tradeoff}, applying MoDE improves
PSNR over the single-deformation baseline.
However, the magnitude of improvement is relatively moderate.
Because MoDE jointly optimizes multiple deformation experts
within a shared canonical representation, the optimization
process becomes more constrained and can be sensitive to
expert interaction.
Nevertheless, the additional training cost remains modest,
incurring only a limited increase in training time compared
to the baseline model.
In contrast, MoE-GS achieves a substantially larger performance gain.
Since it leverages independently trained experts and adopts a
decoupled optimization strategy, each expert can specialize
without being constrained by shared canonical coupling,
resulting in more stable optimization and stronger reconstruction quality.
However, this comes at the expense of significantly higher training cost,
as each expert must be trained individually and an additional
routing optimization stage is required.
Consequently, the increase in training time for MoE-GS is
considerably larger than that of MoDE.

\begin{table}[t]
\centering
\caption{Performance and training cost comparison between MoDE and MoE-GS.
PSNR is averaged over scenes. Training time is reported in relative units when available.} 
\label{tab:tradeoff}
\resizebox{\linewidth}{!}{
\begin{tabular}{l|cc}
\toprule
Method & PSNR $\uparrow$ & Train Time (h) $\downarrow$ \\
\midrule
4DGaussians~\cite{wu20244d}          & 30.61 & 1.5 \\
4DGaussians~\cite{wu20244d} w/ E-D3DGS~\cite{bae2024per}              & 30.97 & 2.2 \\
4DGaussians~\cite{wu20244d} w/ Grid4D~\cite{xu2024grid4d}              & 30.83 & 2.2 \\
\addlinespace[1.5pt]
\rowcolor{lightgray}
\textbf{MoE-GS (N=2)} & \textbf{32.82} & 6.7 \\
\bottomrule
\end{tabular}}
\vspace{-2mm}
\end{table}
\vspace{-3mm}
\section{Conclusion}
\vspace{-1mm}
In this work, we studied multi-deformation modeling for dynamic Gaussian
Splatting from a Mixture-of-Experts perspective.
We showed that the limitations of existing methods arise not from insufficient
model capacity, but from the reliance on a single deformation prior to represent
heterogeneous motion.
Different deformation formulations induce complementary motion behaviors,
motivating principled deformation composition.
We presented two distinct approaches that instantiate this idea under different integration constraints.
\emph{MoDE} composes multiple deformation experts within a shared canonical
Gaussian representation through joint optimization, offering an efficient and
end-to-end extension of existing pipelines.
In contrast, \emph{MoE-GS} decouples expert optimization from expert integration,
enabling flexible composition of independently trained Gaussian models with
diverse motion parameterizations via learned routing.
Rather than advocating a single solution, our work clarifies how different
integration strategies lead to distinct trade-offs in reconstruction fidelity,
training behavior, and computational efficiency.
By explicitly contrasting deformation-level and model-level expert composition,
this study provides a structured view of the design space for dynamic Gaussian
Splatting and offers guidance for selecting appropriate modeling strategies under
different reconstruction scenarios.

\bibliographystyle{IEEEtran}
\bibliography{main}

@article{mildenhall2021nerf,
  author  = {Mildenhall, Ben and Srinivasan, Pratul P and Tancik, Matthew and Barron, Jonathan T and Ramamoorthi, Ravi and Ng, Ren},
  journal = {Communications of the ACM},
  pages   = {99--106},
  title   = {Nerf: Representing scenes as neural radiance fields for view synthesis},
  volume  = {65},
  year    = {2021},
  publisher = {ACM}
}

@article{kerbl20233d,
  author  = {Kerbl, Bernhard and Kopanas, Georgios and Leimk{\"u}hler, Thomas and Drettakis, George},
  journal = {ACM Transactions on Graphics},
  pages   = {139--1},
  title   = {3D Gaussian Splatting for Real-Time Radiance Field Rendering},
  volume  = {42},
  year    = {2023}
}

@article{hinton2015distilling,
  author  = {Hinton, Geoffrey and Vinyals, Oriol and Dean, Jeff},
  journal = {arXiv preprint arXiv:1503.02531},
  pages   = {1--9},
  title   = {Distilling the Knowledge in a Neural Network},
  volume  = {1503},
  year    = {2015}
}

@article{song2023nerfplayer,
  author  = {Song, Liangchen and Chen, Anpei and Li, Zhong and Chen, Zhang and Chen, Lele and Yuan, Junsong and Xu, Yi and Geiger, Andreas},
  journal = {IEEE Transactions on Visualization and Computer Graphics},
  pages   = {2732--2742},
  title   = {Nerfplayer: A Streamable Dynamic Scene Representation with Decomposed Neural Radiance Fields},
  volume  = {29},
  year    = {2023},
  number  = {5}
}

@inproceedings{attal2023hyperreel,
  author    = {Attal, Benjamin and Huang, Jia-Bin and Richardt, Christian and Zollhoefer, Michael and Kopf, Johannes and O’Toole, Matthew and Kim, Changil},
  booktitle = {Proceedings of the IEEE/CVF Conference on Computer Vision and Pattern Recognition},
  pages     = {16610--16620},
  title     = {HyperReel: High-Fidelity 6-DoF Video with Ray-Conditioned Sampling},
  year      = {2023}
}

@inproceedings{li2022neural,
  author    = {Li, Tianye and Slavcheva, Mira and Zollhoefer, Michael and Green, Simon and Lassner, Christoph and Kim, Changil and Schmidt, Tanner and Lovegrove, Steven and Goesele, Michael and Newcombe, Richard and others},
  booktitle = {Proceedings of the IEEE/CVF Conference on Computer Vision and Pattern Recognition},
  pages     = {5521--5531},
  title     = {Neural 3D Video Synthesis from Multi-View Video},
  year      = {2022}
}

@inproceedings{cao2023hexplane,
  author    = {Cao, Ang and Johnson, Justin},
  booktitle = {Proceedings of the IEEE/CVF Conference on Computer Vision and Pattern Recognition},
  pages     = {130--141},
  title     = {HexPlane: A Fast Representation for Dynamic Scenes},
  year      = {2023}
}

@inproceedings{fridovich2023k,
  author    = {Fridovich-Keil, Sara and Meanti, Giacomo and Warburg, Frederik Rahb{\ae}k and Recht, Benjamin and Kanazawa, Angjoo},
  booktitle = {Proceedings of the IEEE/CVF Conference on Computer Vision and Pattern Recognition},
  pages     = {12479--12488},
  title     = {K-Planes: Explicit Radiance Fields in Space, Time, and Appearance},
  year      = {2023}
}

@inproceedings{wang2023mixed,
  author    = {Wang, Feng and Tan, Sinan and Li, Xinghang and Tian, Zeyue and Song, Yafei and Liu, Huaping},
  booktitle = {Proceedings of the IEEE/CVF International Conference on Computer Vision},
  pages     = {19706--19716},
  title     = {Mixed Neural Voxels for Fast Multi-View Video Synthesis},
  year      = {2023}
}

@inproceedings{lin2023high,
  author    = {Lin, Haotong and Peng, Sida and Xu, Zhen and Xie, Tao and He, Xingyi and Bao, Hujun and Zhou, Xiaowei},
  booktitle = {SIGGRAPH Asia 2023 Conference Papers},
  pages     = {1--9},
  title     = {High-Fidelity and Real-Time Novel View Synthesis for Dynamic Scenes},
  year      = {2023}
}

@inproceedings{li2024spacetime,
  author    = {Li, Zhan and Chen, Zhang and Li, Zhong and Xu, Yi},
  booktitle = {Proceedings of the IEEE/CVF Conference on Computer Vision and Pattern Recognition},
  pages     = {8508--8520},
  title     = {Spacetime Gaussian Feature Splatting for Real-Time Dynamic View Synthesis},
  year      = {2024}
}

@inproceedings{wu20244d,
  author    = {Wu, Guanjun and Yi, Taoran and Fang, Jiemin and Xie, Lingxi and Zhang, Xiaopeng and Wei, Wei and Liu, Wenyu and Tian, Qi and Wang, Xinggang},
  booktitle = {Proceedings of the IEEE/CVF Conference on Computer Vision and Pattern Recognition},
  pages     = {20310--20320},
  title     = {4D Gaussian Splatting for Real-Time Dynamic Scene Rendering},
  year      = {2024}
}

@article{lee2024fully,
  author  = {Lee, Junoh and Won, Changyeon and Jung, Hyunjun and Bae, Inhwan and Jeon, Hae-Gon},
  journal = {Advances in Neural Information Processing Systems},
  pages   = {5384--5409},
  title   = {Fully Explicit Dynamic Gaussian Splatting},
  volume  = {37},
  year    = {2024}
}

@inproceedings{bae2024per,
  author    = {Bae, Jeongmin and Kim, Seoha and Yun, Youngsik and Lee, Hahyun and Bang, Gun and Uh, Youngjung},
  booktitle = {European Conference on Computer Vision},
  pages     = {321--335},
  title     = {Per-Gaussian Embedding-Based Deformation for Deformable 3D Gaussian Splatting},
  year      = {2024},
  organization = {Springer}
}

@inproceedings{sabater2017dataset,
  author    = {Sabater, Neus and Boisson, Guillaume and Vandame, Benoit and Kerbiriou, Paul and Babon, Frederic and Hog, Matthieu and Gendrot, Remy and Langlois, Tristan and Bureller, Olivier and Schubert, Arno and others},
  booktitle = {Proceedings of the IEEE Conference on Computer Vision and Pattern Recognition Workshops},
  pages     = {30--40},
  title     = {Dataset and Pipeline for Multi-View Light-Field Video},
  year      = {2017}
}

@inproceedings{yang2024deformable,
  author    = {Yang, Ziyi and Gao, Xinyu and Zhou, Wen and Jiao, Shaohui and Zhang, Yuqing and Jin, Xiaogang},
  booktitle = {Proceedings of the IEEE/CVF Conference on Computer Vision and Pattern Recognition},
  pages     = {20331--20341},
  title     = {Deformable 3D Gaussians for High-Fidelity Monocular Dynamic Scene Reconstruction},
  year      = {2024}
}

@inproceedings{sun20243dgstream,
  author    = {Sun, Jiakai and Jiao, Han and Li, Guangyuan and Zhang, Zhanjie and Zhao, Lei and Xing, Wei},
  booktitle = {Proceedings of the IEEE/CVF Conference on Computer Vision and Pattern Recognition},
  pages     = {20675--20685},
  title     = {3DGStream: On-the-Fly Training of 3D Gaussians for Efficient Streaming of Photo-Realistic Free-Viewpoint Videos},
  year      = {2024}
}

@article{liu2026dynamics,
  title={Dynamics-aware gaussian splatting streaming towards fast on-the-fly 4d reconstruction},
  author={Liu, Zhening and Hu, Yingdong and Zhang, Xinjie and Song, Rui and Shao, Jiawei and Lin, Zehong and Zhang, Jun},
  journal={IEEE Transactions on Visualization and Computer Graphics},
  year={2026},
  publisher={IEEE}
}

@inproceedings{xie2024mode,
  author    = {Xie, Zhitian and Zhang, Yinger and Zhuang, Chenyi and Shi, Qitao and Liu, Zhining and Gu, Jinjie and Zhang, Guannan},
  booktitle = {Proceedings of the AAAI Conference on Artificial Intelligence},
  pages     = {16067--16075},
  title     = {MODE: A Mixture-of-Experts Model with Mutual Distillation Among the Experts},
  year      = {2024}
}

@inproceedings{lepikhin2021gshard,
  author    = {Lepikhin, Dmitry and Lee, HyoukJoong and Xu, Yuanzhong and Chen, Dehao and Firat, Orhan and Huang, Yanping and Krikun, Maxim and Shazeer, Noam and Chen, Zhifeng},
  booktitle = {International Conference on Learning Representations},
  pages     = {1--14},
  title     = {GShard: Scaling Giant Models with Conditional Computation and Automatic Sharding},
  year      = {2021}
}

@article{fedus2022switch,
  author  = {Fedus, William and Zoph, Barret and Shazeer, Noam},
  journal = {Journal of Machine Learning Research},
  pages   = {1--39},
  title   = {Switch Transformers: Scaling to Trillion Parameter Models with Simple and Efficient Sparsity},
  volume  = {23},
  year    = {2022},
  number  = {120}
}

@inproceedings{shazeer2017outrageously,
  author    = {Shazeer, Noam and Mirhoseini, Azalia and Maziarz, Krzysztof and Davis, Andy and Le, Quoc and Hinton, Geoffrey and Dean, Jeff},
  booktitle = {International Conference on Learning Representations},
  pages     = {1--14},
  title     = {Outrageously Large Neural Networks: The Sparsely-Gated Mixture-of-Experts Layer},
  year      = {2017}
}

@inproceedings{ma2018modeling,
  author    = {Ma, Jiaqi and Zhao, Zhe and Yi, Xinyang and Chen, Jilin and Hong, Lichan and Chi, Ed H.},
  booktitle = {Proceedings of the 24th ACM SIGKDD International Conference on Knowledge Discovery \& Data Mining},
  pages     = {1930--1939},
  title     = {Modeling Task Relationships in Multi-Task Learning with Multi-Gate Mixture-of-Experts},
  year      = {2018}
}

@inproceedings{kong2022efficient,
  author    = {Kong, Yuchen and Lu, Xinghua and Shen, Jianzhuang and Liu, Lingyun and Chen, Baoquan},
  booktitle = {European Conference on Computer Vision},
  pages     = {1--12},
  title     = {Efficient Face Forgery Detection with Mixture of Experts},
  year      = {2022}
}

@inproceedings{meng2024moead,
  author    = {Meng, Shiyuan and Meng, Wenchao and Zhou, Qihang and Li, Shizhong and Hou, Weiye and He, Shibo},
  booktitle = {European Conference on Computer Vision},
  pages     = {345--361},
  title     = {MoEAD: A Parameter-Efficient Model for Multi-Class Anomaly Detection},
  year      = {2024},
  organization = {Springer}
}

@inproceedings{park2021hypernerf,
  author    = {Park, Keunhong and Sinha, Utkarsh and Hedman, Peter and Barron, Jonathan T. and Bouaziz, Sofien and Goldman, Dan B. and Martin-Brualla, Ricardo and Seitz, Steven M.},
  booktitle = {SIGGRAPH Asia},
  pages     = {1--12},
  title     = {HyperNeRF: A Higher-Dimensional Representation for Topologically Varying Neural Radiance Fields},
  year      = {2021}
}

@inproceedings{yan20244d,
  author    = {Yan, Jinbo and Peng, Rui and Tang, Luyang and Wang, Ronggang},
  booktitle = {Proceedings of the 32nd ACM International Conference on Multimedia},
  pages     = {7871--7880},
  title     = {4D Gaussian Splatting with Scale-Aware Residual Field and Adaptive Optimization for Real-Time Rendering of Temporally Complex Dynamic Scenes},
  year      = {2024}
}

@article{liu2024swings,
  author  = {Liu, Bangya and Banerjee, Suman},
  journal = {arXiv preprint arXiv:2409.07759},
  pages   = {1--12},
  title   = {Swings: Sliding Window Gaussian Splatting for Volumetric Video Streaming with Arbitrary Length},
  volume  = {2409},
  year    = {2024}
}

@inproceedings{xu2024grid4d,
  title     = {Grid4d: 4d decomposed hash encoding for high-fidelity dynamic gaussian splatting},
  author    = {Xu, Jiawei and Fan, Zexin and Yang, Jian and Xie, Jin},
  booktitle = {Advances in Neural Information Processing Systems},
  volume    = {37},
  pages     = {123787--123811},
  year      = {2024}
}

@article{oksuz2023mocae,
  author  = {Oksuz, Kemal and Kalkan, Sinan and Akbas, Emre},
  title   = {MOCAE: Mixture of Calibrated Experts Significantly Improves Object Detection},
  journal = {arXiv preprint arXiv:2309.14976},
  year    = {2023}
}

@article{wang2004image,
  title={Image quality assessment: from error visibility to structural similarity},
  author={Wang, Zhou and Bovik, Alan C and Sheikh, Hamid R and Simoncelli, Eero P},
  journal={IEEE transactions on image processing},
  volume={13},
  number={4},
  pages={600--612},
  year={2004},
  publisher={IEEE}
}

@inproceedings{zhang2018unreasonable,
  title={The unreasonable effectiveness of deep features as a perceptual metric},
  author={Zhang, Richard and Isola, Phillip and Efros, Alexei A and Shechtman, Eli and Wang, Oliver},
  booktitle={Proceedings of the IEEE conference on computer vision and pattern recognition},
  pages={586--595},
  year={2018}
}

@inproceedings{chen2023mod,
  title={Mod-squad: Designing mixtures of experts as modular multi-task learners},
  author={Chen, Zitian and Shen, Yikang and Ding, Mingyu and Chen, Zhenfang and Zhao, Hengshuang and Learned-Miller, Erik G and Gan, Chuang},
  booktitle={Proceedings of the IEEE/CVF Conference on Computer Vision and Pattern Recognition},
  pages={11828--11837},
  year={2023}
}

@inproceedings{lewis2021base,
  title={Base layers: Simplifying training of large, sparse models},
  author={Lewis, Mike and Bhosale, Shruti and Dettmers, Tim and Goyal, Naman and Zettlemoyer, Luke},
  booktitle={International Conference on Machine Learning},
  pages={6265--6274},
  year={2021},
  organization={PMLR}
}

@article{hazimeh2021dselect,
  title={Dselect-k: Differentiable selection in the mixture of experts with applications to multi-task learning},
  author={Hazimeh, Hussein and Zhao, Zhe and Chowdhery, Aakanksha and Sathiamoorthy, Maheswaran and Chen, Yihua and Mazumder, Rahul and Hong, Lichan and Chi, Ed},
  journal={Advances in Neural Information Processing Systems},
  volume={34},
  pages={29335--29347},
  year={2021}
}

@article{chi2022representation,
  title={On the representation collapse of sparse mixture of experts},
  author={Chi, Zewen and Dong, Li and Huang, Shaohan and Dai, Damai and Ma, Shuming and Patra, Barun and Singhal, Saksham and Bajaj, Payal and Song, Xia and Mao, Xian-Ling and others},
  journal={Advances in Neural Information Processing Systems},
  volume={35},
  pages={34600--34613},
  year={2022}
}

@article{he2021fastmoe,
  title={Fastmoe: A fast mixture-of-expert training system},
  author={He, Jiaao and Qiu, Jiezhong and Zeng, Aohan and Yang, Zhilin and Zhai, Jidong and Tang, Jie},
  journal={arXiv preprint arXiv:2103.13262},
  year={2021}
}

@inproceedings{rajbhandari2022deepspeed,
  title={Deepspeed-moe: Advancing mixture-of-experts inference and training to power next-generation ai scale},
  author={Rajbhandari, Samyam and Li, Conglong and Yao, Zhewei and Zhang, Minjia and Aminabadi, Reza Yazdani and Awan, Ammar Ahmad and Rasley, Jeff and He, Yuxiong},
  booktitle={International conference on machine learning},
  pages={18332--18346},
  year={2022},
  organization={PMLR}
}

@article{yu2024moesys,
  title={Moesys: A distributed and efficient mixture-of-experts training and inference system for internet services},
  author={Yu, Dianhai and Shen, Liang and Hao, Hongxiang and Gong, Weibao and Wu, Huachao and Bian, Jiang and Dai, Lirong and Xiong, Haoyi},
  journal={IEEE Transactions on Services Computing},
  volume={17},
  number={5},
  pages={2626--2639},
  year={2024},
  publisher={IEEE}
}

@inproceedings{he2022fastermoe,
  title={Fastermoe: modeling and optimizing training of large-scale dynamic pre-trained models},
  author={He, Jiaao and Zhai, Jidong and Antunes, Tiago and Wang, Haojie and Luo, Fuwen and Shi, Shangfeng and Li, Qin},
  booktitle={Proceedings of the 27th ACM SIGPLAN Symposium on Principles and Practice of Parallel Programming},
  pages={120--134},
  year={2022}
}

@inproceedings{singh2023hybrid,
  title={A hybrid tensor-expert-data parallelism approach to optimize mixture-of-experts training},
  author={Singh, Siddharth and Ruwase, Olatunji and Awan, Ammar Ahmad and Rajbhandari, Samyam and He, Yuxiong and Bhatele, Abhinav},
  booktitle={Proceedings of the 37th International Conference on Supercomputing},
  pages={203--214},
  year={2023}
}

@article{nie2023flexmoe,
  title={Flexmoe: Scaling large-scale sparse pre-trained model training via dynamic device placement},
  author={Nie, Xiaonan and Miao, Xupeng and Wang, Zilong and Yang, Zichao and Xue, Jilong and Ma, Lingxiao and Cao, Gang and Cui, Bin},
  journal={Proceedings of the ACM on Management of Data},
  volume={1},
  number={1},
  pages={1--19},
  year={2023},
  publisher={ACM New York, NY, USA}
}

@inproceedings{zhai2023smartmoe,
  title={$\{$SmartMoE$\}$: Efficiently training $\{$Sparsely-Activated$\}$ models through combining offline and online parallelization},
  author={Zhai, Mingshu and He, Jiaao and Ma, Zixuan and Zong, Zan and Zhang, Runqing and Zhai, Jidong},
  booktitle={2023 USENIX Annual Technical Conference (USENIX ATC 23)},
  pages={961--975},
  year={2023}
}

@article{lakshminarayanan2017simple,
  title={Simple and scalable predictive uncertainty estimation using deep ensembles},
  author={Lakshminarayanan, Balaji and Pritzel, Alexander and Blundell, Charles},
  journal={Advances in neural information processing systems},
  volume={30},
  year={2017}
}

@incollection{levoy2023light,
  title={Light field rendering},
  author={Levoy, Marc and Hanrahan, Pat},
  booktitle={Seminal Graphics Papers: Pushing the Boundaries, Volume 2},
  pages={441--452},
  year={2023}
}

@article{bergen1991plenoptic,
  title={The plenoptic function and the elements of early vision},
  author={Bergen, James R and Adelson, Edward H},
  journal={Computational models of visual processing},
  volume={1},
  number={8},
  pages={3},
  year={1991}
}

@inproceedings{guo2023forward,
  title={Forward flow for novel view synthesis of dynamic scenes},
  author={Guo, Xiang and Sun, Jiadai and Dai, Yuchao and Chen, Guanying and Ye, Xiaoqing and Tan, Xiao and Ding, Errui and Zhang, Yumeng and Wang, Jingdong},
  booktitle={Proceedings of the IEEE/CVF International Conference on Computer Vision},
  pages={16022--16033},
  year={2023}
}

@article{liu2022devrf,
  title={Devrf: Fast deformable voxel radiance fields for dynamic scenes},
  author={Liu, Jia-Wei and Cao, Yan-Pei and Mao, Weijia and Zhang, Wenqiao and Zhang, David Junhao and Keppo, Jussi and Shan, Ying and Qie, Xiaohu and Shou, Mike Zheng},
  journal={Advances in Neural Information Processing Systems},
  volume={35},
  pages={36762--36775},
  year={2022}
}

@inproceedings{pumarola2021d,
  title={D-nerf: Neural radiance fields for dynamic scenes},
  author={Pumarola, Albert and Corona, Enric and Pons-Moll, Gerard and Moreno-Noguer, Francesc},
  booktitle={Proceedings of the IEEE/CVF conference on computer vision and pattern recognition},
  pages={10318--10327},
  year={2021}
}

@article{wang2021neural,
  title={Neural trajectory fields for dynamic novel view synthesis},
  author={Wang, Chaoyang and Eckart, Ben and Lucey, Simon and Gallo, Orazio},
  journal={arXiv preprint arXiv:2105.05994},
  year={2021}
}

@inproceedings{park2021nerfies,
  title={Nerfies: Deformable neural radiance fields},
  author={Park, Keunhong and Sinha, Utkarsh and Barron, Jonathan T and Bouaziz, Sofien and Goldman, Dan B and Seitz, Steven M and Martin-Brualla, Ricardo},
  booktitle={Proceedings of the IEEE/CVF international conference on computer vision},
  pages={5865--5874},
  year={2021}
}

@inproceedings{shao2023tensor4d,
  title={Tensor4d: Efficient neural 4d decomposition for high-fidelity dynamic reconstruction and rendering},
  author={Shao, Ruizhi and Zheng, Zerong and Tu, Hanzhang and Liu, Boning and Zhang, Hongwen and Liu, Yebin},
  booktitle={Proceedings of the IEEE/CVF Conference on Computer Vision and Pattern Recognition},
  pages={16632--16642},
  year={2023}
}

@article{wang2023masked,
  title={Masked space-time hash encoding for efficient dynamic scene reconstruction},
  author={Wang, Feng and Chen, Zilong and Wang, Guokang and Song, Yafei and Liu, Huaping},
  journal={Advances in neural information processing systems},
  volume={36},
  pages={70497--70510},
  year={2023}
}

@inproceedings{wang2023neural,
  title={Neural residual radiance fields for streamably free-viewpoint videos},
  author={Wang, Liao and Hu, Qiang and He, Qihan and Wang, Ziyu and Yu, Jingyi and Tuytelaars, Tinne and Xu, Lan and Wu, Minye},
  booktitle={Proceedings of the IEEE/CVF Conference on Computer Vision and Pattern Recognition},
  pages={76--87},
  year={2023}
}

@inproceedings{duan20244d,
  title={4d-rotor gaussian splatting: towards efficient novel view synthesis for dynamic scenes},
  author={Duan, Yuanxing and Wei, Fangyin and Dai, Qiyu and He, Yuhang and Chen, Wenzheng and Chen, Baoquan},
  booktitle={ACM SIGGRAPH 2024 Conference Papers},
  pages={1--11},
  year={2024}
}

@inproceedings{lin2024gaussian,
  title={Gaussian-flow: 4d reconstruction with dynamic 3d gaussian particle},
  author={Lin, Youtian and Dai, Zuozhuo and Zhu, Siyu and Yao, Yao},
  booktitle={Proceedings of the IEEE/CVF Conference on Computer Vision and Pattern Recognition},
  pages={21136--21145},
  year={2024}
}

@inproceedings{luiten2024dynamic,
  title={Dynamic 3d gaussians: Tracking by persistent dynamic view synthesis},
  author={Luiten, Jonathon and Kopanas, Georgios and Leibe, Bastian and Ramanan, Deva},
  booktitle={2024 International Conference on 3D Vision (3DV)},
  pages={800--809},
  year={2024},
  organization={IEEE}
}

@inproceedings{kratimenos2024dynmf,
  title={Dynmf: Neural motion factorization for real-time dynamic view synthesis with 3d gaussian splatting},
  author={Kratimenos, Agelos and Lei, Jiahui and Daniilidis, Kostas},
  booktitle={European Conference on Computer Vision},
  pages={252--269},
  year={2024},
  organization={Springer}
}

@inproceedings{duisterhof2024deformgs,
  title={Deformgs: Scene flow in highly deformable scenes for deformable object manipulation},
  author={Duisterhof, Bardienus P and Zhao, Mandi and Yao, Yunchao and Liu, Jia-Wei and Seidenschwarz, Jenny and Shou, Mike Zheng and Ramanan, Deva and Song, Shuran and Birchfield, Stan and Wen, Bowen and others},
  booktitle={International Workshop on the Algorithmic Foundations of Robotics},
  pages={263--282},
  year={2024},
  organization={Springer}
}

@inproceedings{liang2025gaufre,
  title={Gaufre: Gaussian deformation fields for real-time dynamic novel view synthesis},
  author={Liang, Yiqing and Khan, Numair and Li, Zhengqin and Nguyen-Phuoc, Thu and Lanman, Douglas and Tompkin, James and Xiao, Lei},
  booktitle={2025 IEEE/CVF Winter Conference on Applications of Computer Vision (WACV)},
  pages={2642--2652},
  year={2025},
  organization={IEEE}
}

@inproceedings{huang2024sc,
  title={Sc-gs: Sparse-controlled gaussian splatting for editable dynamic scenes},
  author={Huang, Yi-Hua and Sun, Yang-Tian and Yang, Ziyi and Lyu, Xiaoyang and Cao, Yan-Pei and Qi, Xiaojuan},
  booktitle={Proceedings of the IEEE/CVF conference on computer vision and pattern recognition},
  pages={4220--4230},
  year={2024}
}

@inproceedings{yu2024cogs,
  title={Cogs: Controllable gaussian splatting},
  author={Yu, Heng and Julin, Joel and Milacski, Zolt{\'a}n {\'A} and Niinuma, Koichiro and Jeni, L{\'a}szl{\'o} A},
  booktitle={Proceedings of the IEEE/CVF Conference on Computer Vision and Pattern Recognition},
  pages={21624--21633},
  year={2024}
}

@inproceedings{feichtenhofer2019slowfast,
  title={Slowfast networks for video recognition},
  author={Feichtenhofer, Christoph and Fan, Haoqi and Malik, Jitendra and He, Kaiming},
  booktitle={Proceedings of the IEEE/CVF international conference on computer vision},
  pages={6202--6211},
  year={2019}
}

@book{bartels1995introduction,
  title={An introduction to splines for use in computer graphics and geometric modeling},
  author={Bartels, Richard H and Beatty, John C and Barsky, Brian A},
  year={1995},
  publisher={Morgan Kaufmann}
}

@inproceedings{shoemake1985animating,
  title={Animating rotation with quaternion curves},
  author={Shoemake, Ken},
  booktitle={Proceedings of the 12th annual conference on Computer graphics and interactive techniques},
  pages={245--254},
  year={1985}
}

@article{bucilua2006model,
  title={Model compression, in proceedings of the 12 th acm sigkdd international conference on knowledge discovery and data mining},
  author={Bucilua, C and Caruana, Rich and Niculescu-Mizil, Alexandru},
  journal={New York, NY, USA},
  volume={3},
  year={2006}
}

@article{ba2014deep,
  title={Do deep nets really need to be deep?},
  author={Ba, Jimmy and Caruana, Rich},
  journal={Advances in neural information processing systems},
  volume={27},
  year={2014}
}

@article{yu2025moe,
  title={MoE-Adapters++: Towards More Efficient Continual Learning of Vision-Language Models via Dynamic Mixture-of-Experts Adapters},
  author={Yu, Jiazuo and Huang, Zichen and Zhuge, Yunzhi and Zhang, Lu and Hu, Ping and Wang, Dong and Lu, Huchuan and He, You},
  journal={IEEE Transactions on Pattern Analysis and Machine Intelligence},
  year={2025},
  publisher={IEEE}
}

@inproceedings{yu2024boosting,
  title={Boosting continual learning of vision-language models via mixture-of-experts adapters},
  author={Yu, Jiazuo and Zhuge, Yunzhi and Zhang, Lu and Hu, Ping and Wang, Dong and Lu, Huchuan and He, You},
  booktitle={Proceedings of the IEEE/CVF Conference on Computer Vision and Pattern Recognition},
  pages={23219--23230},
  year={2024}
}

@article{li2025uni,
  title={Uni-moe: Scaling unified multimodal llms with mixture of experts},
  author={Li, Yunxin and Jiang, Shenyuan and Hu, Baotian and Wang, Longyue and Zhong, Wanqi and Luo, Wenhan and Ma, Lin and Zhang, Min},
  journal={IEEE Transactions on Pattern Analysis and Machine Intelligence},
  year={2025},
  publisher={IEEE}
}

@article{kheradmand20243d,
  title={3d gaussian splatting as markov chain monte carlo},
  author={Kheradmand, Shakiba and Rebain, Daniel and Sharma, Gopal and Sun, Weiwei and Tseng, Yang-Che and Isack, Hossam and Kar, Abhishek and Tagliasacchi, Andrea and Yi, Kwang Moo},
  journal={Advances in Neural Information Processing Systems},
  volume={37},
  pages={80965--80986},
  year={2024}
}

@inproceedings{chen2025dash,
  title={DASH: 4D Hash Encoding with Self-Supervised Decomposition for Real-Time Dynamic Scene Rendering},
  author={Chen, Jie and Hu, Zhangchi and Wu, Peixi and Zhu, Huyue and Li, Hebei and Sun, Xiaoyan},
  booktitle={Proceedings of the IEEE/CVF International Conference on Computer Vision},
  pages={26349--26359},
  year={2025}
}

@inproceedings{wu2025localdygs,
  title={LocalDyGS: Multi-view Global Dynamic Scene Modeling via Adaptive Local Implicit Feature Decoupling},
  author={Wu, Jiahao and Peng, Rui and Jiao, Jianbo and Yang, Jiayu and Tang, Luyang and Xiong, Kaiqiang and Liang, Jie and Yan, Jinbo and Liu, Runling and Wang, Ronggang},
  booktitle={Proceedings of the IEEE/CVF International Conference on Computer Vision},
  pages={9519--9529},
  year={2025}
}

@article{chen2026haif,
  title={Haif-gs: Hierarchical and induced flow-guided gaussian splatting for dynamic scene},
  author={Chen, Jianing and Li, Zehao and Cai, Yujun and Jiang, Hao and Qian, Chengxuan and Kang, Juyuan and Gao, Shuqin and Zhao, Honglong and Mao, Tianlu and Zhang, Yucheng},
  journal={Advances in Neural Information Processing Systems},
  volume={38},
  pages={125539--125563},
  year={2026}
}

@inproceedings{jiang2025timeformer,
  title={Timeformer: Capturing temporal relationships of deformable 3d gaussians for robust reconstruction},
  author={Jiang, Dadong and Hou, Zhi and Ke, Zhihui and Yang, Xianghui and Zhou, Xiaobo and Qiu, Tie},
  booktitle={Proceedings of the IEEE/CVF International Conference on Computer Vision},
  pages={8721--8732},
  year={2025}
}

@inproceedings{wang2025freetimegs,
  title={FreeTimeGS: Free Gaussian Primitives at Anytime Anywhere for Dynamic Scene Reconstruction},
  author={Wang, Yifan and Yang, Peishan and Xu, Zhen and Sun, Jiaming and Zhang, Zhanhua and Chen, Yong and Bao, Hujun and Peng, Sida and Zhou, Xiaowei},
  booktitle={Proceedings of the Computer Vision and Pattern Recognition Conference},
  pages={21750--21760},
  year={2025}
}

@inproceedings{gao20257dgs,
  title={7DGS: Unified spatial-temporal-angular Gaussian splatting},
  author={Gao, Zhongpai and Planche, Benjamin and Zheng, Meng and Choudhuri, Anwesa and Chen, Terrence and Wu, Ziyan},
  booktitle={Proceedings of the IEEE/CVF International Conference on Computer Vision},
  pages={26316--26325},
  year={2025}
}

@article{chen2025hac++,
  title={Hac++: Towards 100x compression of 3d gaussian splatting},
  author={Chen, Yihang and Wu, Qianyi and Lin, Weiyao and Harandi, Mehrtash and Cai, Jianfei},
  journal={IEEE Transactions on Pattern Analysis and Machine Intelligence},
  year={2025},
  publisher={IEEE}
}

@article{qu2024z,
  title={Z-splat: Z-axis gaussian splatting for camera-sonar fusion},
  author={Qu, Ziyuan and Vengurlekar, Omkar and Qadri, Mohamad and Zhang, Kevin and Kaess, Michael and Metzler, Christopher and Jayasuriya, Suren and Pediredla, Adithya},
  journal={IEEE Transactions on Pattern Analysis and Machine Intelligence},
  year={2024},
  publisher={IEEE}
}

@article{mi2025learning,
  title={Learning heterogeneous mixture of scene experts for large-scale neural radiance fields},
  author={Mi, Zhenxing and Yin, Ping and Xiao, Xue and Xu, Dan},
  journal={IEEE Transactions on Pattern Analysis and Machine Intelligence},
  year={2025},
  publisher={IEEE}
}

@article{yin2025ms,
  title={MS-NeRF: Multi-space neural radiance fields},
  author={Yin, Ze-Xin and Jiao, Peng-Yi and Qiu, Jiaxiong and Cheng, Ming-Ming and Ren, Bo},
  journal={IEEE Transactions on Pattern Analysis and Machine Intelligence},
  year={2025},
  publisher={IEEE}
}

@article{verbin2024ref,
  title={Ref-nerf: Structured view-dependent appearance for neural radiance fields},
  author={Verbin, Dor and Hedman, Peter and Mildenhall, Ben and Zickler, Todd and Barron, Jonathan T and Srinivasan, Pratul P},
  journal={IEEE Transactions on Pattern Analysis and Machine Intelligence},
  volume={47},
  number={11},
  pages={9426--9437},
  year={2024},
  publisher={IEEE}
}

@inproceedings{jin2026moe,
  title={MoE-GS: Mixture of Experts for Dynamic Gaussian Splatting},
  author={Jin, In-Hwan and Mun, Hyeongju and Kim, Joonsoo and Yun, Kugjin and Kong, Kyeongbo},
  booktitle={International Conference on Learning Representations},
  year={2026}
}

@inproceedings{joo2015panoptic,
  title={Panoptic studio: A massively multiview system for social motion capture},
  author={Joo, Hanbyul and Liu, Hao and Tan, Lei and Gui, Lin and Nabbe, Bart and Matthews, Iain and Kanade, Takeo and Nobuhara, Shohei and Sheikh, Yaser},
  booktitle={Proceedings of the IEEE international conference on computer vision},
  pages={3334--3342},
  year={2015}
}

@inproceedings{liu2025modgs,
  title={Modgs: Dynamic gaussian splatting from casually-captured monocular videos with depth priors},
  author={Liu, Qingming and Liu, Yuan and Wang, Jiepeng and Lyu, Xianqiang and Wang, Peng and Wang, Wenping and Hou, Junhui},
  booktitle={International Conference on Learning Representations},
  volume={2025},
  pages={97048--97074},
  year={2025}
}

@inproceedings{yang2024real,
  title={Real-time photorealistic dynamic scene representation and rendering with 4d gaussian splatting},
  author={Yang, Zeyu and Yang, Hongye and Pan, Zijie and Zhang, Li},
  booktitle={International Conference on Learning Representations},
  volume={2024},
  pages={9142--9159},
  year={2024}
}

@article{yang20244d,
  title={4d gaussian splatting: Modeling dynamic scenes with native 4d primitives},
  author={Yang, Zeyu and Pan, Zijie and Zhu, Xiatian and Zhang, Li and Feng, Jianfeng and Jiang, Yu-Gang and Torr, Philip HS},
  journal={arXiv preprint arXiv:2412.20720},
  year={2024}
}

@inproceedings{zhang2025mega,
  title={Mega: Memory-efficient 4d gaussian splatting for dynamic scenes},
  author={Zhang, Xinjie and Liu, Zhening and Zhang, Yifan and Ge, Xingtong and He, Dailan and Xu, Tongda and Wang, Yan and Lin, Zehong and Yan, Shuicheng and Zhang, Jun},
  booktitle={Proceedings of the IEEE/CVF International Conference on Computer Vision},
  pages={27828--27838},
  year={2025}
}

@inproceedings{cho20264d,
  title={4d scaffold gaussian splatting with dynamic-aware anchor growing for efficient and high-fidelity dynamic scene reconstruction},
  author={Cho, Woong Oh and Cho, In and Kim, Seoha and Bae, Jeongmin and Uh, Youngjung and Kim, Seon Joo},
  booktitle={Proceedings of the AAAI Conference on Artificial Intelligence},
  volume={40},
  number={5},
  pages={3363--3371},
  year={2026}
}

@inproceedings{wang2025shape,
  title={Shape of motion: 4d reconstruction from a single video},
  author={Wang, Qianqian and Ye, Vickie and Gao, Hang and Zeng, Weijia and Austin, Jake and Li, Zhengqi and Kanazawa, Angjoo},
  booktitle={Proceedings of the IEEE/CVF International Conference on Computer Vision},
  pages={9660--9672},
  year={2025}
}
 
\begin{IEEEbiography}
[{\includegraphics[width=1in,height=1.25in,clip,keepaspectratio]{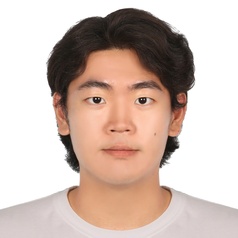}}]
{In-Hwan Jin}
is currently pursuing an
M.S. degree in Electrical and Electronics Engineering at Pusan National University, Busan, Republic of Korea. His current research interests include image processing and computer vision, particularly in 3D vision and dynamic scene reconstruction.
\end{IEEEbiography}

\begin{IEEEbiography}
[{\includegraphics[width=1in,height=1.25in,clip,keepaspectratio]{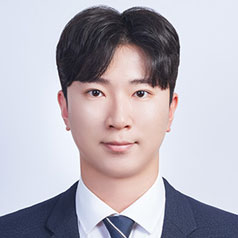}}]
{Hyeongju Mun}
received the
B.S. degree in Electrical and Electronics Engineering from Pusan National
University, Busan, Republic of Korea,
in 2025. He is currently pursuing an
M.S. degree in Electrical and Electronics Engineering
at Pusan National University, Busan, Republic of
Korea. His current research interests include image
processing, computer vision, machine learning, and
deep learning.
\end{IEEEbiography}

\begin{IEEEbiography}
[{\includegraphics[width=1in,height=1.25in,clip,keepaspectratio]{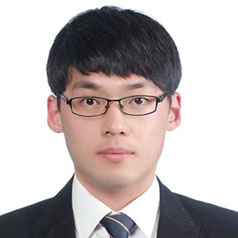}}]
{Joonsoo Kim}
received his B.S. and
Ph.D. degrees in Electrical Engineering from Seoul National University
(SNU), Seoul, Republic of Korea, in
2012 and 2017, respectively. He has
been with the Electronics and Telecommunications
Research Institute since 2017, where he is currently a
principal member of the research staff in the Immersive
Media Research Section. His current research interests include light field displays, autostereoscopic 3D displays, computer vision, and machine learning.
\end{IEEEbiography}

\begin{IEEEbiography}
[{\includegraphics[width=1in,height=1.25in,clip,keepaspectratio]{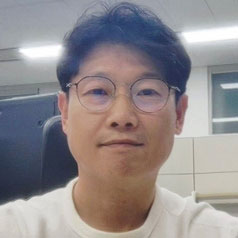}}]
{Kugjin Yun}
received the M.S. degree
in Computer Engineering from
Chungbuk National University,
Cheongju, Republic of Korea, in 2001,
and the Ph.D. degree in Electronics
and Radio Engineering from Kyung Hee University,
Seoul, Republic of Korea, in 2016. In 2001, he joined the Electronics and Telecommunications Research Institute. He is currently a principal member of the research staff with the Immersive Media Research Section, Media Research Division. His research interests include augmented reality (AR), virtual reality (VR), light field (LF) media processing, machine learning, and realistic media technology.
\end{IEEEbiography}

\begin{IEEEbiography}
[{\includegraphics[width=1in,height=1.25in,clip,keepaspectratio]{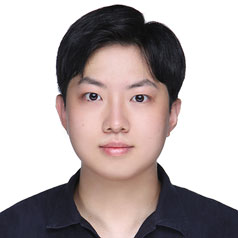}}]
{Kyeongbo Kong}
received the
B.S. degree in Electronics Engineering
from Sogang University, Seoul,
Republic of Korea, in 2015, and the
M.S. and Ph.D. degrees in Electrical
Engineering from the Pohang University of Science
and Technology (POSTECH), Pohang, Republic of
Korea, in 2017 and 2020, respectively. From 2020 to
2021, he worked as a postdoctoral fellow with the
Department of Electrical Engineering, POSTECH,
Pohang, Republic of Korea. From 2021 to 2023, he
was an assistant professor in the Media School at
Pukyong National University, Busan. He is currently
an associate professor of Electrical and Electronics
Engineering at Pusan National University. His
research interests include image processing, computer
vision, machine learning, and deep learning.
\end{IEEEbiography}

\clearpage
\appendix
\label{sup}
\textbf{Overview}

This appendix includes detailed implementation information in Appendix~\ref{sup_impelemenation}, additional quantitative and qualitative results in Appendix~\ref{sup_additonal}, More details of ablation studies on MoE-GS in Appendix~\ref{sup_additonal_abl}, Gaussian-Level Interpretation and Geometry Evaluation Appendix~\ref{MoE_3D_interp}, analysis of expert specialization in Appendix~\ref{expert_analysis}

\subsubsection{Referenced in the Main Paper}
The sections listed below are directly referenced in the main paper for further details:
\begin{itemize}
    \item Optimization strategy of MoE-GS (Appendix~\ref{sup_two-stage})
    \item Implementation details of Expert Training (Appendix~\ref{sup_expert_training})
    \item Implementation details of Router Training (Appendix~\ref{sup_router_training})
    \item Implementation details of Single-Pass Multi-Expert Rendering (Appendix~\ref{sup_single_pass_rendering})
    \item Quantitative Results (Appendix~\ref{sup_additonal_quanti})
    \item Qualitative Results (Appendix~\ref{sup_additonal_quali})
    \item Details of Training Cost and Budget Sensitivity 
    (Appendix~\ref{sup_additonal_abl_train})
    \item Details of Distillation Strategies 
    (Appendix~\ref{sup_weigth_ablation})
    \item Per-Gaussian Contributions and Responsibilities
    (Appendix~\ref{MoE_3D_per_gau})
    \item Post-hoc Gaussian Fusion using Lifting Weights
    (Appendix~\ref{Post_hoc})
    \item Multi-view Depth Consistency Evaluation
    (Appendix~\ref{MDC})
    \item Analysis of Expert Specialization (Appendix~\ref{expert_analysis})  
\end{itemize}

\subsection{Implementation Details}
\label{sup_impelemenation}
\noindent
This section provides implementation and training details of the proposed MoE-GS framework.
We first describe the overall two-stage training strategy, which separates expert model training from router optimization to ensure stable convergence and prevent dominance by faster-converging experts.
Next, we present the training setup for individual expert models, including initialization and baseline-aligned hyperparameters.
Finally, we outline architectural details of the proposed Volume-aware Pixel Router, which enables spatially and temporally coherent expert blending in the MoE-GS pipeline.

\subsubsection{Two-Stage Training Strategy}
\label{sup_two-stage}
To ensure stable convergence and balanced optimization, we adopt a two-stage training strategy that decouples expert training from router optimization.
Jointly training both components can lead to suboptimal convergence, as faster-converging experts tend to dominate the gating process early on, leaving others underutilized and under-optimized.
To mitigate this, we first train each expert model independently to ensure that it can reconstruct the entire scene without relying on other experts. Once trained, all expert parameters are frozen.
In the second stage, we optimize the routing components—specifically the per-Gaussian parameters ${w_i, w_i^{dir}, w_i^{time}}$ and the MLP $\Phi$—to learn an adaptive gating strategy that dynamically selects and blends experts based on spatial, temporal, and view-dependent cues.

\subsubsection{Stage 1: Expert Training}
\label{sup_expert_training}
In the first step of MoE-GS training, each expert model is independently optimized before integration into the MoE framework. Since MoE-GS reconstructs dynamic scenes by blending the outputs of multiple experts, it is critical that each expert achieves its best possible performance. To this end, we retain the original training strategies proposed in their respective works without modification.
All expert models are initialized using point clouds generated by COLMAP.

\begin{itemize}
    \item Ex4DGS~\cite{lee2024fully} is initialized with sparse point clouds obtained via Structure-from-Motion (SfM).
    \item 4DGaussians~\cite{wu20244d} and E-D3DGS\cite{bae2024per} are initialized with downsampled versions of these dense point clouds.
    \item For STG~\cite{li2024spacetime}, we follow its original strategy, merging point clouds from all frames to obtain a globally consistent initialization that serves as a strong prior for optimizing Gaussian attributes.
\end{itemize}

Each expert is trained using its original learning rate schedule, as the hyperparameters are specifically tuned to each model’s architecture and deformation representation.

\subsubsection{Stage 2: Router Training}
\label{sup_router_training}
In the second stage of training, we optimize the Volume-aware Pixel Router while keeping all expert models frozen. 
This allows the router to focus solely on learning effective expert blending strategies without being influenced by the convergence rate of individual experts.
Specifically, we optimize per-Gaussian parameters ${w_i, w_i^{dir}, w_i^{time}}$ and the MLP $\Phi$, which collectively determine the final expert weights for blending. 
All three parameters are fully learnable. To avoid introducing any handcrafted directional or temporal bias, we initialize $w_i^{dir}$ and $w_i^{time}$ using neutral near-zero constants, allowing the router to gradually learn meaningful directional and temporal sensitivities directly from data.
Joint training of experts and the router often leads to suboptimal expert utilization, as rapidly converging experts can dominate the early stages of routing. Our two-stage approach avoids this issue by ensuring that all experts are first trained to full capacity, allowing the router to later learn how to combine their outputs most effectively.

\begin{figure}[t]

\centering
\includegraphics[width=\linewidth]{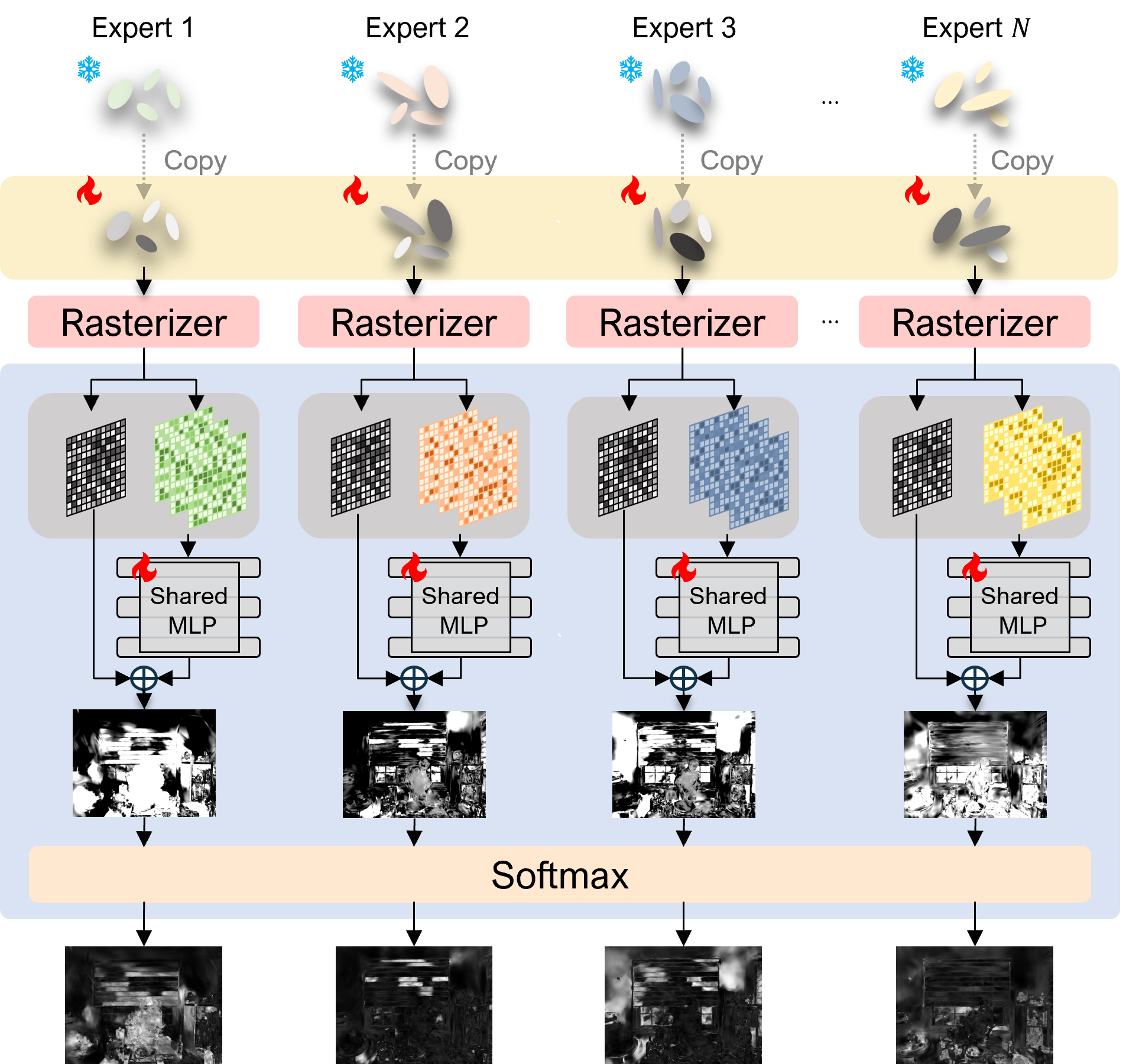}
\caption{Architectural details of the Volume-aware Pixel Router.
}
\label{fig:sup_router}
\end{figure}

\textbf{Volume-aware Pixel Router.}
To generate routing weights, our router begins with the per-Gaussian weights $w_i^{\text{per}}$ defined in 3D space. As illustrated in Figure~\ref{fig:sup_router}, each expert’s Gaussian is duplicated, and its color attribute is replaced with a learnable weights $w_i^{\text{per}}$. The remaining attributes—such as position, scale, and rotation—are directly copied from the pre-trained expert Gaussians. This enables $w_i^{\text{per}}$ to make the router aware of the expert’s volumetric structure, allowing the 2D splatted weights to reflect how each expert behaves under different viewing directions and time steps.
To further model view- and time-dependent variation, we also splat auxiliary features $w_{2D}^{dir}$, $w_{2D}^{time}$, and the ray direction $r$, which are fed into a lightweight MLP. The output of this MLP is added to $w_{2D}$ in a residual manner to produce the final routing weights.

To promote spatial coherence, we adopt a convolutional MLP architecture that leverages local pixel context. For computational efficiency, this MLP is shared across all experts. We empirically set the learning rate to 0.05 for the shared MLP and $w_i^{time}$, and to 0.5 for $w_i$ and $w_i^{dir}$.
This modular and coherent design enables the router to learn adaptive, per-pixel expert blending strategies that generalize effectively across diverse dynamic scenes.

\subsubsection{Single-Pass Multi-Expert Rendering}
\label{sup_single_pass_rendering}
To efficiently deploy MoE-GS, we complement our design with an optimized rendering pipeline. In particular, independently rasterizing each expert triggers repeated kernel launches and separate memory traversal over each expert’s Gaussian buffer, significantly increasing memory IO pressure.
To address this bottleneck, we adopt a Single-Pass Multi-Expert Rendering strategy, which processes all Gaussians in a single batched pass and eliminates these redundant kernel launches while still preserving expert-specific outputs.
For completeness, we also investigated two alternative implementations for rendering multiple experts: (1) sequential expert execution with CPU offloading, and (2) distributing experts across multiple GPUs.

\textbf{Sequential execution} becomes prohibitively slow due to repeated PCIe transfers between the host and device. Gaussian Splatting requires high-frequency random access to Gaussian attributes (positions, scales, SH coefficients, opacities) during splatting. Moving these buffers back and forth over PCIe significantly increases latency and prevents any effective kernel fusion, leading to very large slowdowns in practice.

\textbf{Multi-GPU distribution} also provides limited benefit. Each expert resides in a heterogeneous and non-alignable 3D deformation space, so distributing experts across GPUs requires duplicating each expert’s Gaussian buffer as well as synchronizing per-pixel accumulations across devices. This synchronization step introduces substantial inter-GPU communication overhead, often outweighing the parallelism benefits and resulting in negligible wall-clock speedup.

These observations motivate our choice of the Single-Pass Multi-Expert Rendering strategy as the most stable and efficient trade-off for MoE-GS workloads.

\vspace{-2mm}
\subsection{Additional Quantitative and Qualitative Results}
\label{sup_additonal}
\subsubsection{Quantitative results}
\label{sup_additonal_quanti}
To comprehensively evaluate MoE-GS across perceptual, structural, and pixel-wise metrics, Tables~\ref{tab:appen_n3v_qunati} and~\ref{tab:appen_tech_quanti} present per-scene quantitative results on the N3V~\cite{li2022neural} and Technicolor~\cite{sabater2017dataset} datasets. Table~\ref{tab:appen_n3v_qunati} reports N3V results using a 4-expert configuration, where MoE-GS consistently outperforms individual experts across all scenes and metrics, demonstrating strong generalization. Table~\ref{tab:appen_tech_quanti} shows results on Technicolor with a 3-expert setup using E-D3DGS~\cite{bae2024per}, STG~\cite{li2024spacetime}, and Ex4DGS~\cite{lee2024fully}. MoE-GS maintains strong performance even with fewer experts, achieving top or near-top performance across most scenes and the highest overall average across all evaluation metrics.
To assess its scalability to monocular settings, Table~\ref{tab:monocular} evaluates MoE-GS on the HyperNeRF dataset~\cite{park2021hypernerf}. Despite the limited input views, MoE-GS transfers well to this setting, maintaining high fidelity and demonstrating strong adaptability beyond the multi-view reconstruction scenario.

In addition, Table~\ref{tab:distillation_ablation} presents an ablation study of our distillation strategies on the Technicolor dataset, reporting per-scene results and highlighting the contribution of each component to final performance.

\subsubsection{Qualitative results}
\label{sup_additonal_quali}
\begin{figure*}[t]
  \centering
  \includegraphics[width=\textwidth]{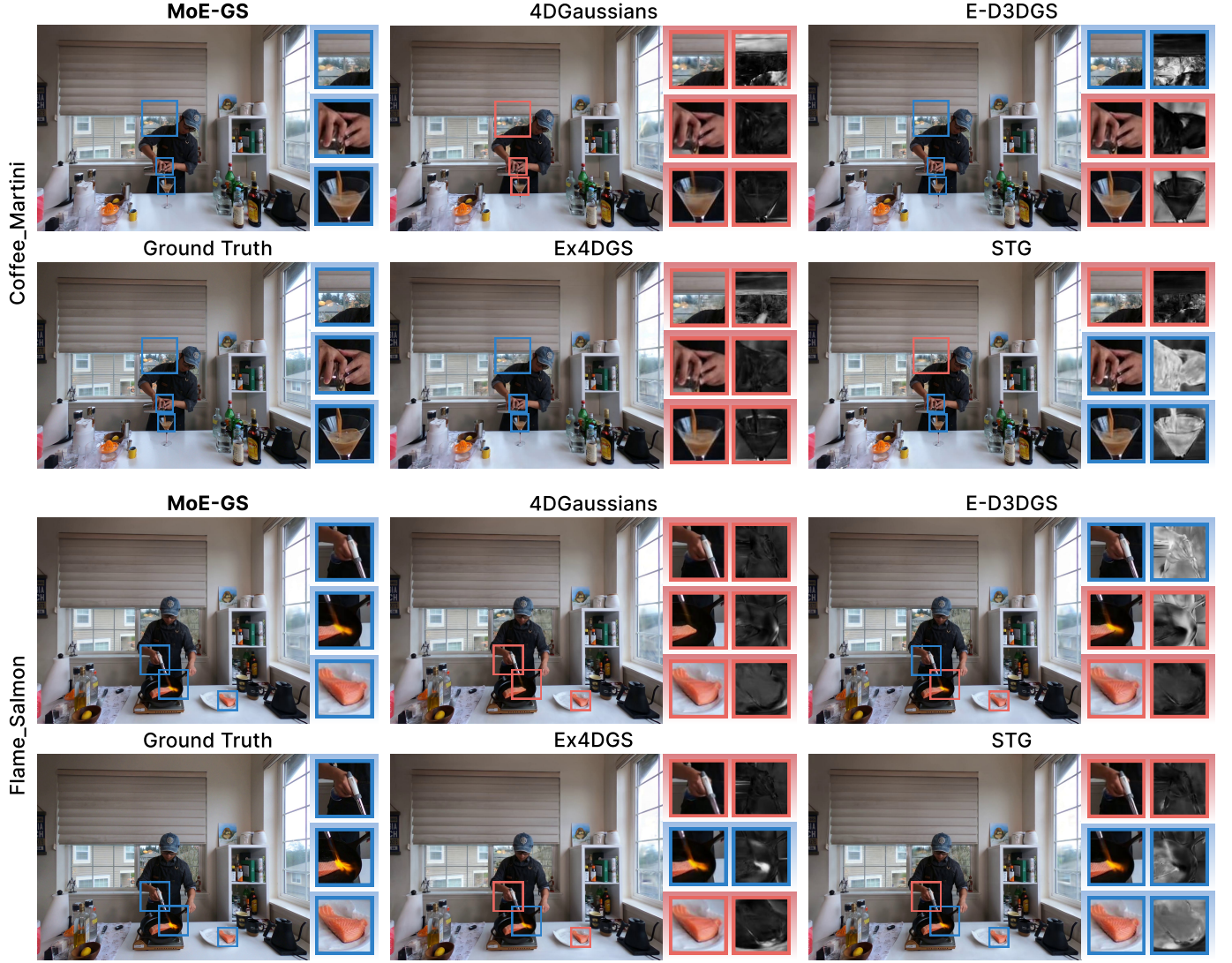}
  \caption{\textbf{Additional N3V Qualitative Results.} Comparison of our MoE-GS with other dynamic Gaussian splatting methods on the Neural 3D Video dataset~\cite{li2022neural}.}
  \label{n3vsupple}
  \vspace{-5pt}
\end{figure*}

\begin{figure*}[t]
\begin{center}
\centerline{\includegraphics[width=\textwidth]{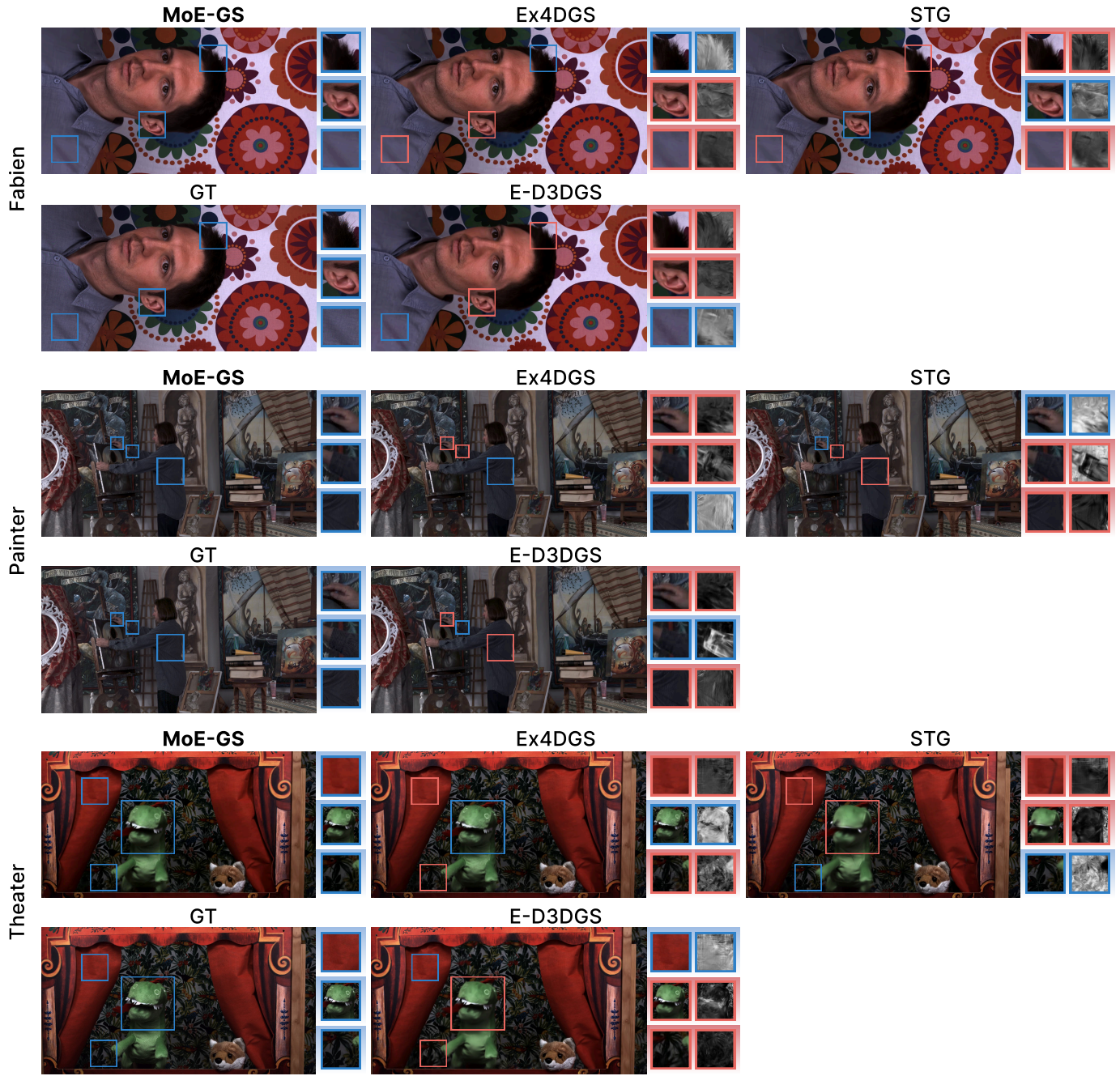}}
\caption{\textbf{Additional Qualitative Results on the Technicolor Dataset~\cite{sabater2017dataset}.} Visual comparison of our MoE-GS with other dynamic Gaussian splatting methods.}
\vspace{-3pt}
\label{techsupple}
\vspace{-10pt}
\end{center}
\end{figure*}

\begin{figure*}[t]
\begin{center}
\centerline{\includegraphics[width=\textwidth]{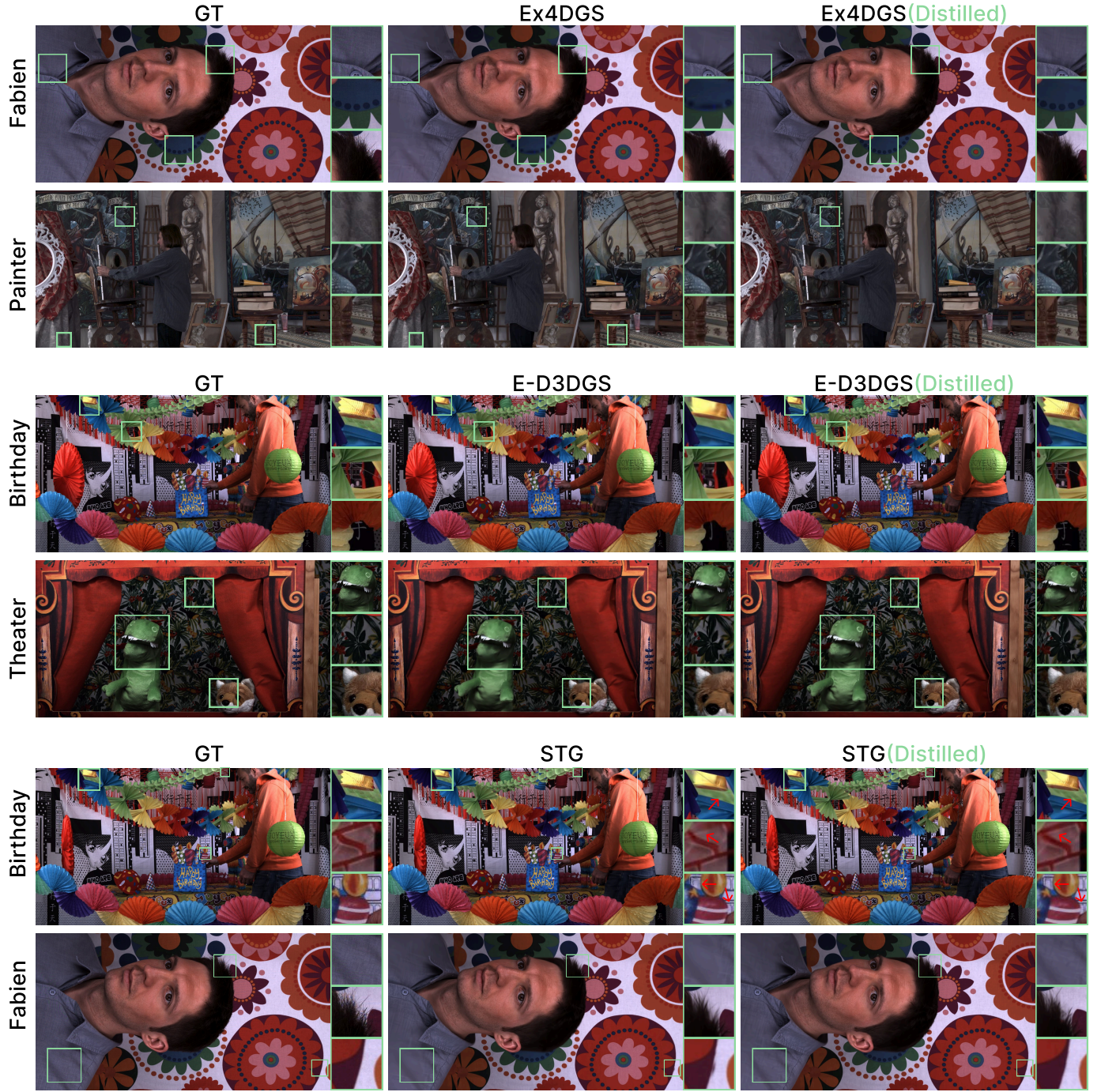}}
\caption{Visual comparison between retrained and distilled expert models on the Technicolor dataset~\cite{sabater2017dataset}.}
\vspace{-3pt}
\label{distillupple}
\vspace{-20pt}
\end{center}
\end{figure*}

Figures~\ref{n3vsupple},~\ref{techsupple} present additional qualitative comparisons of MoE-GS across different datasets.  
MoE-GS effectively routes scene regions to the most suitable experts, resulting in high-fidelity reconstructions that outperform individual models.  
These results further demonstrate the model's ability to adaptively combine specialized expert outputs for diverse dynamic scenes. In addition, Figure~\ref{distillupple} shows distillation qualitative results on the Technicolor dataset.

\subsubsection{Evaluation on Large-Motion Benchmarks}
\label{large_motion}
To further validate the effectiveness of MoE-GS under challenging dynamic scenarios, we additionally evaluate the proposed framework on two large-motion benchmarks: PanopticSports~\cite{joo2015panoptic} and D-NeRF~\cite{pumarola2021d}. Compared to N3V~\cite{li2022neural} and Technicolor~\cite{sabater2017dataset}, these datasets contain substantially larger and more complex motions, providing a more demanding testbed for expert specialization and composition.
Tables~\ref{tab:panoptic_large_motion} and~\ref{tab:dnerf_large_motion} report quantitative results using the MoE-GS (N=2) configuration. Across both datasets, MoE-GS consistently outperforms the individual experts and achieves the best average reconstruction quality. These results indicate that the proposed expert composition strategy generalizes beyond relatively static scenes and remains effective under large-motion conditions.

\vspace{-2mm}
\subsection{More Details on Ablation Studies of MoE-GS}
\label{sup_additonal_abl}
\vspace{-1mm}
\subsubsection{Training Cost and Budget Sensitivity}
\label{sup_additonal_abl_train}
We further analyze the training efficiency of MoE-GS from the perspectives of expert training budgets and router overhead.

\noindent\textbf{Partial expert training.}
Table~\ref{tab:abl_train_budget} reports the effect of partial expert training under different training budgets.
A budget of 100\% corresponds to the full training time of each expert
(approximately 4.2 hours for E-D3DGS~\cite{bae2024per}, 2.2 hours for
STG~\cite{li2024spacetime}, and 1.5 hours for
4DGaussians~\cite{wu20244d}), while 50\%, 20\%, and 10\% represent
proportional reductions in wall-clock time.
Thus, the budget axis directly reflects the actual training time per expert.

\noindent\textbf{Router overhead.}
The router itself is lightweight, adding less than 5\% computation relative to expert training. Consequently, its overhead is negligible in practice. These results indicate that the overall training cost of MoE-GS scales far more favorably than linearly with the number of experts.

\subsubsection{Distillation Weighting Strategies}
\label{sup_weigth_ablation}
Distillation in our framework aims to transfer the complementary 
reconstruction strengths identified by MoE-GS back into a single expert.
We explore whether assigning per-pixel weights to the MoE-guided loss 
improves this transfer, and compare two strategies:
(i) w/o weighting, and 
(ii) gating-based weighting derived from MoE-GS. 
All experiments use E\text{-}D3DGS~\cite{bae2024per} on the Technicolor~\cite{sabater2017dataset} dataset.
Using MoE-derived routing weights yields clear improvements over unweighted
distillation, as shown in Table~\ref{tab:distill_weight}.
These routing weights capture how different experts contribute
to each spatial–temporal region, providing supervision signals that encode
complementary deformation priors rather than relying on pixel-level error
statistics.
As a result, the distilled expert benefits from MoE-GS’s
region-specific strengths and achieves higher reconstruction quality.

\clearpage
\vspace{2mm}
\begin{table*}[t]
\centering
\caption{Additional Quantitative Results on Experts and MoE-GS for the N3V Dataset \cite{li2022neural}. {\small\textdagger}: Models were trained on a dataset split into 150 frames..}
\label{tab:appen_n3v_qunati}
\resizebox{0.9\textwidth}{!}{%
\begin{tabular}{@{}ccccccccccc@{}}
\toprule
\multirow{2}{*}{\raisebox{-0.6ex}{Model}}
                                  & \multicolumn{3}{c}{Coffee Martini}        & \multicolumn{3}{c}{Cook Spinach}          & \multicolumn{3}{c}{Cut Roast Beef}        \\ \cmidrule(lr){2-4} \cmidrule(lr){5-7}
                                  \cmidrule(lr){8-10}
             & PSNR  & SSIM & LPIPS & PSNR  & SSIM & LPIPS & PSNR  & SSIM & LPIPS             \\ \midrule
\multicolumn{1}{c|}{4DGaussians \cite{wu20244d}}      & 29.09 & 0.923 & \multicolumn{1}{c|}{0.066}  & 32.78 & 0.955 & \multicolumn{1}{c|}{0.041}  & 33.15 & 0.954 & 0.048              \\
\multicolumn{1}{c|}{E-D3DGS \cite{bae2024per}}         & 30.04 & 0.930 & \multicolumn{1}{c|}{0.058}  & 33.11 & 0.961 & \multicolumn{1}{c|}{0.041}  & 33.85 & 0.958 & 0.042              \\
\multicolumn{1}{c|}{STG {\small\textdagger} \cite{li2024spacetime}}          & 28.16 & 0.927 & \multicolumn{1}{c|}{0.061}  & 33.09 & 0.961 & \multicolumn{1}{c|}{0.033}  & 34.15 & 0.964 & 0.032              \\
\multicolumn{1}{c|}{Ex4DGS~\cite{lee2024fully}}  & 28.72 & 0.918 & \multicolumn{1}{c|}{0.070}  & 33.24 & 0.956 & \multicolumn{1}{c|}{0.042}  & 33.73 & 0.958 & 0.040              \\
\addlinespace[1.5pt]
\rowcolor{lightgray}
\multicolumn{1}{c|}{MoE-GS (N=4)} & \textbf{30.43} & \textbf{0.940} & \multicolumn{1}{c|}{\textbf{0.054}}  & \textbf{34.24} & \textbf{0.966} & \multicolumn{1}{c|}{\textbf{0.031}}  & \textbf{35.08} & \textbf{0.968} & \textbf{0.030}              \\ \bottomrule
\end{tabular}%
}

\resizebox{0.9\textwidth}{!}{%
\begin{tabular}{@{}ccccccccccc@{}}
\toprule
                                 \multirow{2}{*}{\raisebox{-0.6ex}{Model}}
                                  & \multicolumn{3}{c}{Flame Salmon}        & \multicolumn{3}{c}{Flame Steak}          & \multicolumn{3}{c}{Sear Steak}        \\ \cmidrule(lr){2-4} \cmidrule(lr){5-7}
                                  \cmidrule(lr){8-10}
             & PSNR  & SSIM & LPIPS & PSNR  & SSIM & LPIPS & PSNR  & SSIM & LPIPS             \\ \midrule
\multicolumn{1}{c|}{4DGaussians \cite{wu20244d}}      & 29.76 & 0.928 & \multicolumn{1}{c|}{0.062}  & 31.81 & 0.962 & \multicolumn{1}{c|}{0.032}  & 32.01 & 0.964 & 0.032              \\
\multicolumn{1}{c|}{E-D3DGS \cite{bae2024per}}         & 30.49 & 0.936 & \multicolumn{1}{c|}{0.054}  & 32.77 & 0.960 & \multicolumn{1}{c|}{0.037}  & 33.70 & 0.964 & 0.033              \\
\multicolumn{1}{c|}{STG {\small\textdagger} \cite{li2024spacetime}}          & 29.09 & 0.928 & \multicolumn{1}{c|}{0.057}  & 33.25 & 0.968 & \multicolumn{1}{c|}{\textbf{0.026}}  & 33.77 & 0.969 & \textbf{0.026}              \\
\multicolumn{1}{c|}{Ex4DGS~\cite{lee2024fully}}  & 29.33 & 0.925 & \multicolumn{1}{c|}{0.066}  & 33.91 & 0.963 & \multicolumn{1}{c|}{0.034}  & 33.69 & 0.960 & 0.035              \\
\addlinespace[1.5pt]
\rowcolor{lightgray}
\multicolumn{1}{c|}{MoE-GS (N=4)} & \textbf{30.92} & \textbf{0.942} & \multicolumn{1}{c|}{\textbf{0.049}}  & \textbf{34.38} & \textbf{0.972} & \multicolumn{1}{c|}{\textbf{0.026}}  & \textbf{34.42} & \textbf{0.972} & \textbf{0.026}              \\ \bottomrule
\end{tabular}%
}
\end{table*}

\vspace{-2mm}
\begin{table*}[t]
\centering
\caption{Additional Quantitative Results on Experts and MoE-GS for the Technicolor Dataset \cite{sabater2017dataset}.}
\label{tab:appen_tech_quanti}
\resizebox{0.9\textwidth}{!}{%
\begin{tabular}{@{}ccccccccccc@{}}
\toprule
\multirow{2}{*}{\raisebox{-0.6ex}{Model}}
                                  & \multicolumn{3}{c}{Birthday}        & \multicolumn{3}{c}{Fabien}          & \multicolumn{3}{c}{Painter}        \\ \cmidrule(lr){2-4} \cmidrule(lr){5-7}
                                  \cmidrule(lr){8-10}
             & PSNR  & SSIM & LPIPS & PSNR  & SSIM & LPIPS & PSNR  & SSIM & LPIPS             \\ \midrule
\multicolumn{1}{c|}
{DyNeRF~\cite{li2022neural}}       & 29.20 & \NA & \multicolumn{1}{c|}{0.067}  & 32.76 & \NA & \multicolumn{1}{c|}{0.242}  & 35.95 & \NA & 0.146              \\
{HyperReel~\cite{attal2023hyperreel}}       & 29.99 & 0.922 & \multicolumn{1}{c|}{0.053}  & 34.70 & 0.895 & \multicolumn{1}{c|}{0.186}  & 35.91 & 0.923 & 0.117              \\
\multicolumn{1}{c|}{4DGaussians \cite{wu20244d}}      & 30.87 & 0.904 & \multicolumn{1}{c|}{0.087}  & 33.56 & 0.854 & \multicolumn{1}{c|}{0.186}  & 34.36 & 0.884 & 0.136              \\
\multicolumn{1}{c|}{STG \cite{li2024spacetime}}          & 31.90 & 0.940 & \multicolumn{1}{c|}{\textbf{0.044}}  & 35.70 & 0.904 & \multicolumn{1}{c|}{\textbf{0.114}}  & 37.07 & 0.928 & 0.093              \\
\multicolumn{1}{c|}{Ex4DGS~\cite{lee2024fully}}  & 32.36 & 0.941 & \multicolumn{1}{c|}{0.045}  & 35.19 & 0.896 & \multicolumn{1}{c|}{0.124}  & 36.66 & 0.932 & 0.091              \\
\addlinespace[1.5pt]
\rowcolor{lightgray}
\multicolumn{1}{c|}{MoE-GS (N=3)} & \textbf{33.26} & \textbf{0.947} & \multicolumn{1}{c|}{0.049}  & \textbf{36.26} & \textbf{0.908} & \multicolumn{1}{c|}{0.121}  & \textbf{37.63} & \textbf{0.939} & \textbf{0.083}              \\ \bottomrule
\end{tabular}%
}

\resizebox{0.9\textwidth}{!}{%
\begin{tabular}{@{}ccccccccccc@{}}
\toprule
\multirow{2}{*}{\raisebox{-0.6ex}{Model}}
                                  & \multicolumn{3}{c}{Theater}        & \multicolumn{3}{c}{Train}          & \multicolumn{3}{c}{Average}        \\ \cmidrule(lr){2-4} \cmidrule(lr){5-7}
                                  \cmidrule(lr){8-10}
             & PSNR  & SSIM & LPIPS & PSNR  & SSIM & LPIPS & PSNR  & SSIM & LPIPS             \\ \midrule
\multicolumn{1}{c|}
{DyNeRF~\cite{li2022neural}}       & 29.53 & \NA & \multicolumn{1}{c|}{0.188}  & 31.58 & \NA & \multicolumn{1}{c|}{0.067}  & 31.80 & \NA & 0.142              \\
{HyperReel~\cite{attal2023hyperreel}}       & \textbf{33.32} & 0.895 & \multicolumn{1}{c|}{\textbf{0.115}}  & 29.74 & 0.895 & \multicolumn{1}{c|}{0.072}  & 32.73 & 0.906 & 0.109              \\
\multicolumn{1}{c|}{4DGaussians \cite{wu20244d}}      & 29.81 & 0.841 & \multicolumn{1}{c|}{0.155}  & 25.35 & 0.730 & \multicolumn{1}{c|}{0.166}  & 30.79 & 0.843 & 0.146              \\
\multicolumn{1}{c|}{STG \cite{li2024spacetime}}          & 31.08 & 0.879 & \multicolumn{1}{c|}{0.140}  & 32.32 & 0.937 & \multicolumn{1}{c|}{\textbf{0.045}}  & 33.61 & 0.918 & 0.087              \\
\multicolumn{1}{c|}{Ex4DGS~\cite{lee2024fully}}  & 31.79 & 0.882 & \multicolumn{1}{c|}{0.130}  & 31.39 & 0.928 & \multicolumn{1}{c|}{0.055}  & 33.48 & 0.916 & 0.089              \\
\addlinespace[1.5pt]
\rowcolor{lightgray}
\multicolumn{1}{c|}{MoE-GS (N=3)} & 32.88 & \textbf{0.900} & \multicolumn{1}{c|}{\textbf{0.115}}  & \textbf{32.89} & \textbf{0.944} & \multicolumn{1}{c|}{0.046}  & \textbf{34.58} & \textbf{0.928} & \textbf{0.083}              \\ \bottomrule
\end{tabular}%
}
\end{table*}

\begin{table*}[t]
\centering
\caption{Comparison results on the HyperNeRF dataset \cite{park2021hypernerf}.}
\resizebox{0.6\textwidth}{!}{
\label{tab:monocular}
\begin{tabular}{lccccc}
\toprule
\multirow{2}{*}{Model} & \multicolumn{5}{c}{PSNR (dB) $\uparrow$} \\
\cmidrule(lr){2-6}
 & 3dprinter & banana & broom & chicken & Average \\
\midrule
4DGaussians~\cite{wu20244d} & 22.16 & 22.90 & 20.88 & 30.12 & 24.02 \\
E-D3DGS~\cite{bae2024per}     & 22.41 & 23.38 & 20.07 & 29.11 & 23.74 \\
\addlinespace[1.5pt]
\rowcolor{lightgray}
\textbf{MoE-GS (N=2)} & \textbf{22.84} & \textbf{24.75} & \textbf{21.26} & \textbf{30.37} & \textbf{24.81} \\
\bottomrule
\end{tabular}}
\end{table*}

\begin{table*}[t!]
\centering
\caption{Ablation study on distillation strategies evaluating the effect of routing-weight-based adaptive supervision on the Technicolor dataset~\cite{sabater2017dataset}.}
\label{tab:distillation_ablation}
\resizebox{\textwidth}{!}{%
\begin{tabular}{@{}cc|ccc|ccc|ccc@{}}
\toprule
\multirow{2}{*}{\raisebox{-0.6ex}{Model}}
& \multirow{2}{*}{\raisebox{-0.6ex}{Training}} & \multicolumn{3}{c}{Birthday} & \multicolumn{3}{c}{Fabien} & \multicolumn{3}{c}{Painter} \\
\cmidrule(lr){3-5} \cmidrule(lr){6-8} \cmidrule(lr){9-11}
& & PSNR & SSIM & LPIPS & PSNR & SSIM & LPIPS & PSNR & SSIM & LPIPS \\
\midrule
\multirow{3}{*}{E-D3DGS~\cite{bae2024per}} 
  & Retrained (GT Loss)           & 32.05 & 0.936   & 0.050 & 34.7 & 0.878   & 0.171 & 36.26 & 0.931   & 0.089 \\
  & Distilled (w/o Weight)       & 32.17 & 0.942 & 0.048 & 34.8 & 0.883 & 0.164 & 36.77 & 0.934 & 0.089 \\
  & \cellcolor{lightgray}Distilled (Ours)              & \cellcolor{lightgray} \textbf{32.24} & \cellcolor{lightgray} \textbf{0.946} &
  \cellcolor{lightgray}
  \textbf{0.038} & \cellcolor{lightgray} \textbf{35.68} & \cellcolor{lightgray} \textbf{0.902} & \cellcolor{lightgray} \textbf{0.120} & \cellcolor{lightgray} \textbf{37.20} & \cellcolor{lightgray} \textbf{0.939} & \cellcolor{lightgray} \textbf{0.078} \\
\midrule
\multirow{3}{*}{STG~\cite{li2024spacetime}} 
  & Retrained (GT Loss)           & 33.15 & 0.947 & \textbf{0.038} & 34.87 & 0.901 & \textbf{0.117} & 33.61 & 0.907 & 0.098 \\
  & Distilled (w/o Weight)       & 33.27 & 0.948 & 0.040 & 34.99 & 0.901 & 0.188 & \textbf{33.66} & 0.907 & 0.099 \\
  & \cellcolor{lightgray} Distilled (Ours)              & \cellcolor{lightgray} \textbf{33.46} & \cellcolor{lightgray} \textbf{0.949} & \cellcolor{lightgray} 0.039 & \cellcolor{lightgray} \textbf{35.01} & \cellcolor{lightgray} \textbf{0.902} & \cellcolor{lightgray} \textbf{0.117} & \cellcolor{lightgray} 33.51 & \cellcolor{lightgray} \textbf{0.908} & \cellcolor{lightgray} \textbf{0.094} \\
\midrule
\multirow{3}{*}{Ex4DGS~\cite{lee2024fully}} 
  & Retrained (GT Loss)           & 32.18 & 0.944 & 0.039 & 35.33 & 0.896 & 0.124 & 36.40 & 0.930 & 0.094 \\
  & Distilled (w/o Weight)       & 32.39 & 0.945 & 0.041 & 35.44 & 0.896 & 0.126 & 36.37 & 0.930 & 0.095 \\
  & \cellcolor{lightgray} Distilled (Ours)              &\cellcolor{lightgray} \textbf{32.41} & \cellcolor{lightgray} \textbf{0.946} & \cellcolor{lightgray} \textbf{0.038} & \cellcolor{lightgray} \textbf{35.88} & \cellcolor{lightgray} \textbf{0.903} & \cellcolor{lightgray} \textbf{0.115} & \cellcolor{lightgray} \textbf{36.87} & \cellcolor{lightgray} \textbf{0.935} & \cellcolor{lightgray} \textbf{0.086} \\
\bottomrule
\end{tabular}
}

\vspace{1em}
\centering
\resizebox{\textwidth}{!}{%
\begin{tabular}{@{}cc|ccc|ccc|ccc@{}}
\toprule
\multirow{2}{*}{\raisebox{-0.6ex}{Model}}
& \multirow{2}{*}{\raisebox{-0.6ex}{Training}} & \multicolumn{3}{c}{Theater} & \multicolumn{3}{c}{Train} & \multicolumn{3}{c}{Average} \\
\cmidrule(lr){3-5} \cmidrule(lr){6-8} \cmidrule(lr){9-11}
& & PSNR & SSIM & LPIPS & PSNR & SSIM & LPIPS & PSNR & SSIM & LPIPS \\
\midrule
\multirow{3}{*}{E-D3DGS~\cite{bae2024per}} 
  & Retrained (GT Loss)           & 30.49 & 0.871 & 0.148 & 30.92 & 0.896 & 0.097 & 32.88 & 0.902 & 0.111 \\
  & Distilled (w/o Weight)       & 31.73 & 0.878 & 0.150 & 30.85 & 0.898 & 0.099 & 33.26 & 0.907 & 0.110 \\
  & \cellcolor{lightgray} Distilled (Ours)              & \cellcolor{lightgray} \textbf{31.79} & \cellcolor{lightgray} \textbf{0.887} & \cellcolor{lightgray} \textbf{0.124} & \cellcolor{lightgray} \textbf{31.46} & \cellcolor{lightgray} \textbf{0.900} & \cellcolor{lightgray} \textbf{0.095} & \cellcolor{lightgray} \textbf{33.67} & \cellcolor{lightgray} \textbf{0.915} & \cellcolor{lightgray} \textbf{0.091} \\
\midrule
\multirow{3}{*}{STG~\cite{li2024spacetime}} 
  & Retrained (GT Loss)           & 30.28 & 0.876   & 0.126 & 32.26 & 0.942   & \textbf{0.036} & 32.83 & 0.915   & 0.083 \\
  & Distilled (w/o Weight)       & 31.23 & \textbf{0.883} & \textbf{0.124} & \textbf{32.30} & \textbf{0.943} & 0.039 & 33.09 & \textbf{0.917} & 0.084 \\
  & \cellcolor{lightgray} Distilled (Ours)              & \cellcolor{lightgray} \textbf{31.35} & \cellcolor{lightgray} \textbf{0.883} & \cellcolor{lightgray} \textbf{0.124} & \cellcolor{lightgray} 32.20 & \cellcolor{lightgray} \textbf{0.943} & \cellcolor{lightgray} 0.038 & \cellcolor{lightgray} \textbf{33.11} & \cellcolor{lightgray} \textbf{0.917} & \cellcolor{lightgray} \textbf{0.082} \\
\midrule
\multirow{3}{*}{Ex4DGS~\cite{lee2024fully}} 
  & Retrained (GT Loss)           & 31.85 & 0.886 & 0.122 & 32.11 & 0.934 & 0.050 & 33.57 & 0.918 & 0.086 \\
  & Distilled (w/o Weight)       & 31.78 & 0.884 & 0.126 & 32.14 & 0.935 & 0.053 & 33.62 & 0.918 & 0.088 \\
  & \cellcolor{lightgray} Distilled (Ours)              & \cellcolor{lightgray} \textbf{32.04} & \cellcolor{lightgray} \textbf{0.890} & \cellcolor{lightgray} \textbf{0.111} & \cellcolor{lightgray} \textbf{32.37} & \cellcolor{lightgray} \textbf{0.939} & \cellcolor{lightgray} \textbf{0.045} & \cellcolor{lightgray} \textbf{33.91} & \cellcolor{lightgray} \textbf{0.923} & \cellcolor{lightgray} \textbf{0.079} \\
\bottomrule
\end{tabular}
}
\end{table*}

\begin{table*}[t]
\centering
\caption{Comparison results on PanopticSports dataset~\cite{joo2015panoptic}. N = 2: \{4DGaussians~\cite{wu20244d}, Grid4D~\cite{xu2024grid4d}\}.}
\label{tab:panoptic_large_motion}
\resizebox{0.72\textwidth}{!}{
\begin{tabular}{lccccccc}
\toprule
Model & Basketball & Boxes & Football & Juggle & Softball & Tennis & Average \\
\midrule
4DGaussians & 26.48 & 25.27 & 26.35 & 26.51 & 27.32 & 25.57 & 26.25 \\
Grid4D      & 26.12 & 26.15 & 27.81 & 26.82 & 25.65 & 26.54 & 26.69 \\
\rowcolor{lightgray}
\textbf{MoE-GS (N=2)}
             & \textbf{27.11} & \textbf{26.81} & \textbf{27.90} & \textbf{27.62} & \textbf{27.45} & \textbf{26.77} & \textbf{27.24} \\
\bottomrule
\end{tabular}
}
\vspace{-1mm}
\end{table*}

\begin{table*}[t]
\centering
\caption{Comparison results on D-NeRF dataset~\cite{pumarola2021d}. N = 2: \{DeformableGS~\cite{yang2024deformable}, Grid4D~\cite{xu2024grid4d}\}.}
\label{tab:dnerf_large_motion}
\resizebox{\textwidth}{!}{
\begin{tabular}{lccccccccc}
\toprule
Model & Bouncingballs & Hellwarrior & Hook & Jumpingjacks & Lego & Mutant & Standup & Trex & Average \\
\midrule
DeformableGS & 41.45 & 41.39 & 36.77 & 37.55 & 24.93 & 42.03 & 43.73 & 38.22 & 40.16 \\
Grid4D       & 42.51 & 42.93 & 38.94 & 39.20 & 24.88 & 43.95 & 46.62 & 40.40 & 42.08 \\
\rowcolor{lightgray}
\textbf{MoE-GS (N=2)} 
             & \textbf{43.01} & \textbf{43.10} & \textbf{39.22} & \textbf{39.56} & \textbf{25.03} & \textbf{44.11} & \textbf{46.80} & \textbf{40.72} & \textbf{42.36} \\
\bottomrule
\end{tabular}
}
\end{table*}

\begin{table*}[t!]
\centering
\caption{Effect of expert training budget on MoE-GS (N=3). $N\!=\!3$: \{Ex4DGS~\cite{lee2024fully}, STG~\cite{li2024spacetime}, E-D3DGS~\cite{bae2024per}\}.}
\label{tab:abl_train_budget}
\resizebox{0.5\linewidth}{!}{%
\begin{tabular}{l|cccc}
\toprule
Model & 100\% & 50\% & 20\% & 10\% \\
\midrule
E-D3DGS~\cite{bae2024per}       & 32.33 & 32.19 & 31.87 & 30.60 \\
STG~\cite{li2024spacetime}           & 31.92 & 31.76 & 31.41 & 31.19 \\
4DGaussians~\cite{wu20244d}   & 31.43 & 31.02 & 30.90 & 30.64 \\
\rowcolor{lightgray}
\textbf{MoE-GS (N=3)} & \textbf{33.23} & \textbf{32.71} & \textbf{32.60} & \textbf{32.14} \\
\bottomrule
\end{tabular}
}
\vspace{-1mm}
\end{table*}

\begin{table*}
\centering
\small
\caption{Effect of routing-weighted distillation.}
\label{tab:distill_weight}
\vspace{-1mm}
\begin{tabular}{lccc}
\toprule
Method & PSNR (dB) $\uparrow$ & SSIM $\uparrow$ & LPIPS $\downarrow$ \\
\midrule
E-D3DGS (Baseline)     & 32.88 & 0.902 & 0.111 \\
w/o Weight            & 33.26 & 0.907 & 0.110 \\
\addlinespace[1.5pt]
\rowcolor{lightgray}
\textbf{Routing Weight (Ours)}  & \textbf{33.67} & \textbf{0.918} & \textbf{0.091} \\
\bottomrule
\end{tabular}
\end{table*}

\FloatBarrier
\clearpage 

\clearpage

\FloatBarrier

\newpage
\subsection{Gaussian-Level Interpretation and Geometry Evaluation}
\label{MoE_3D_interp}
 \subsubsection{Per-Gaussian Contributions and Responsibilities}
\label{MoE_3D_per_gau}
As described in Section~\ref{sec_moegs}, the router outputs per-pixel expert weights
$G'_k(u)$, while each Gaussian $j$ of expert $k$ contributes to pixel $u$
through its volumetric transmittance $T_{k,j}(u)$ and opacity $\alpha_{k,j}(u)$.
We first define the raw volumetric contribution of Gaussian $g_{k,j}$ as
\begin{equation}
C_{k,j}(u)=T_{k,j}(u)\,\alpha_{k,j}(u),
\end{equation}
which measures how much Gaussian $j$ influences the ray at pixel $u$.
To lift the pixel-level routing back to the Gaussian domain, we weight this
contribution by the router’s pixel gate:
\begin{equation}
\tilde{C}_{k,j}(u)=G'_k(u)\,C_{k,j}(u).
\end{equation}
Because different Gaussians vary in visibility, opacity, and pixel coverage,
their raw gated contributions $\tilde{C}_{k,j}$ reflect both geometric scale and
routing strength. To isolate the routing effect, we normalize by each Gaussian’s
total volumetric contribution:
\begin{equation}
\bar{R}_{k,j} =
\frac{\sum_{u \in \mathcal{U}} \tilde{C}_{k,j}(u)}
     {\sum_{u \in \mathcal{U}} C_{k,j}(u)},
\end{equation}
where $\mathcal{U}$ denotes the set of all pixel rays from all training views.
The resulting $\bar{R}_{k,j}$ measures how strongly Gaussian $j$ is utilized
within expert $k$ under the router’s pixel-level gates, capturing its
effective contribution in the image regions where it is visible. 
This interpretation arises naturally from our volume-aware router: because 
Gaussian contributions are computed in 3D before rasterization, pixel-level 
gating can be consistently traced back to individual Gaussians, enabling a 
geometry-informed rather than RGB-level interpretation of MoE-GS.

Moreover, different experts operate in heterogeneous deformation spaces, so their
Gaussians do not form a one-to-one correspondence in 3D. Pixel space, however,
is a shared observation domain across all experts: each Gaussian contributes to
the same set of rays through $T_{k,j}(u)\alpha_{k,j}(u)$.  
The router weights $G'_k(u)$ therefore gate \emph{geometry-conditioned}
volumetric contributions, making the lifting operation well-defined and ensuring
that the resulting Gaussian responsibilities preserve consistent 3D semantics
rather than reflecting any form of 2D blending.

\subsubsection{Post-hoc Gaussian Fusion using Lifting Weights}
\label{Post_hoc}

Using the normalized responsibilities $\bar{R}_{k,j}$ from Section~\ref{MoE_3D_per_gau}, we construct
a post-hoc unified Gaussian model without any retraining.  
For each Gaussian $j$ in expert $k$, we keep all geometry attributes—
position, scale, rotation, and SH coefficients—unchanged, and adjust only the
opacity based on the responsibility weight:
\begin{equation}
\alpha^{\mathrm{fused}}_{k,j}
= \bar{R}_{k,j}\,\alpha_{k,j}.
\end{equation}
The final fused model is obtained by simply concatenating all Gaussians from
all experts, using the fused opacities.  
This unified Gaussian set is rendered with the standard 3D Gaussian Splatting
volume renderer—\emph{without} any 2D compositing.  
The result preserves each expert’s geometric structure while modulating its
influence according to the router-driven responsibilities, yielding a coherent
volumetric representation that reflects geometry-informed routing rather than
image-space blending.

\subsubsection{Multi-view Depth Consistency Evaluation}
\label{MDC}
To quantitatively evaluate the geometry of our post-hoc fused Gaussian model, 
we compute the Multi-view Depth Consistency (MDC), defined as the mean 
reprojection error between depth maps across all viewpoint pairs at the same 
timestamp.  
Given rendered depth maps $\{D_i^t\}$ from viewpoints $\{v_i\}$ at time $t$, 
we measure consistency over all ordered pairs $(i,j)$ by reprojecting the depth 
from view $i$ into view $j$ and comparing it against $D_j^t$:
\begin{equation}
\mathrm{MDC}
=
\frac{1}{|\mathcal{P}|}
\sum_{(i,j)\in\mathcal{P}}
\big\|
\Pi_{v_j}(D_i^t) - D_j^t
\big\|_1,
\end{equation}
where $\mathcal{P}$ denotes the set of all ordered view pairs, and 
$\Pi_{v_j}(D_i^t)$ is the reprojection of $D_i^t$ into view $j$ using the known 
camera poses at timestamp $t$.

\subsection{Analysis of Expert-Specific Motion Behavior}
\label{expert_analysis}
\begin{figure}[t]
\begin{center}
\centerline{\includegraphics[width=\columnwidth]{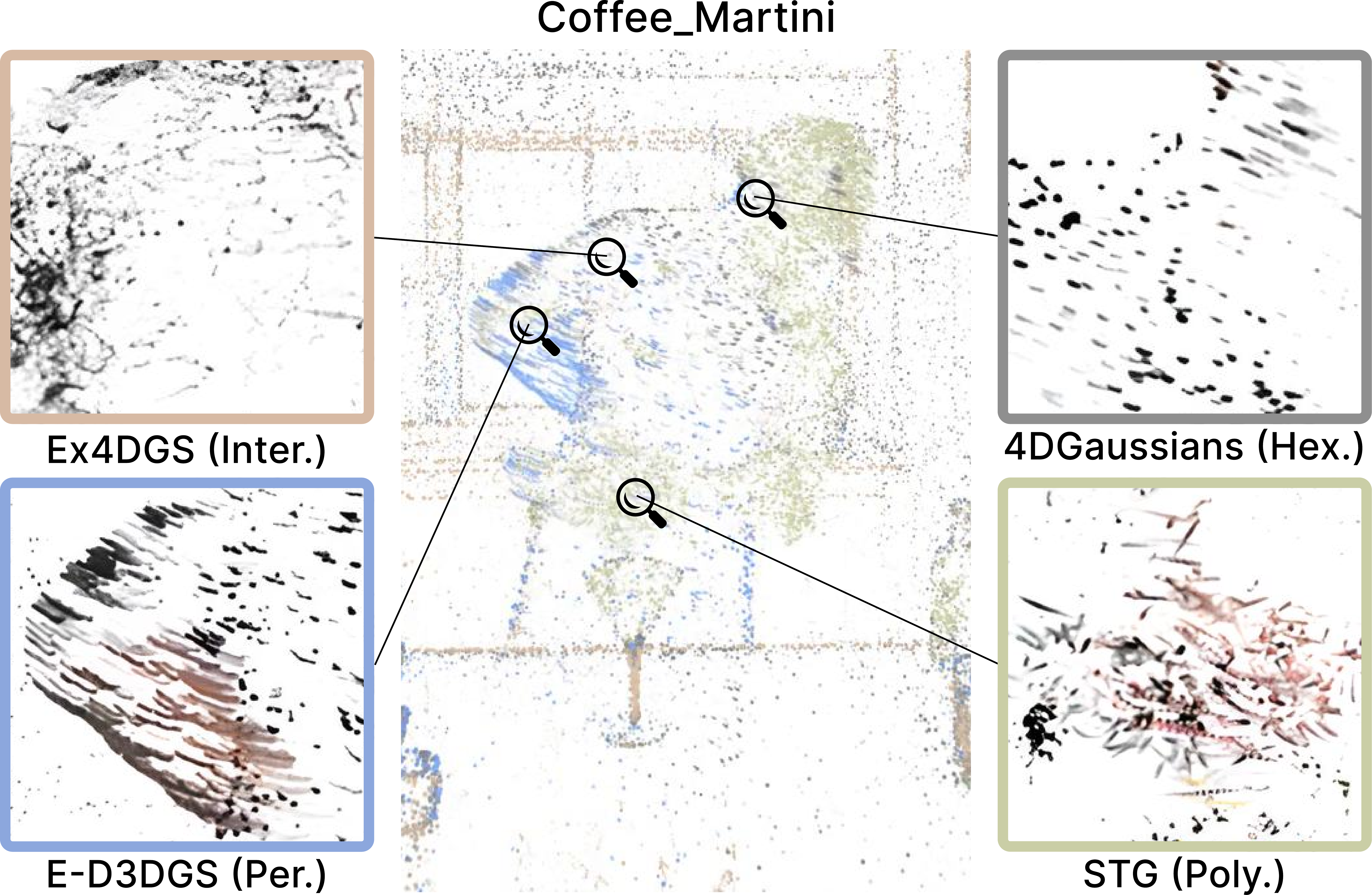}}
\caption{\textbf{Motion Trajectory 1} Comparison across dynamic Gaussian Splatting methods.}
\vspace{-3pt}
\label{motion}
\vspace{-10pt}
\end{center}
\end{figure}

\begin{figure}[t]
\begin{center}
\centerline{\includegraphics[width=\columnwidth]{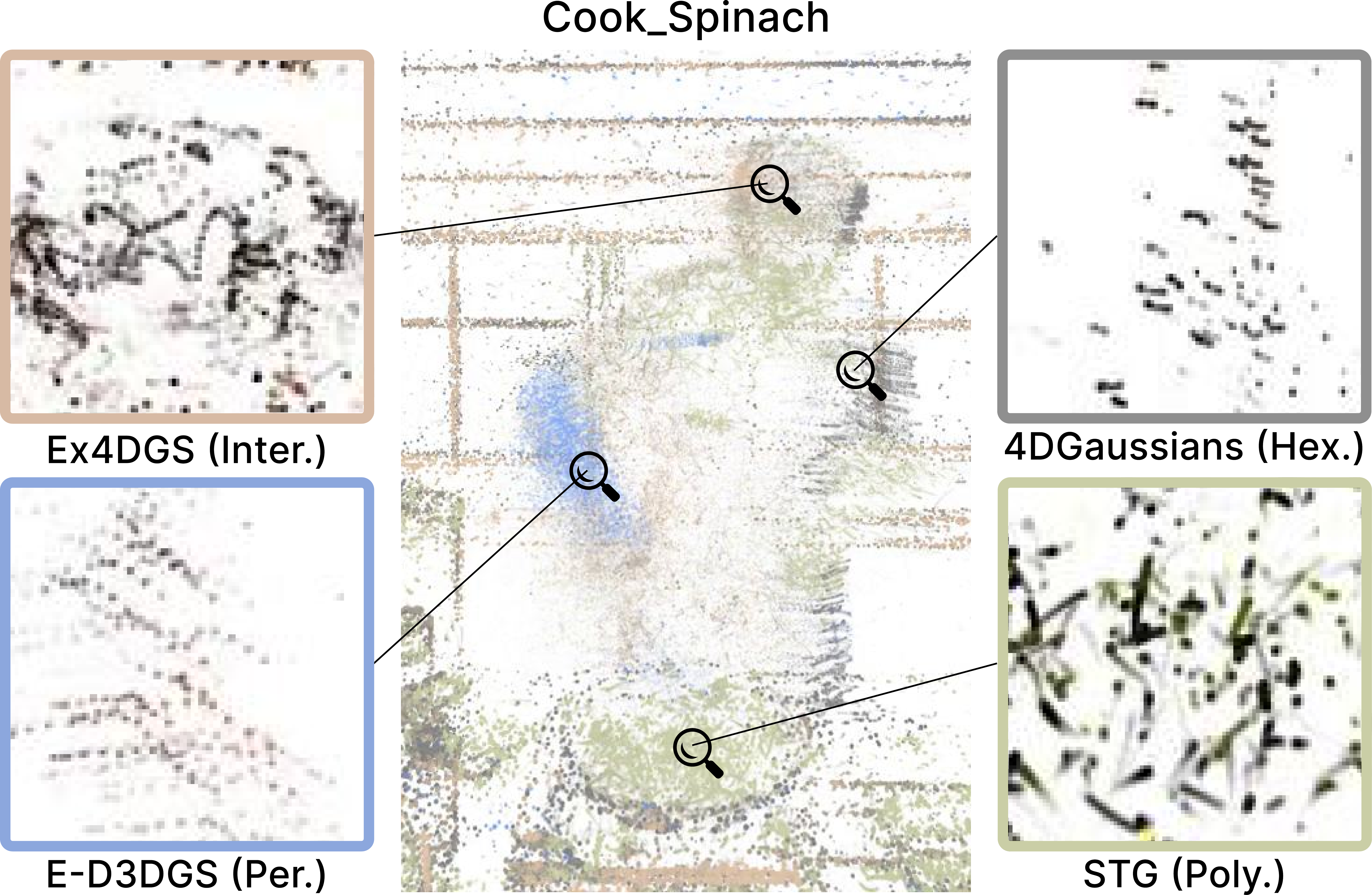}}
\caption{\textbf{Motion Trajectory 2} Comparison across dynamic Gaussian Splatting methods.}
\vspace{-3pt}
\label{motion2}
\vspace{-10pt}
\end{center}
\end{figure}

\begin{figure}[t]
\begin{center}
\centerline{\includegraphics[width=\columnwidth]{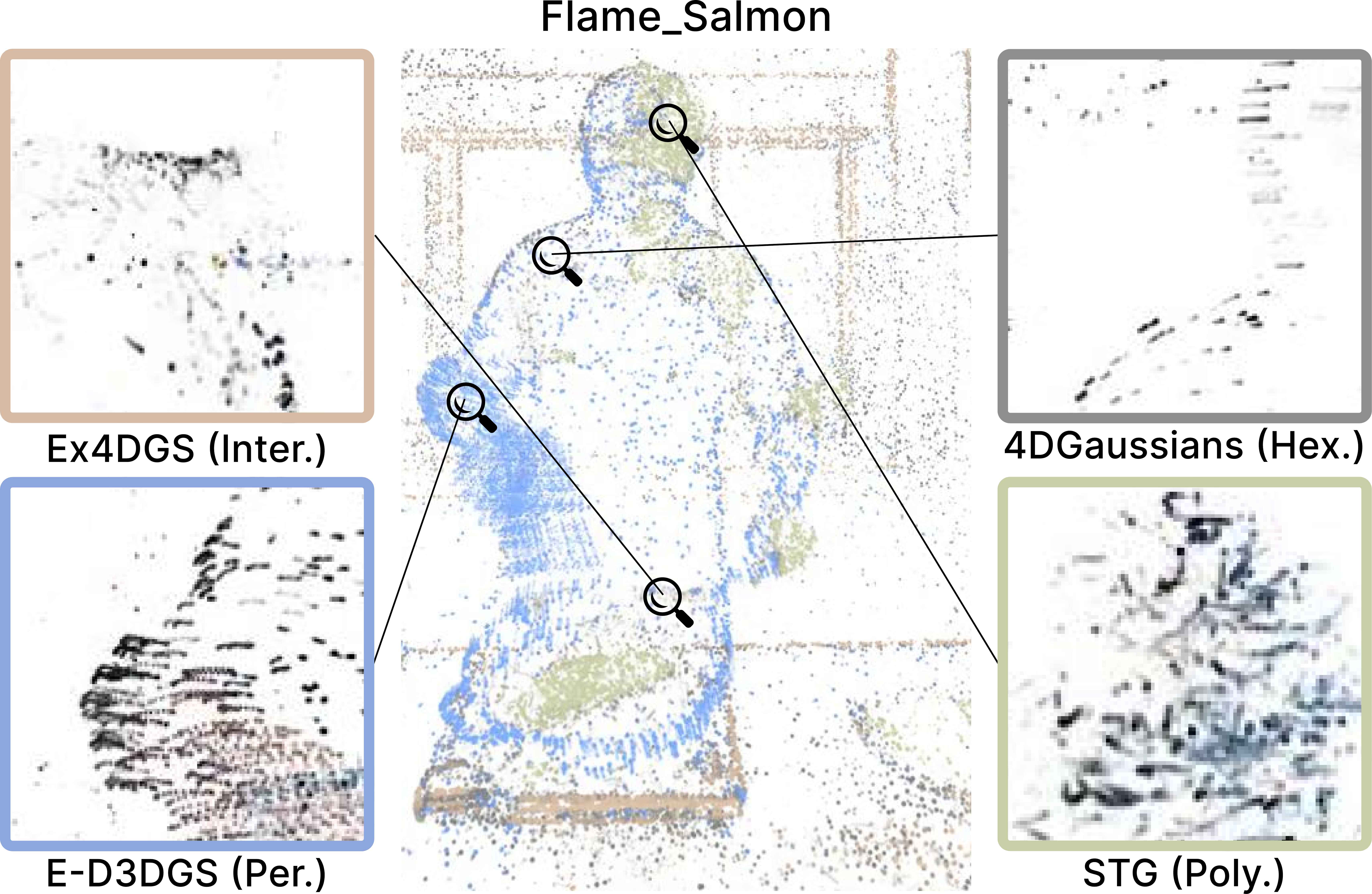}}
\caption{\textbf{Motion Trajectory 3} Comparison across dynamic Gaussian Splatting methods.}
\vspace{-3pt}
\label{motion3}
\vspace{-10pt}
\end{center}
\end{figure}

To further support the expert-specific motion analysis, we include additional trajectory examples across multiple scenes (Figs.~\ref{motion}, \ref{motion2}, and \ref{motion3}).
Although all methods operate on Gaussian primitives, their deformation priors impose fundamentally different inductive biases, leading to distinct behaviors across spatial regions and temporal motion regimes. Below, we summarize the key tendencies of each expert based on these examples.

\noindent\textbf{4DGaussians~\cite{wu20244d} (HexPlane canonical deformation).}
4DGaussians produces short, smooth, and highly regular trajectories due to its shared HexPlane canonical representation. The deformation MLP conditions on features interpolated from a global position--time grid, causing spatially adjacent Gaussians to receive highly correlated deformation signals. This results in strong spatial regularization and minimal per-Gaussian variation, which is beneficial for \emph{static or low-motion regions}. 
Regions associated with 4DGaussians consistently exhibit compact and spatially coherent trajectories across the visualized examples. This tendency is particularly evident in regions with limited motion, such as the head or upper-body regions in Coffee Martini (Fig.~\ref{motion}) and Cook Spinach (Fig.~\ref{motion2}), where neighboring Gaussians show highly similar motion patterns. Similar compact and regular trajectories are also observed in Flame Salmon (Fig.~\ref{motion3}), suggesting that this behavior remains consistent across different scenes.
However, the same bias can hinder the representation of \emph{high-speed or rapidly changing motion}, where the canonical features may not vary sufficiently to encode fine-grained displacement.

\noindent\textbf{E-D3DGS~\cite{bae2024per} (per-Gaussian volumetric deformation).}
E-D3DGS exhibits directionally consistent and comparatively higher-velocity trajectories. This behavior arises from two architectural elements: (i) a two-branch deformation network that separately models coarse and fine motion, allowing the model to represent \emph{fast and detailed dynamics}; and (ii) per-Gaussian embeddings with local embedding regularization, which encourage neighboring Gaussians to share similar deformation features and form spatially coherent motion clusters.
Across the visualized examples, regions associated with E-D3DGS correspond to the most dynamic articulated parts, such as the arms in Coffee Martini (Fig.~\ref{motion}) and Cook Spinach (Fig.~\ref{motion2}), and the upper-body/arm region in Flame Salmon (Fig.~\ref{motion3}). These regions repeatedly exhibit longer and directionally aligned trajectories, indicating locally coherent yet high-velocity motion.
These repeated patterns indicate that E-D3DGS is particularly effective for \emph{high-velocity motion} that still preserves local spatial coherence.

\noindent\textbf{Ex4DGS~\cite{lee2024fully} (keyframe interpolation).}
Ex4DGS produces highly diverse and often ``free-form'' trajectories, even among spatially adjacent Gaussians. This behavior stems from its interpolation-based formulation, where each Gaussian independently updates its position between keyframes without relying on a shared deformation field or an explicit spatial-coherence constraint. As a result, nearby Gaussians may move with different magnitudes or directions. Across the visualized examples in Coffee Martini (Fig.~\ref{motion}), Cook Spinach (Fig.~\ref{motion2}), and Flame Salmon (Fig.~\ref{motion3}), regions associated with Ex4DGS exhibit more scattered trajectories than those associated with canonical deformation-based experts, particularly where neighboring Gaussians follow different motion directions. This flexibility is useful for \emph{irregular or rapidly changing motion}, especially when abrupt and multi-directional displacements are required.

\noindent\textbf{STG~\cite{li2024spacetime} (polynomial trajectory model).}
STG produces globally smooth, low-curvature trajectories due to its low-order polynomial motion parameterization. Unlike Ex4DGS, which relies on independent interpolation, STG learns per-Gaussian polynomial coefficients while sharing the same polynomial motion form across Gaussians. This provides moderate local flexibility while maintaining a strong bias toward smoothly varying and spatially aligned motion. Across the visualized examples, regions associated with STG consistently exhibit smooth trajectories with limited curvature and strong directional consistency. This tendency is particularly evident in regions exhibiting smooth and coherent motion, such as the tumbler in Coffee Martini (Fig.~\ref{motion}), the spinach bundle in Cook Spinach (Fig.~\ref{motion2}), and the hat in Flame Salmon (Fig.~\ref{motion3}), where motion is largely rigid or evolves gradually over time. These observations indicate that STG is well suited for \emph{rigid or near-rigid global motion} and \emph{low-frequency temporal changes}.

We emphasize that these motion tendencies are representative rather than absolute, as real-world scenes often contain mixed or ambiguous motion regimes that do not align perfectly with any single deformation prior. Nevertheless, the recurring patterns observed across multiple scenes suggest that different deformation formulations induce consistent expert-specific motion behaviors. This observation explains why no existing dynamic GS method consistently dominates across all spatial or temporal regions and further motivates the MoE-GS formulation, which adaptively combines experts with complementary motion priors.

\vfill

\end{document}